%% file: template.tex
\documentclass[preprints,article,accept,moreauthors,pdftex]{Definitions/mdpi}
\usepackage{pgfplots}
\usepackage{siunitx}
\usepackage{amssymb}
\usetikzlibrary{fadings}
\usetikzlibrary{patterns}
\usetikzlibrary{shadows.blur}
\usetikzlibrary{shapes}
\usepackage{pgfplotstable}
\pgfplotsset{compat=1.17}

\newcommand{\R}{\mathbb{R}}

\firstpage{1} 
\pubvolume{xx}
\issuenum{1}
\articlenumber{5}
\pubyear{2019}
\copyrightyear{2019}
\history{Received: date; Accepted: date; Published: date}

\Title{\LARGE \bf
Deeply supervised UNet for semantic segmentation to assist dermatopathological assessment of Basal Cell Carcinoma (BCC)
}

\Author{Jean Le'Clerc Arrastia $^{1}$, Nick Heilenkötter$^{1}$, Daniel Otero Baguer $^{1}$, Lena Hauberg-Lotte$^{1}$, Tobias Boskamp$^{2}$, Sonja Hetzer$^{3}$, Nicole Duschner$^{3}$, Jörg Schaller$^{3}$, and Peter Maaß $^{1}$}

\AuthorNames{J. Le'Clerc Arrastia, N. Heilenkötter, D. Otero Baguer, L. Hauberg-Lotte, T. Boskamp, S. Hetzer, N. Duschner, J. Schaller, and P. Maass}

\address{%
$^{1}$ \quad University of Bremen, ZeTeM, Bremen, Germany\\
$^{2}$ \quad SCiLS, Bruker Daltonik, Bremen, Germany\\
$^{3}$ \quad Dermatopathologie Duisburg, Duisburg, Germany}

\corres{Correspondence: leclerc@uni-bremen.de}

\abstract{%
Accurate and fast assessment of resection margins is an essential part of a dermatopathologist's clinical routine. In this work, we successfully develop a deep learning method to assist the pathologists by marking critical regions that have a high probability of exhibiting pathological features in Whole Slide Images (WSI). We focus on detecting Basal Cell Carcinoma (BCC) through semantic segmentation using several models based on the UNet architecture. The study includes $650$ WSI with $3443$ tissue sections in total. Two clinical dermatopathologists annotated the data, marking tumor tissues' exact location on $100$ WSI. The rest of the data, with ground-truth section-wise labels, is used to further validate and test the models. We analyze two different encoders for the first part of the  UNet network and two additional training strategies: a) deep supervision, b) linear combination of decoder outputs, and obtain some interpretations about what the network's decoder does in each case. The best model achieves over \SI{96}{\%}, accuracy, sensitivity, and specificity on the test set.}

\keyword{Digital Pathology; Whole Slide Image; Basal Cell Carcinoma; Skin Cancer; Deep Learning; UNet}

\begin{document}

\section{Introduction}

Basal Cell Carcinoma (BCC) is the most common malignant skin cancer with an increasing incidence by up to \SI{10}{\%} a year \cite{Wong794}. It can be locally destructive and is an essential source of morbidity for patients, mainly when located on the face. Thus, it must be adequately treated, despite its slow growth~\cite{Sahl1995}. Although BCC can be effectively managed through surgical excision, determining the most suitable surgical margins is often not trivial. Complete removal of the pathological tissue is the key to a successful surgical treatment. Initially, the tumor is removed with a safety margin of surrounding tissue, and it is sent to a laboratory for analysis. If remaining tumor parts are detected in the margins, further surgery may be performed. For BCC, approximately \SI{10}{\%} of the tumors recur after the usual removal surgeries \cite{Crowson2006}. The resection margins' microscopic control can reduce the recurrence rate to \SI{1}{\%} \cite{Miller1991, Liersch2014}.

As a part of the laboratory routine, each tissue biopsy is cut into several slices in the center of the probe and close to its edges. The pathologist can only be sure that the surgery was successful if no tumor is present in the latter ones. However, this task is time-consuming and error-prone, even for skilled pathologists. Examination under the microscope is currently the most common practice. Still, pathology laboratories can use scanners to digitalize the samples and obtain Whole Slide Images (WSI) to benefit from a Computer-Aided Diagnostic system. We aim to develop an artificial intelligence method that can support pathologists in providing fast, reliable, and reproducible decisions in this context.

The first and most crucial step we do is to automatically highlight where the tumors are located with the highest probability. That also allows the pathologist to interpret the computer-based decisions better. To this end, we obtain a deep learning-based semantic segmentation of a WSI into two classes: \textit{Tumor} and \textit{Normal}, i.e., each pixel of tissue in the image is assigned to one of these labels. Following, we decide for each section, whether it contains tumors or not, based on the segmentation, and account for possible model errors.

\subsection{Related work}

Deep learning has shown great potential to address several problems in understanding, reconstructing, and reasoning about images. In particular, convolutional neural network approaches have been actively used for classification and segmentation tasks in a wide field of applications \cite{litjens201760, Chen2018, survey2020}, ranging from robot vision and understanding to the support of critical medical tasks \cite{unet2015, oktay2018attention, etmann2019deep, behrmann2017bioinf}. The nearly human-expert performance achieved in some medical imaging applications \cite{esteva2017dermatologist, gulshan2016development} has come to show the capabilities and potential of these algorithms.

Besides the success of deep learning methods in various histological imaging tasks \cite{Janowczyk2016, Iizuka2020, Kriegsmann2020}, to the best of our knowledge, there are only a few works \cite{olsen2018diagnostic} on tumor recognition in dermatopathological microscope images. We argue that the automated diagnostic of microscopic images of the human skin can be incredibly challenging due to the large variety of relevant data. Dermatological features like the structure of the extracted tissue depend on several aspects, e.g., the body part or the patient's skin type. Therefore, the pursued deep learning model has to be more robust towards such changes than models that segment histological images of inner organs. An extensive database, such as the one we use for this study,  is essential for reliable digital dermatopathology solutions.

Most recent semantic image segmentation methods are based on fully convolutional networks (FCN) \cite{long2015fully}. In this work, we use the UNet \cite{unet2015} as base architecture, which has been successfully applied in a large variety of medical image analysis applications \cite{milletari2016v, li2018h, oktay2018attention}. Recent work has also used a UNet to segment the epidermis from the skin slice \cite{Oskal2019}. While a previous study \cite{olsen2018diagnostic} obtained promising results in diagnosing a BCC subtype, we focus on the automatic detection of BCC in general, including several subtypes such as sclerodermiform, nodular, and superficial.

\section{Methods}

\subsection{Data collection and data parts}

\begin{figure}[h]
    \centering
    \input{figures/example-annotations}
    \caption{Detailed annotations used for training, validation and data balancing.}
    \label{fig:training_data}
\end{figure}
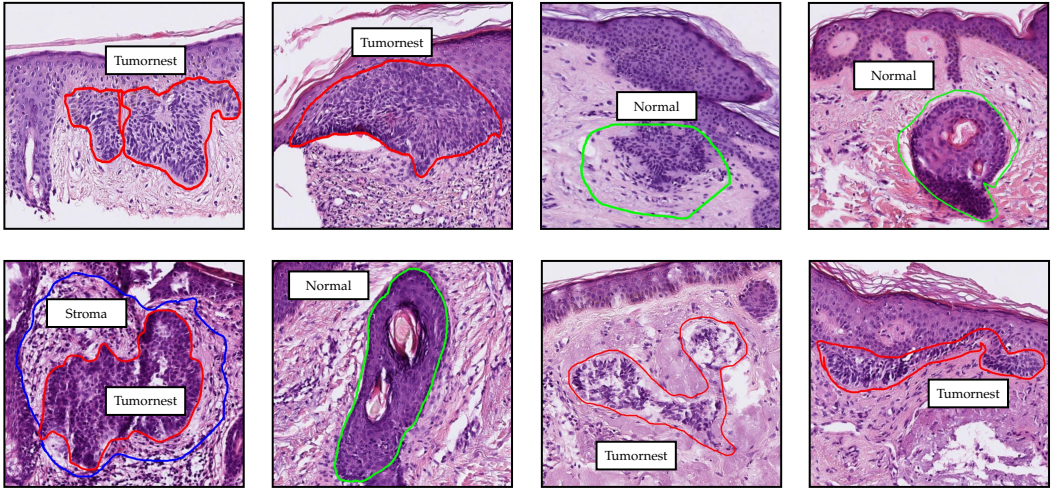

The tissue biopsy samples were prepared in the ``Dermatopathologie Duisburg'' laboratory using standard protocols. After fixation in formalin and the usual paraffin embedding, the probe was cut into slices of $\approx$ \SI{3}{\micro\metre} thickness, and standard hematoxylin and eosin (H\&E) staining was applied. The samples were digitized using a ``Hamamatsu NanoZoomer S360'' scanner with a $20$x lens. In total, this study includes $650$ WSI annotated in different manners and used for training, validation, and testing. Two clinical dermatopathologists annotated $100$ WSI, which were included in the \textit{Train} and \textit{Validation I} parts of the data, see Table~\ref{tab:datasets}. The annotations contain different interest regions of the tissue: \textit{Tumornest}, indicates where tumor islands are; \textit{Stroma}, highlights supporting tissue around the tumor islands; and \textit{Normal}, contains parts of the tissue that look similar to tumors, such as hair follicles, actinic keratosis, or cysts. However, only the \textit{Tumornest} annotations were done exhaustively, whereas the others were annotated only in some cases. For that reason, we only use \textit{Tumornest} annotations for training, and leave the other types for data balancing. Some annotation examples are shown in Figure~\ref{fig:training_data}.

\begin{table}[h]
    \centering
    \begin{tabular}{l|l|c|l|l}
    \toprule
         Part & Slides & Detailed annotations & Tumor sections & Normal sections\\
         \toprule
         Training & $85$ & \checkmark & $188$ & $209$ \\
         Validation I & $15$ & \checkmark & $53$ & $31$ \\
         Validation II & $229$ & & $392$ & $608$ \\
         Test & $321$ & & $1119$ & $843$ \\
         \bottomrule
    \end{tabular}
    \caption{Distribution of the slides, sections and type of annotations. All slides contain section-wise annotations with \textit{Tumor} or \textit{Normal} labels.}
    \label{tab:datasets}
\end{table}

\begin{figure}[ht]
    \centering
    \input{figures/example_slides}
    \caption{Example ground-truth labels (\textit{Tumor} in red and \textit{Normal} in green) for two WSI from the \textit{Test} data.}
    \label{fig:test_data}
\end{figure}
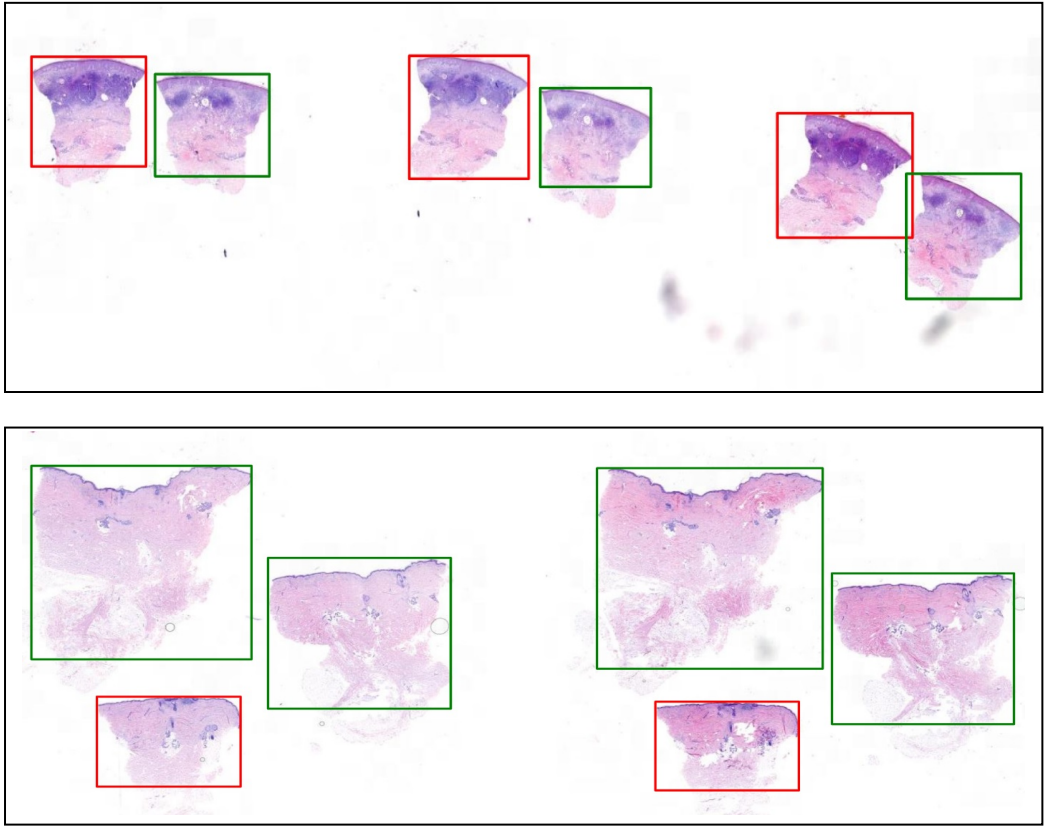

Due to the huge number of WSI, it was not feasible to create detailed annotations for all the slides. Two parts of the data, namely \textit{Validation II} and \textit{Test} only contain section-wise labels. That means that for each section of tissue in the WSI, there is a label that indicates whether there is some tumor inside or not, see Figure~\ref{fig:test_data}. To speed up the section-wise annotation process, we first generated bounding boxes and assigned the label \textit{Normal} or \textit{Tumor} resulting from one of our models. The pathologists then efficiently corrected these labels by just changing those which were wrong, see Section~\ref{sec:blind_study}. Table~\ref{tab:datasets} contains the amount of WSI for each part of the data and the number of sections.

\subsection{Model architecture}

The neural network designs we use in this work follow the UNet architecture \cite{unet2015}, a fully convolutional encoder-decoder network with skip-connections between the encoder blocks and their symmetric decoder blocks. We use two different encoders and a standard decoder, similar to the original one \cite{unet2015} with only minor changes. The input to the network is a patch $x \in \R^{512\times512\times3}$, and the output is a segmentation map $\Phi(x) \in [0,1]^{512\times512\times2}$. The segmentation map contains two matrices of size $512\times512$, which correspond to the predicted pixel-wise probabilities for each of the two classes: \textit{Tumor} and \textit{Normal}. Since we use the Softmax activation to compute the segmentation map, this can be seen as only one matrix of size $512\times512$ with the \textit{Tumor} probabilities.

\subsubsection{Encoder}

The first encoder follows nearly the same design as the original one \cite{unet2015} with $5$ blocks, which successively down-sample the spatial resolution to increasingly catch higher-level features. Each of the blocks doubles the number of feature channels and halves the spatial resolution. It uses two $3 \times 3$ convolutions, where the first one uses a stride of $2$. The second encoder is exactly the convolutional backbone of a ResNet34 \cite{he2016deep}, as can be observed in Figure~\ref{fig:architecture}, and also contains $5$ blocks.

Both encoders contain an initial block with a $7 \times 7$ convolution with 64 filters and a stride of $2$, which decreases the input resolution by half. That allowed us to effectively enlarge the patches' size ($512\times512$) to incorporate a larger context without a significant increase in computation and memory consumption.

\subsubsection{Decoder}

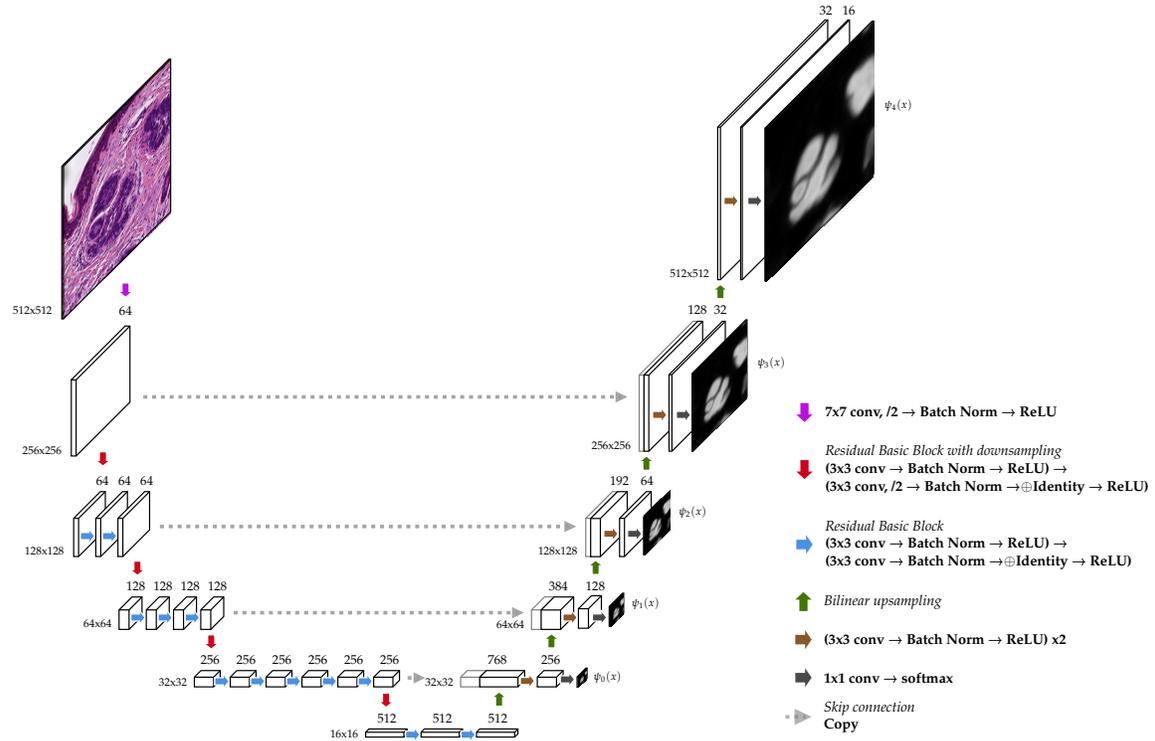
\begin{figure}[ht]
    \centering
    \scalebox{0.6}{\input{figures/architecture}}
    \caption{UNet architecture with a ResNet-34 encoder. The output of the additional $1 \times 1$ convolution after Softmax is shown next to each decoder block.}
    \label{fig:architecture}
\end{figure}

The decoder contains an expanding path that seeks to build a segmentation map from the encoded features, see Figure~\ref{fig:architecture}. It has the same number of blocks as the encoder. Each decoding block duplicates the spatial resolution while halving the number of feature channels. It performs a bilinear up-sampling and concatenates the result with the output of its symmetric block in the encoder ({\it skip-connection}). Next, it applies two $3 \times 3$ convolutions with the same number of filters. Unlike the original decoder proposed in \cite{unet2015}, it has one extra block that does not obtain any \textit{skip-connection}. The output of the last decoder block is passed to a $1 \times 1$ convolution that computes the final \textit{score-map}.

Additionally, at each block of the decoder, we added an extra $1 \times 1$ convolution, i.e., a linear combination of the block's final feature maps, to produce an intermediate \textit{score-map}. We used these maps in two settings: \textbf{a)} deep supervision and \textbf{b)} to merge them through a linear combination to produce the final output. Except for the $1 \times 1$ convolutions, we always use a batch- normalization layer \cite{ioffe2015batch} and a ReLU activation function right after each convolution in the whole network. Furthermore, we always use the corresponding padding to keep the same spatial size unchanged by the convolution's kernel.

\subsection{Model training}

For training the models, we used the \textit{Train} set, which contains $85$ annotated WSI. We extracted small patches of size $512\times512$ at $10$x level of magnification, on which we performed the semantic segmentation. In almost all slides, the tumor-free tissue is dominant; therefore, it was necessary to balance the training data to avoid biases and improve the performance. We did this based on the amounts of pixels belonging to the $3$ types of annotations: \textit{Tumornest} ($T$), \textit{Stroma} ($S$), and \textit{Normal} ($N$). The re-sampling was done as depicted in Table~\ref{table:resampling}. That means that some of the patches were used several times during a training epoch. That is the case for patches that contain tumor, are close to tumor areas (\textit{Stroma}), or contain normal regions which look similar to tumors (\textit{Normal}). Additionally, we did extra over-sampling for patches with high tumor density.

During training, all patches were extensively augmented using: random rotations, scaling, smoothing, color variations, and elastic deformations to increase the variety of the data effectively. Additionally, we used the Focal-Loss with $\gamma=2.0$ \cite{Lin2018}, which has a similar effect to down-weighting the easy examples, to make their contribution to the total loss smaller. We did not include any re-sampling or augmentation for the validation dataset.

\begin{table}[ht]
    \centering
    \begin{tabular}{l|r|r}
    \toprule
     & Before & After \\
    \midrule
    $T < \SI{0.05}{\%}$ & 175,771 & 175,771 \\
    $T \geq \SI{0.05}{\%}$ & 9,537 & 30,000 \\
    $T \geq \SI{10}{\%}$ & 5,528 & 10,000 \\
    $S \geq \SI{0.05}{\%}$ & 9,096 & 20,000 \\
    $N \geq \SI{0.05}{\%}$ & 9,458 & 20,000 \\
    \midrule
    Total patches & 209,390 & \textbf{ 255,771} \\
    \midrule
    Pixel unbalance & 78.48 & \textbf{ 16.81} \\
    \bottomrule
    \end{tabular}
    \caption{Patches distribution and pixel-wise unbalance before and after re-sampling the data. The pixel unbalance is computed as the ratio between the amounts of tumor-free and tumor pixels.}
    \label{table:resampling}
\end{table}

We trained the models using a maximum of $40$ epochs, a batch-size of $64$, $lr=5\cdot 10^{-4}$, and a scheduler to multiply the learning rate by $0.8$ every $5$ epochs. Optimization was performed with the Adam method \cite{adam} and all computations were done on $4$ NVIDIA GeForce GTX 1080 Ti.

\subsubsection{Deep supervision}

The deep supervision strategy consists of forcing the decoder blocks' outputs to yield a meaningful segmentation map. We compare each of the decoder blocks' output with the corresponding down-sampled version of the target segmentation map and add the discrepancy to the total loss. This technique was originally introduced in \cite{Lee2015}, for obtaining transparency and robustness of the features extracted in the middle of the network and helping address the vanishing gradient problem. In this case, it allows gradient information to flow back directly from the loss to every block of the decoder. Some recent works  \cite{Zhu2017, Zongwei2018} used a similar idea for training a UNet.

Let $x \in \R^{512\times512\times3}$ be an input patch and its corresponding target segmentation map ${y\in \{0,1\}^{512\times512\times2}}$. We define $\psi_\ell(x)$ as the output of the $\ell$-block of the decoder (after Softmax), see Figure~\ref{fig:architecture}. The contribution to the loss function from this single data point $(x,\,y)$ is then defined as
\begin{equation}
    loss(x,\,y) = \sum_{\ell=0}^{k-1}  fl\left(\psi_\ell(x),\, \Pi_\ell(y)\right),
\end{equation}
where $k=5$ is the number of decoder blocks, $\Pi_\ell:\, \{0,1\}^{512\times512\times2} \to \{0,1\}^{32 \cdot 2^\ell \times 32 \cdot 2^\ell \times 2}$ is a down-sampling operator, and $fl$ is the Focal-Loss. We also tried using different weights for each of the scales, but we did not observe any improvement. This strategy does not involve any change on the previously described architecture; it only needs the $1\times1$ convolution and the Softmax at the end of each block of the decoder. The final output is given by $\Phi(x)=\psi_{k-1}(x)$.

\subsubsection{Linear merge}

The second strategy merges the decoder outputs through a linear combination. We add a linear layer with weights $w\in \R^k$ to the architecture that computes the final output, i.e.,
\begin{equation}
    \Phi(x) = \text{Softmax}\left(\sum_{\ell=0}^{k-1} w_{\ell} \cdot \Gamma_\ell(\hat{\psi}_\ell(x))\right),
\end{equation}
where $\hat{\psi}_\ell(x)$ is the score-map computed at the $\ell$-block of the decoder, i.e., the same as $\psi_\ell(x)$ but without the Softmax activation, and $\Gamma_l:\, \R^{32 \cdot 2^\ell \times 32 \cdot 2^\ell \times 2} \to \R^{512\times512\times2}$ is a bilinear up-sampling operator. In this case, the Softmax is only applied after the linear combination. The weights $w$ are trained together with the other parameters of the model. Note that the standard model without this strategy is a special case, since the model could learn to assign $w_{k-1}=1.0$ and $w_\ell = 0$ for $0 \leq \ell < k$.

\subsection{Section-wise classification}

The final classification task for each section of the WSI (see Figure \ref{fig:test_data}) is to decide whether it contains tumor or not. First of all, we use the trained model to generate a heatmap by combining several patches' predictions and highlighting the tissue's parts with the highest probabilities to contain tumor. We extract the patches so that they cover the whole tissue area and have at least $256$ pixels (\SI{50}{\%}) overlapping at every border. Following, we apply a prediction threshold to obtain a binary mask and find connected regions. To account for some possible model errors and reduce the false positive rate, we filter out regions below a certain area threshold. If any predicted tumor area is left in the section, it is classified as \textit{Tumor}; otherwise, it is classified as \textit{Normal}. The prediction and area thresholds are selected for each model independently as part of the model selection.

\subsection{Model selection}

We trained our implementation of the original UNet and three other variants using a ResNet34 encoder. The first alternative does not employ any decoder strategy (ResNet34-UNet), the second one uses the deep supervision strategy (ResNet34-UNet + DS), and the third one uses the linear combination of the decoder outputs (ResNet34-UNet + Linear). During training, we evaluated the models on the \textit{Validation I} part of the data. We used the Intersection over Union (IoU) metric to select the models from the best five training epochs for each setting. Afterward, we evaluated these models on the section-wise classification task in the \textit{Validation I} and \textit{Validation II} parts of the data to select the best model together with the prediction and area thresholds. To this end, we used a grid-search and the $F_\beta$ score
\begin{equation}
    F_\beta = \frac{(1+\beta^2) \cdot \text{precision} \cdot \text{recall}}{\beta^2 \cdot \text{precision} + \text{recall}},
\end{equation}
with $\beta=1.5$, to give higher importance to the recall/sensitivity.

\begin{figure}
    \centering
    \scalebox{0.95}{\input{figures/predictions}}
    \caption{Example of inference using the analyzed settings on patches from the \textit{Validation I} part of the data.}
    \label{fig:predictions}
\end{figure}
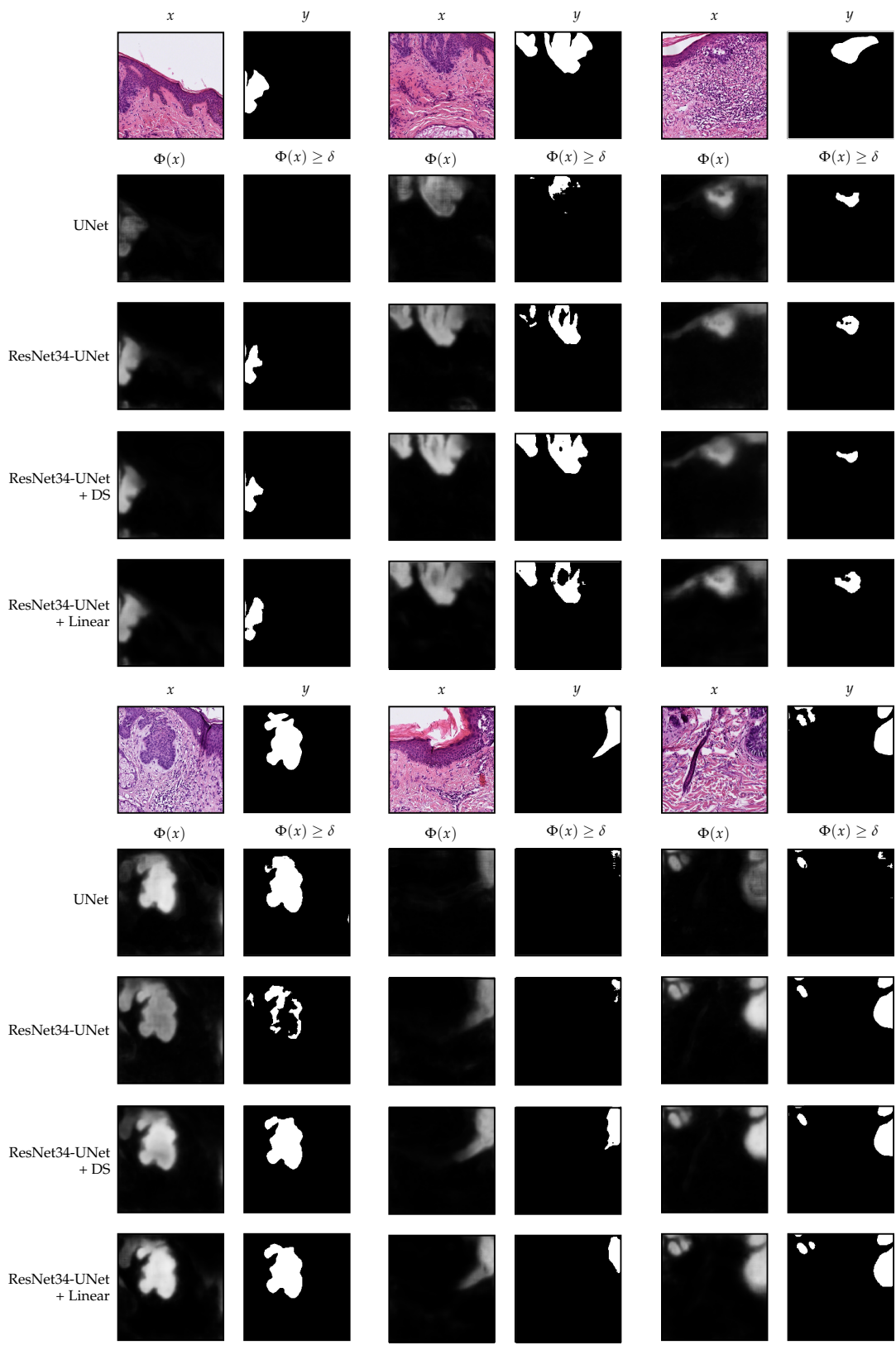

The performance of the best models and the selected thresholds are shown in Table~\ref{tab:validation}. Additionally, Figure~\ref{fig:predictions} presents the outputs of the models and the resulting masks after applying the corresponding thresholds for some patches from the \textit{Validation I} data. 

\begin{table}[h]
    \centering
    \begin{tabular}{l|c|c|c|c|c|c}
        \toprule
        Setting &
        \begin{tabular}{@{}c@{}}Prediction \\ Threshold\end{tabular} &
        \begin{tabular}{@{}c@{}}Tumor-Area \\ Threshold (\SI{}{\micro\meter^2})\end{tabular} &
        Accuracy & Sensitivity & Specificity & $F_{\beta}$\\
        \midrule
        UNet & 0.45 & 8960 & 0.985 & {\bfseries 1.000} & 0.975 & 0.989\\
        ResNet34-UNet & 0.60 & 3840 & 0.993 & 0.998 & 0.989 & 0.994\\
        ResNet34-UNet + DS & 0.60 & 5120 & 0.994 & 0.996 & 0.992 & 0.993\\
        ResNet34-UNet + Linear & 0.65 & 2560 & {\bfseries 0.996} & {\bfseries 1.000} & {\bfseries 0.994} & {\bfseries 0.997}\\
        \bottomrule
    \end{tabular}
    \caption{Results on the section-wise classification task for all sections from the \textit{Validation~I} and \textit{Validation~II} parts of the data. The values correspond to the best models and selected thresholds.}
    \label{tab:validation}
\end{table}

\section{Results}

\begin{figure}[h]
    \centering
    \scalebox{0.9}{\input{figures/misclassifications_test}}
    \caption{Number of wrongly classified sections on the {\it Test} part of the data. There are $1962$ sections in total.}
    \label{fig:misclassifications-test}
\end{figure}
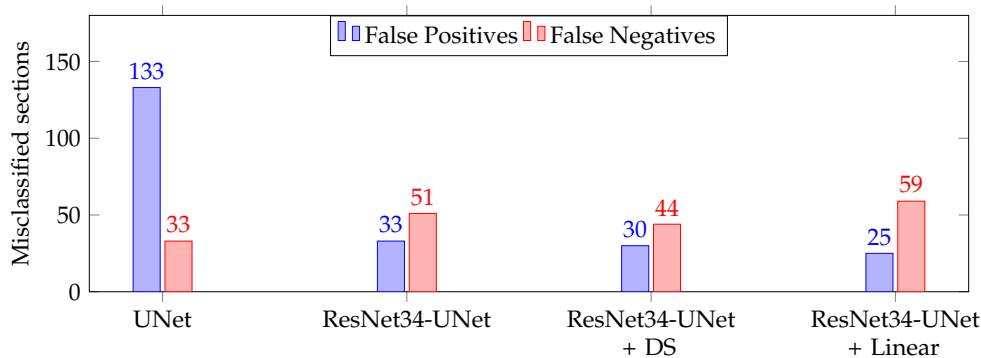

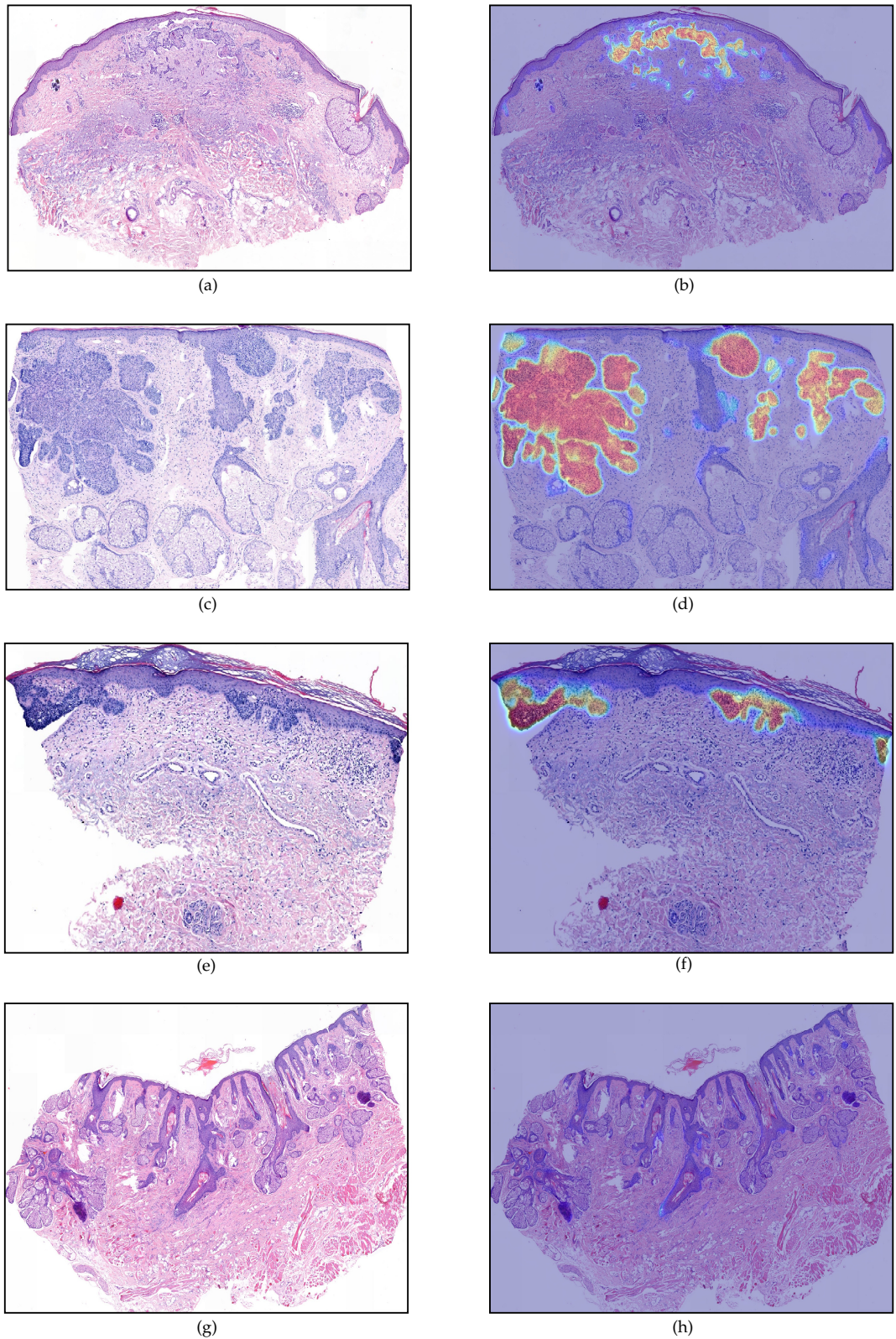
\begin{figure}
    \centering
    \scalebox{0.95}{\input{figures/heatmaps}}
    \caption{Generated heatmaps (ResNet34-UNet + DS) for sections from the \textit{Test} part of the data. The images show a variety of BCC subtypes that were part of the dataset a) sclerodermiform BCC, c) nodular BCC, e) superficial BCC, g) no tumor. As heatmap b) suggests, the exact segmentation of sclerodermiform BCC can be quite challenging.}
    \label{fig:heatmaps}
\end{figure}

Finally, we evaluated the selected models from the validation phase on the \textit{Test} data. The results are depicted in Table \ref{tab:test-results} and Figure~\ref{fig:misclassifications-test} shows the number of wrongly classified sections by each model. The ResNet34-UNet + DS achieved the best results, obtaining more than \SI{96}{\%} overall accuracy, sensitivity, and specificity. It wrongly classified only $74$ ($30$ FP and $44$ FN) out of $1962$ sections. Figure~\ref{fig:heatmaps} shows some correctly classified sections for different BCC subtypes and the corresponding heatmaps generated with this model.

\begin{table}[h]
    \centering
    \begin{tabular}{l|c|c|c|c}
        \toprule
        Setting & Accuracy & Sensitivity & Specificity & $F_{\beta}$\\
        \midrule
        UNet & 0.915 & {\bfseries0.971} & 0.842 & 0.945\\
        ResNet34-UNet & 0.957 & 0.954 & 0.961 & 0.959\\
        ResNet34-UNet + DS & {\bfseries 0.962} &  0.961 & 0.964 & {\bfseries 0.964}\\
        ResNet34-UNet + Linear & 0.957 & 0.947 & {\bfseries 0.970} & 0.956\\
        \bottomrule
    \end{tabular}
    \caption{Results on the section-wise classification task for all sections from the \textit{Test} part of the data.}
    \label{tab:test-results}
\end{table}

\begin{figure}
    \centering
    \scalebox{0.95}{\input{figures/false_positive}}
    \caption{Generated heatmap (ResNet34-UNet + DS) for part of a section from the \textit{Test} data which was wrongly classified as a section with tumor (false positive). In this case, the model identified a hair-follicle as tumor, as it can be seen on the highlighted area in (b).}
    \label{fig:false_positive}
\end{figure}
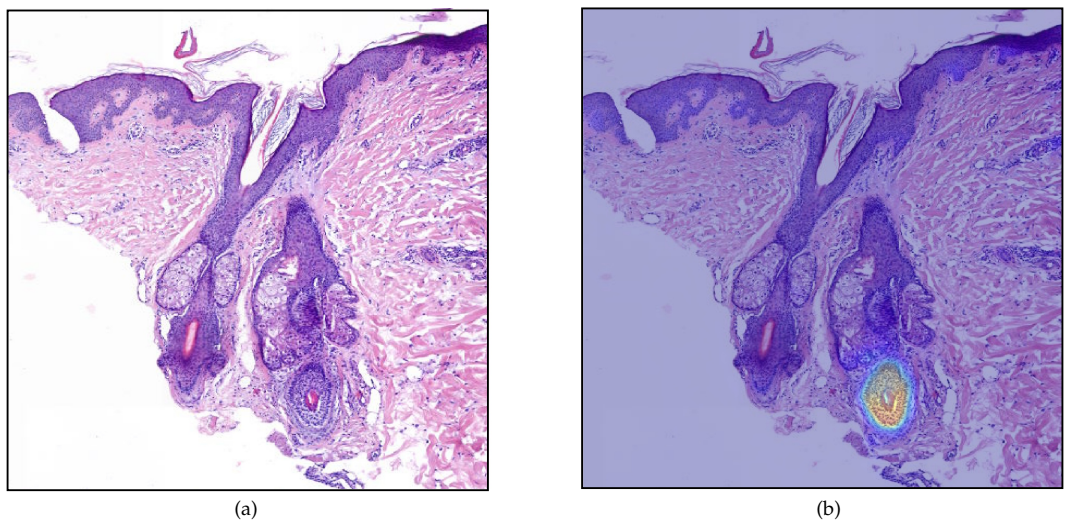

In most false negative cases, the model detects the tumor but without enough confidence. Due to the selected prediction and area thresholds these sections are then classified as \textit{Normal}. On the other hand, false positives are often due to a hair follicle or other skin structure that is identified as a tumor, see for example Figure~\ref{fig:false_positive}.

The ResNet34-UNet + DS was the second-best in the model selection phase. We argue that the slightly different results on the test set are due to the fact that the validation data is biased. Initially, the \textit{Validation II} part was larger, but we selected challenging slides, based on the pathologists' feedback, which were then fully annotated and included in the \textit{Training} part. All in all, the classification error was reduced from \SI{8.5}{\%} (baseline UNet) to \SI{3.8}{\%}. The baseline UNet, which uses the standard encoder, has excellent sensitivity but quite a low specificity. In contrast, the models with the ResNet34 encoder are more balanced and exhibit better performance.

\subsection{Discussion and interpretability}

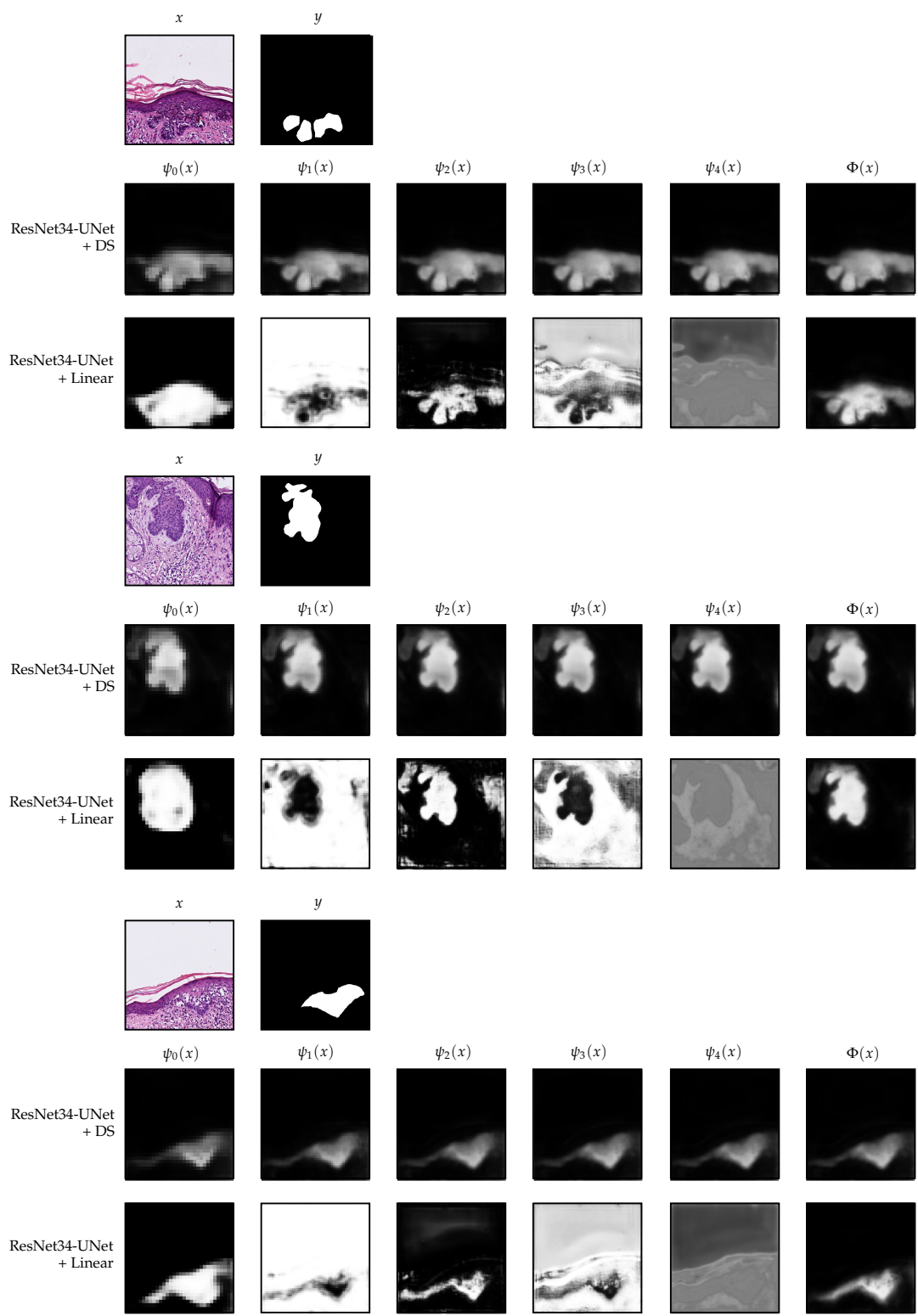
\begin{figure}
    \centering
    \scalebox{0.95}{\input{figures/decoder_outputs}}
    \caption{Decoder outputs for each block of the decoder in the deep supervision and linear merge settings. The patches belong to the \textit{Validation I} part of the data. For the linear merge strategy, the segmentation maps shown in the figure are after applying the Softmax operation, which we do in this case only for visualization purposes.}
    \label{fig:decoder_outputs}
\end{figure}

We now analyze in more detail the models trained with the two additional strategies, namely ResNet34-UNet + DS and ResNet34-UNet + Linear. In Figure~\ref{fig:decoder_outputs}, we show the decoder outputs for some patches from the \textit{Validation I} data using these models. 

In the deep supervision case, one can observe that, indeed, it is possible to guide the UNet to produce a meaningful segmentation already from the first block ($\psi_0$) of the decoder, see Figure~\ref{fig:decoder_outputs}. The difference between $\psi_0(x)$ and $\psi_4(x)$ is only the resolution and the number of details on the borders. We observed that already in $\psi_0(x)$, the model knows where the tumor is if there is any. Therefore, we separately evaluated the performance of $\psi_0$ and the other decoder blocks on the final classification task, using the same prediction and area threshold selected for the original model. The results are nearly the same as when using the whole model, see Table~\ref{tab:test-results-blocks}; in the case of $\psi_0$, the difference is only $5$ sections, whereas in $\psi_1$ it is only $1$. Those sections were wrongly classified because the predicted tumor area was right at the limit. 

Moreover, performing inference using the encoder followed by only the decoder's first block $\psi_0$ is much faster and boosts the heatmap generation process's speed. The original model takes approximately \SI{30}{s} per WSI (several sections) on average on an NVIDIA GeForce GTX 1080 Ti. In contrast, the reduced one needs approximately only \SI{20}{s}, which represents a reduction of $\SI{33}{\%}$ of the time.

\begin{table}[H]
    \centering
    \begin{tabular}{l|c|c|c|c}
        \toprule
        Block & Accuracy & Sensitivity & Specificity & $F_{\beta}$\\
        \midrule
        $\psi_0$ & 0.9597 & 0.959 & 0.961 & 0.962\\
        $\psi_1$ & 0.9617 & 0.961 & 0.963 & 0.964\\
        $\psi_2$ & 0.9617 & 0.961 & 0.963 & 0.964\\
        $\psi_3$ & {\bfseries 0.9622} &  {\bfseries 0.961} & {\bfseries 0.964} & {\bfseries 0.964}\\
        $\psi_4$ & {\bfseries 0.9622} &  {\bfseries 0.961} & {\bfseries 0.964} & {\bfseries 0.964}\\
        \bottomrule
    \end{tabular}
    \caption{Results on the test data for the ResNet34-UNet + DS, where the heatmaps are generated using the output of each decoder block.}
    \label{tab:test-results-blocks}
\end{table}

On the other hand, the setting that uses a linear merge of the decoder blocks' outputs has a fascinating behavior. In this case, the model has more freedom since we do not guide the decoder blocks to output meaningful segmentations. We only include the final linear combination in the loss function. In Figure \ref{fig:decoder_outputs}, one can observe that in $\psi_0(x)$, the model identifies a rough approximation of the tumor's location, which is much larger than the final result. Following, in the subsequent blocks, it creates more details. The learned weights assigned to each block were $w = [0.1590,\, -0.1645,\,  0.1603,\, -0.2963,\,0.0036]$. Even though the observed behavior makes sense and offers some interpretability, it seems that, at least for the final classification task, it does not bring any advantage. As we have seen, only one block of the decoder seems to be good enough. The linear merge strategy might be more effective if the aim is to obtain very exact borders.

\subsection{Blind study}
\label{sec:blind_study}
The results presented in this work were obtained as part of a blind study. At the moment of training the section-wise annotations for the \textit{Test} data did not exist yet. We created an initial classification based on the ResNet34-UNet model, and the pathologists checked them and corrected the wrong ones. That allowed us to obtain the section-wise annotations of the \textit{Test} data in a very time-efficient manner. The process was done using our experimental online platform \footnote{\url{https://digipath-viewer.math.uni-bremen.de/}}, which was created for reviewing and visualizing WSI, annotations, and model predictions \footnote{This platform was also used to obtain some of the detailed annotations for the \textit{Training} and \textit{Validation I} data.}. Finally, after the wrong labels were corrected, we computed the final scores for all our models.
%

\section{Conclusions}

In this work, we used a UNet architecture with two different encoders and several training strategies for performing automatic detection of BCC on skin histological images. Training the network with deep supervision was the decisive factor for improving the final performance. After trained with this strategy, each decoder's block focused on obtaining a segmentation with more details than the previous one but did not add or remove any tumor content. We found out that the decoder's first block is enough for obtaining nearly the best results on the classification task. That implies a substantial speed improvement on the forward pass of the network and, therefore, on the heatmap generation process (\SI{33}{\%} time reduction). We performed the final evaluation on a rather large \textit{Test} dataset compared to the \textit{Training} part of the data. Still, the best model obtained a \SI{96.2}{\%} accuracy and similar sensitivity and specificity on the section-wise classification task. These results are very promising and show the potential of deep learning methods to assist dermapatopathological assessment of BCC.

\vspace{6pt} 





\acknowledgments{Jean Le'Clerc Arrastia and Daniel Otero Baguer acknowledge the support by the Deutsche Forschungsgemeinschaft (DFG) within the framework of GRK 2224/1 ``$\pi^3$: Parameter Identification -- Analysis, Algorithms, Applications''.}

\conflictsofinterest{The authors declare no conflict of interest.The funders had no role in the design of the study; in the collection, analyses, or interpretation of data; in the writing of the manuscript, or in the decision to publish the results.} 






\reftitle{References}


\externalbibliography{yes}
\bibliography{references.bib}



\end{document}

%% file: figures/example-annotations.tex
\tikzset{every picture/.style={line width=0.75pt}} 

\begin{tikzpicture}[x=0.75pt,y=0.75pt,yscale=-1,xscale=1]

\draw (479.93,65.86) node  {\includegraphics[width=90.53pt,height=85.55pt]{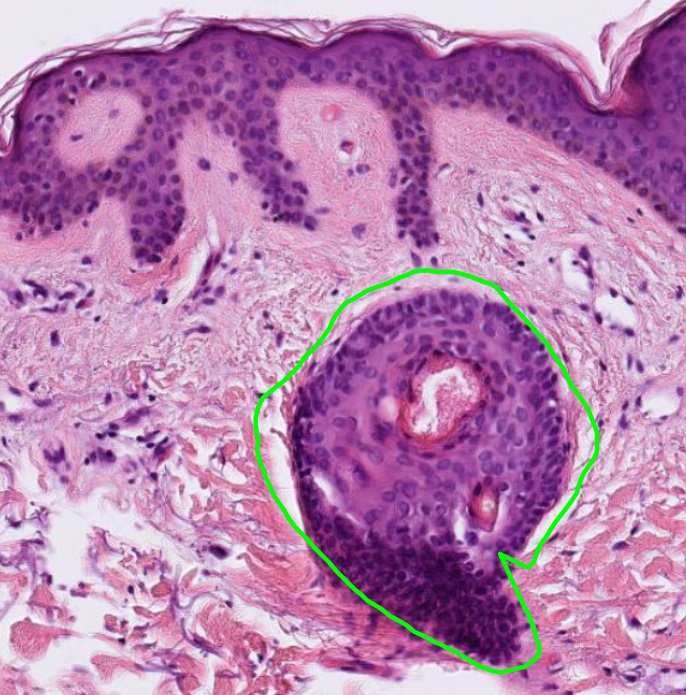}};
\draw (345.26,196.08) node  {\includegraphics[width=90.53pt,height=85.55pt]{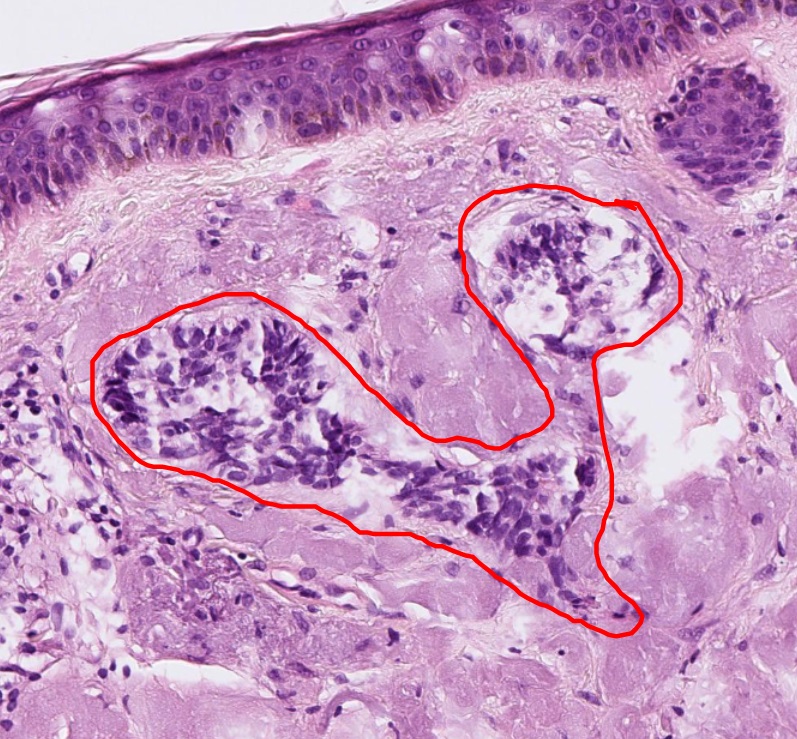}};
\draw (209.52,196.08) node  {\includegraphics[width=90.53pt,height=85.55pt]{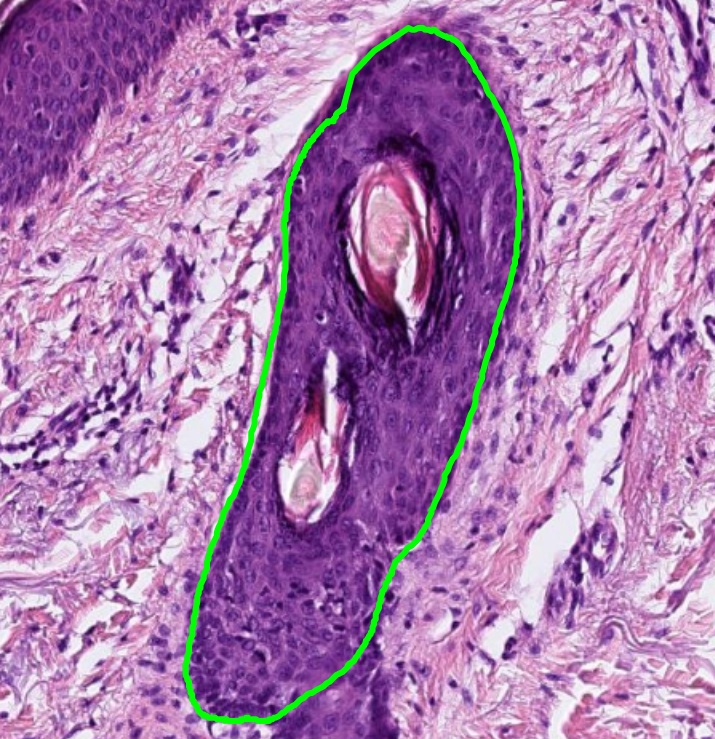}};
\draw (74.19,196.08) node  {\includegraphics[width=90.53pt,height=85.55pt]{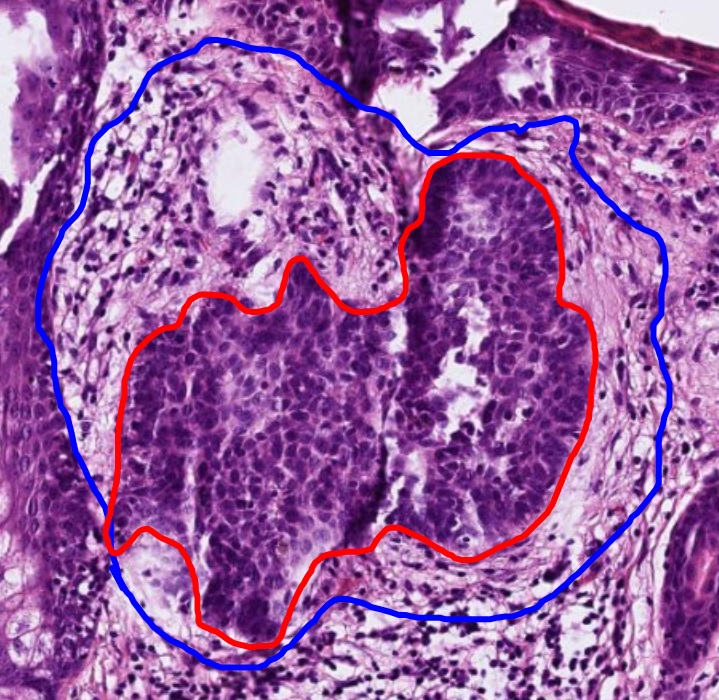}};
\draw (74.19,65.81) node  {\includegraphics[width=90.53pt,height=85.55pt]{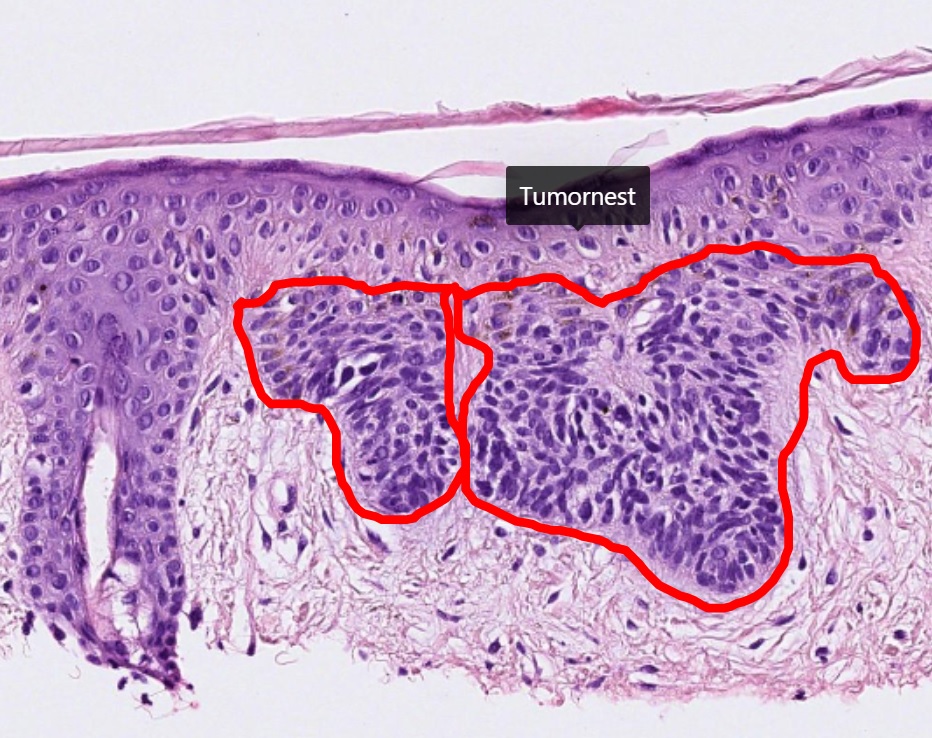}};
\draw (209.52,65.81) node  {\includegraphics[width=90.53pt,height=85.55pt]{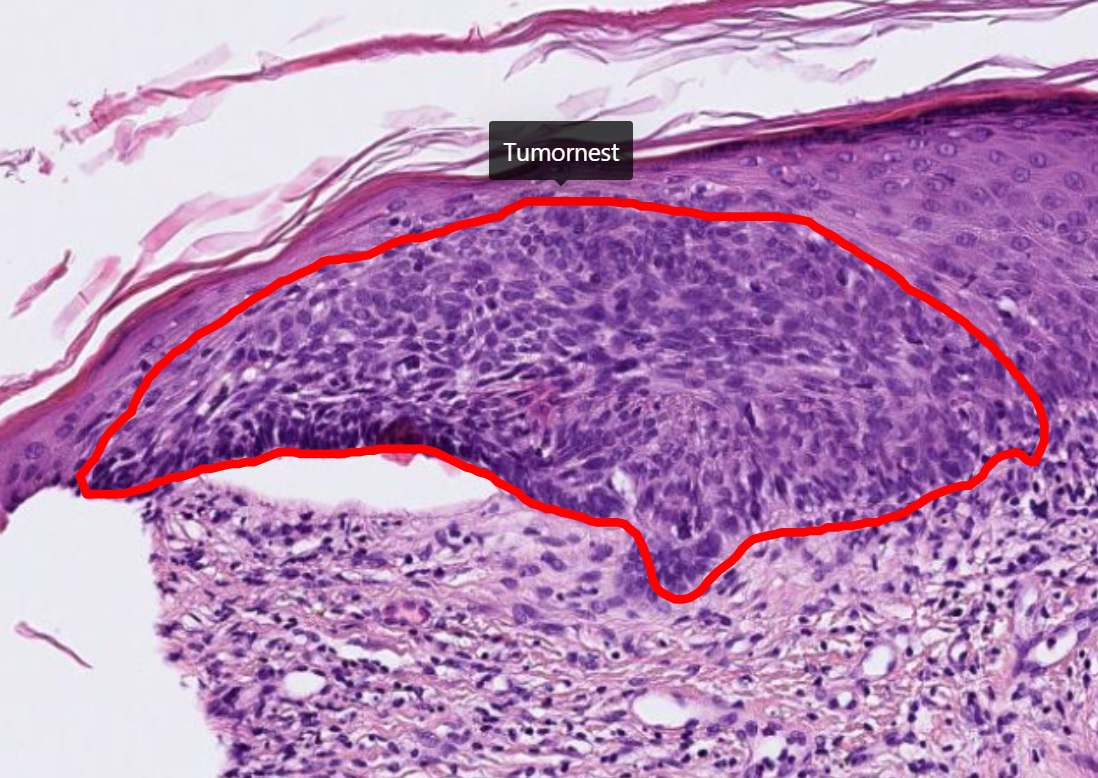}};
\draw (344.5,65.86) node  {\includegraphics[width=90.02pt,height=85.55pt]{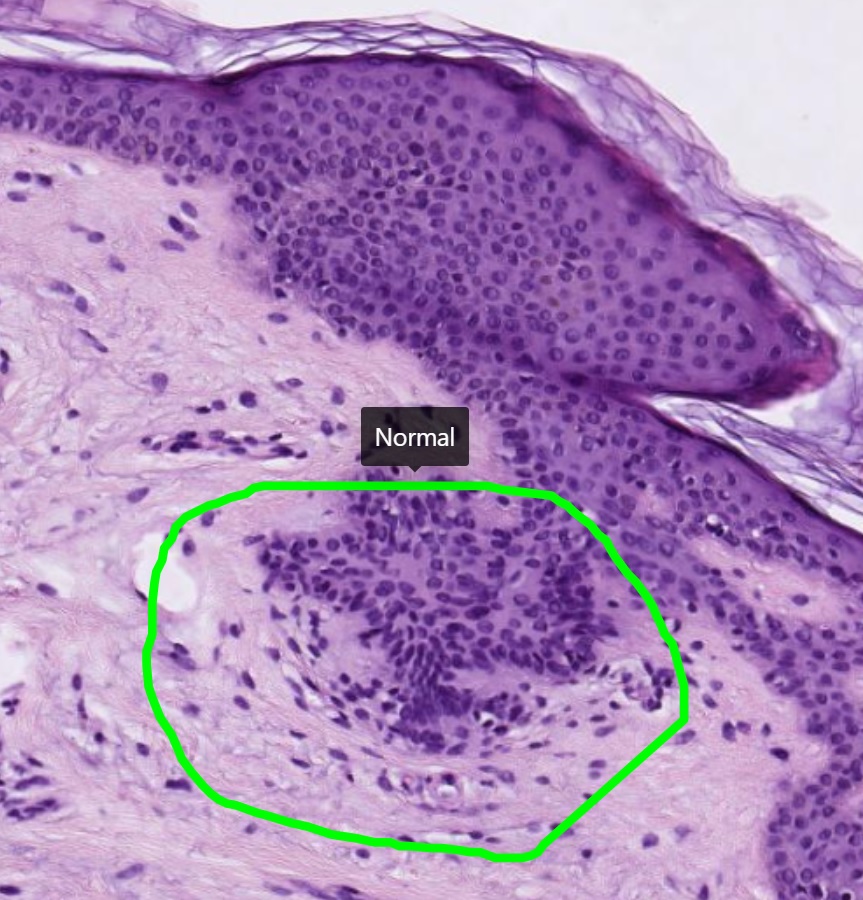}};
\draw   (13.84,8.78) -- (134.55,8.78) -- (134.55,122.84) -- (13.84,122.84) -- cycle ;
\draw   (149.16,8.78) -- (269.87,8.78) -- (269.87,122.84) -- (149.16,122.84) -- cycle ;
\draw   (284.49,8.83) -- (405.2,8.83) -- (405.2,122.89) -- (284.49,122.89) -- cycle ;
\draw   (13.84,139.05) -- (134.55,139.05) -- (134.55,253.11) -- (13.84,253.11) -- cycle ;
\draw   (149.16,139.05) -- (269.87,139.05) -- (269.87,253.11) -- (149.16,253.11) -- cycle ;
\draw   (284.9,139.05) -- (405.61,139.05) -- (405.61,253.11) -- (284.9,253.11) -- cycle ;
\draw (480.34,196.08) node  {\includegraphics[width=90.53pt,height=85.55pt]{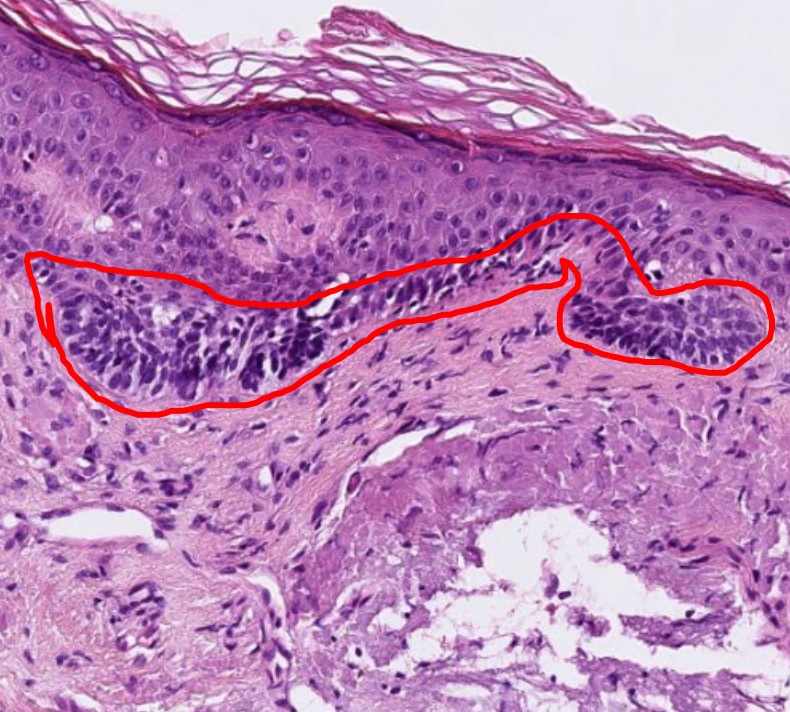}};
\draw   (419.57,8.83) -- (540.28,8.83) -- (540.28,122.89) -- (419.57,122.89) -- cycle ;
\draw   (419.99,139.05) -- (540.7,139.05) -- (540.7,253.11) -- (419.99,253.11) -- cycle ;

\draw  [fill={rgb, 255:red, 255; green, 255; blue, 255 }  ,fill opacity=1 ]  (65.04,30.76) -- (105.04,30.76) -- (105.04,44.76) -- (65.04,44.76) -- cycle  ;
\draw (85.04,37.76) node  [font=\tiny] [align=left] {\begin{minipage}[lt]{24.208798680217544pt}\setlength\topsep{0pt}
\begin{center}
Tumornest
\end{center}

\end{minipage}};
\draw  [fill={rgb, 255:red, 255; green, 255; blue, 255 }  ,fill opacity=1 ]  (189.38,21.63) -- (229.38,21.63) -- (229.38,35.63) -- (189.38,35.63) -- cycle  ;
\draw (209.38,28.63) node  [font=\tiny] [align=left] {\begin{minipage}[lt]{24.208798680217544pt}\setlength\topsep{0pt}
\begin{center}
Tumornest
\end{center}

\end{minipage}};
\draw  [fill={rgb, 255:red, 255; green, 255; blue, 255 }  ,fill opacity=1 ]  (322.87,53.84) -- (362.87,53.84) -- (362.87,67.84) -- (322.87,67.84) -- cycle  ;
\draw (342.87,60.84) node  [font=\tiny] [align=left] {\begin{minipage}[lt]{24.208798680217622pt}\setlength\topsep{0pt}
\begin{center}
Normal
\end{center}

\end{minipage}};
\draw  [fill={rgb, 255:red, 255; green, 255; blue, 255 }  ,fill opacity=1 ]  (35.34,158.38) -- (75.34,158.38) -- (75.34,172.38) -- (35.34,172.38) -- cycle  ;
\draw (55.34,165.38) node  [font=\tiny] [align=left] {\begin{minipage}[lt]{24.208798680217562pt}\setlength\topsep{0pt}
\begin{center}
Stroma
\end{center}

\end{minipage}};
\draw  [fill={rgb, 255:red, 255; green, 255; blue, 255 }  ,fill opacity=1 ]  (65.04,202.77) -- (105.04,202.77) -- (105.04,216.77) -- (65.04,216.77) -- cycle  ;
\draw (85.04,209.77) node  [font=\tiny] [align=left] {\begin{minipage}[lt]{24.208798680217544pt}\setlength\topsep{0pt}
\begin{center}
Tumornest
\end{center}

\end{minipage}};
\draw  [fill={rgb, 255:red, 255; green, 255; blue, 255 }  ,fill opacity=1 ]  (157.43,144.36) -- (197.43,144.36) -- (197.43,158.36) -- (157.43,158.36) -- cycle  ;
\draw (177.43,151.36) node  [font=\tiny] [align=left] {\begin{minipage}[lt]{24.208798680217775pt}\setlength\topsep{0pt}
\begin{center}
Normal
\end{center}

\end{minipage}};
\draw  [fill={rgb, 255:red, 255; green, 255; blue, 255 }  ,fill opacity=1 ]  (312.48,230.31) -- (352.48,230.31) -- (352.48,244.31) -- (312.48,244.31) -- cycle  ;
\draw (332.48,237.31) node  [font=\tiny] [align=left] {\begin{minipage}[lt]{24.208798680217544pt}\setlength\topsep{0pt}
\begin{center}
Tumornest
\end{center}

\end{minipage}};
\draw  [fill={rgb, 255:red, 255; green, 255; blue, 255 }  ,fill opacity=1 ]  (442.44,38.39) -- (482.44,38.39) -- (482.44,52.39) -- (442.44,52.39) -- cycle  ;
\draw (462.44,45.39) node  [font=\tiny] [align=left] {\begin{minipage}[lt]{24.208798680217697pt}\setlength\topsep{0pt}
\begin{center}
Normal
\end{center}

\end{minipage}};
\draw  [fill={rgb, 255:red, 255; green, 255; blue, 255 }  ,fill opacity=1 ]  (480,198.94) -- (520,198.94) -- (520,212.94) -- (480,212.94) -- cycle  ;
\draw (500,205.94) node  [font=\tiny] [align=left] {\begin{minipage}[lt]{24.208798680217697pt}\setlength\topsep{0pt}
\begin{center}
Tumornest
\end{center}

\end{minipage}};

\end{tikzpicture}

%% file: figures/example_slides.tex
\tikzset{every picture/.style={line width=0.75pt}} 

\begin{tikzpicture}[x=0.75pt,y=0.75pt,yscale=-1,xscale=1]

\draw (277.59,608.52) node  {\includegraphics[width=378.93pt,height=145.07pt]{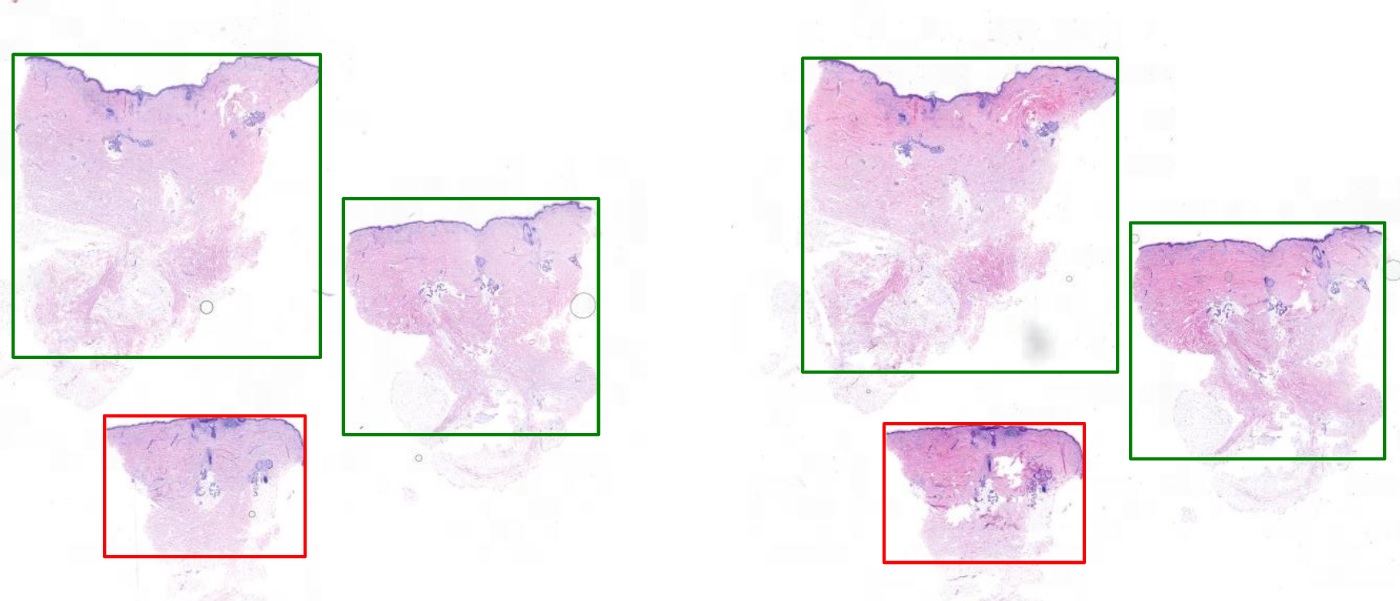}};
\draw (277.59,393.66) node  {\includegraphics[width=392.39pt,height=147.32pt]{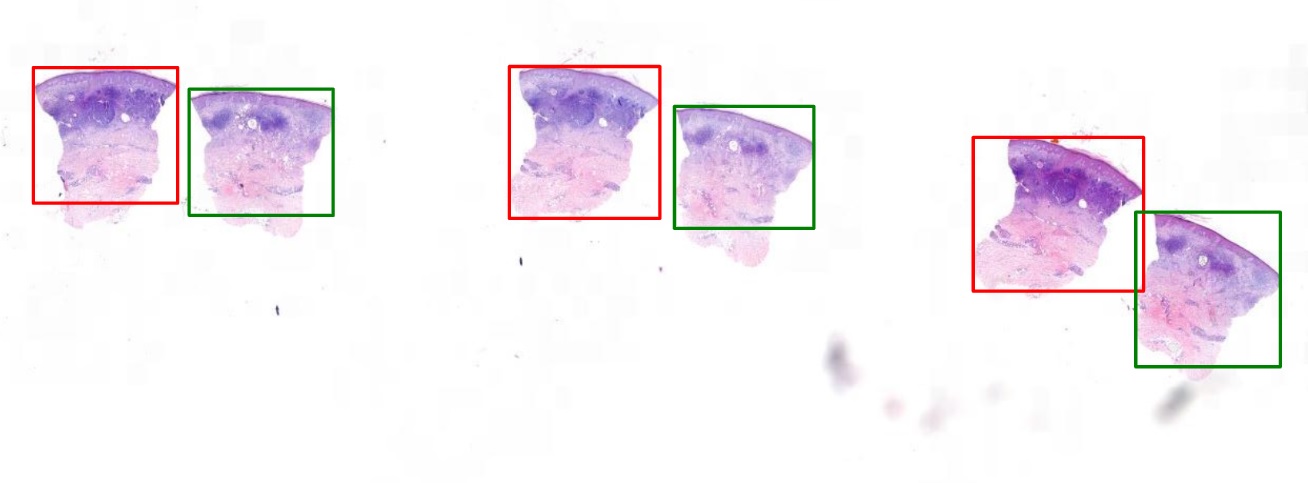}};
\draw   (16,295.44) -- (539.18,295.44) -- (539.18,491.87) -- (16,491.87) -- cycle ;
\draw   (16,509.81) -- (539.18,509.81) -- (539.18,710.22) -- (16,710.22) -- cycle ;

\end{tikzpicture}

%% file: figures/architecture.tex
\tikzset{every picture/.style={line width=0.75pt}} 

\begin{tikzpicture}[x=0.75pt,y=0.75pt,yscale=-1,xscale=1]

\draw (93.68,160.24) node [rotate=-315.65,xslant=0.99] {\includegraphics[width=94.79pt,height=69.24pt]{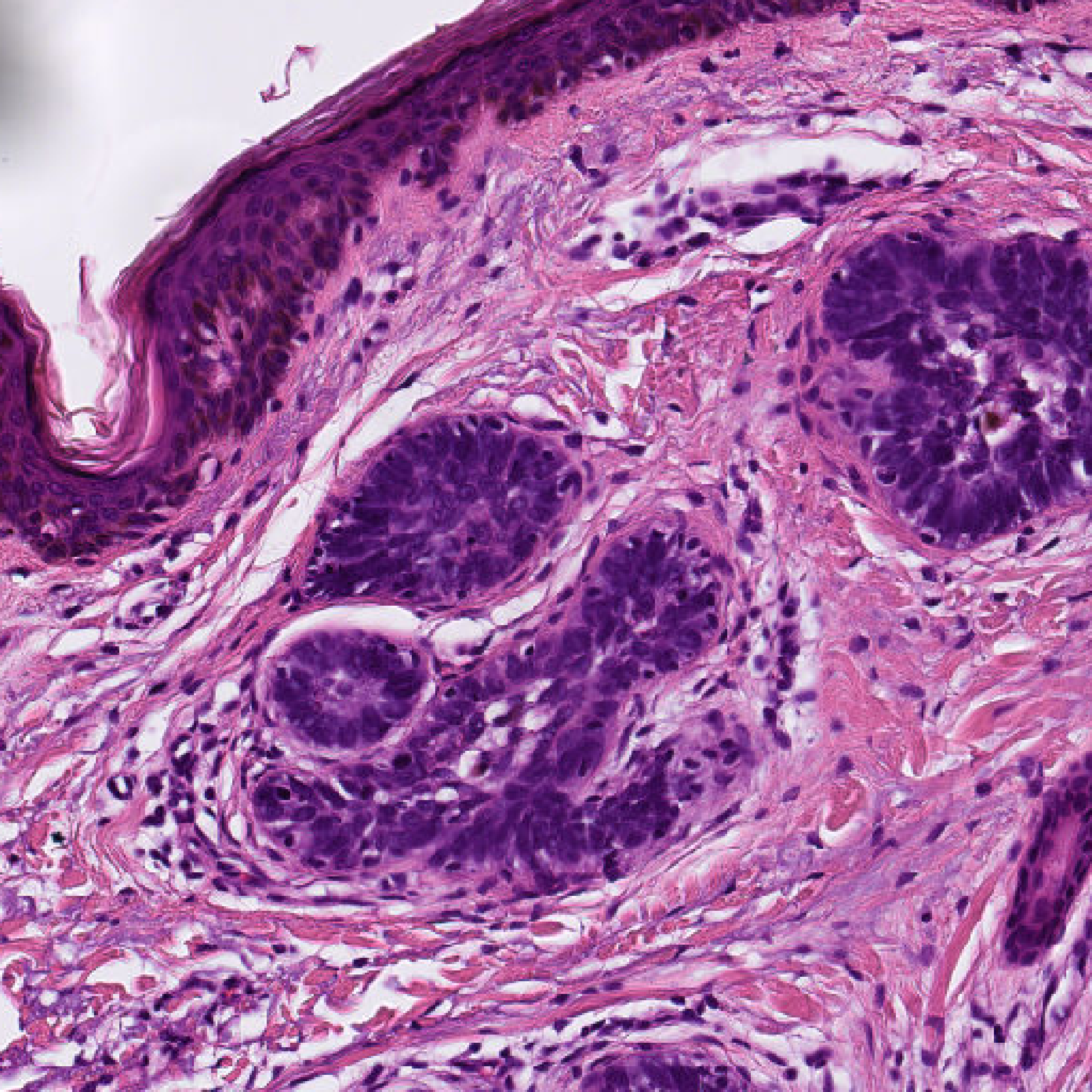}};
\draw  [color={rgb, 255:red, 128; green, 128; blue, 128 }  ,draw opacity=1 ][fill={rgb, 255:red, 255; green, 255; blue, 255 }  ,fill opacity=1 ] (533.15,316.49) -- (578.4,271.24) -- (582.45,271.24) -- (582.45,335.29) -- (537.2,380.54) -- (533.15,380.54) -- cycle ; \draw  [color={rgb, 255:red, 128; green, 128; blue, 128 }  ,draw opacity=1 ] (582.45,271.24) -- (537.2,316.49) -- (533.15,316.49) ; \draw  [color={rgb, 255:red, 128; green, 128; blue, 128 }  ,draw opacity=1 ] (537.2,316.49) -- (537.2,380.54) ;
\draw  [color={rgb, 255:red, 128; green, 128; blue, 128 }  ,draw opacity=1 ][fill={rgb, 255:red, 255; green, 255; blue, 255 }  ,fill opacity=1 ] (488.24,437.01) -- (510.71,414.54) -- (514.84,414.54) -- (514.84,446.68) -- (492.38,469.14) -- (488.24,469.14) -- cycle ; \draw  [color={rgb, 255:red, 128; green, 128; blue, 128 }  ,draw opacity=1 ] (514.84,414.54) -- (492.38,437.01) -- (488.24,437.01) ; \draw  [color={rgb, 255:red, 128; green, 128; blue, 128 }  ,draw opacity=1 ] (492.38,437.01) -- (492.38,469.14) ;
\draw  [fill={rgb, 255:red, 255; green, 255; blue, 255 }  ,fill opacity=1 ] (492.38,437.13) -- (514.97,414.54) -- (523.11,414.54) -- (523.11,446.65) -- (500.53,469.24) -- (492.38,469.24) -- cycle ; \draw   (523.11,414.54) -- (500.53,437.13) -- (492.38,437.13) ; \draw   (500.53,437.13) -- (500.53,469.24) ;
\draw  [fill={rgb, 255:red, 255; green, 255; blue, 255 }  ,fill opacity=1 ] (516.65,437.01) -- (539.11,414.54) -- (543.25,414.54) -- (543.25,446.68) -- (520.79,469.14) -- (516.65,469.14) -- cycle ; \draw   (543.25,414.54) -- (520.79,437.01) -- (516.65,437.01) ; \draw   (520.79,437.01) -- (520.79,469.14) ;
\draw  [fill={rgb, 255:red, 255; green, 255; blue, 255 }  ,fill opacity=1 ] (537.2,316.49) -- (582.45,271.24) -- (586.5,271.24) -- (586.5,335.29) -- (541.25,380.54) -- (537.2,380.54) -- cycle ; \draw   (586.5,271.24) -- (541.25,316.49) -- (537.2,316.49) ; \draw   (541.25,316.49) -- (541.25,380.54) ;
\draw  [fill={rgb, 255:red, 255; green, 255; blue, 255 }  ,fill opacity=1 ] (558.25,316.12) -- (603.13,271.24) -- (605.55,271.24) -- (605.55,335.67) -- (560.68,380.54) -- (558.25,380.54) -- cycle ; \draw   (605.55,271.24) -- (560.68,316.12) -- (558.25,316.12) ; \draw   (560.68,316.12) -- (560.68,380.54) ;
\draw  [fill={rgb, 255:red, 255; green, 255; blue, 255 }  ,fill opacity=1 ] (599.13,107.91) -- (689.29,17.74) -- (691.63,17.74) -- (691.63,146.07) -- (601.46,236.24) -- (599.13,236.24) -- cycle ; \draw   (691.63,17.74) -- (601.46,107.91) -- (599.13,107.91) ; \draw   (601.46,107.91) -- (601.46,236.24) ;
\draw  [fill={rgb, 255:red, 255; green, 255; blue, 255 }  ,fill opacity=1 ] (618.46,107.87) -- (708.58,17.74) -- (709.96,17.74) -- (709.96,146.12) -- (619.83,236.24) -- (618.46,236.24) -- cycle ; \draw   (709.96,17.74) -- (619.83,107.87) -- (618.46,107.87) ; \draw   (619.83,107.87) -- (619.83,236.24) ;
\draw  [color={rgb, 255:red, 128; green, 128; blue, 128 }  ,draw opacity=1 ][fill={rgb, 255:red, 255; green, 255; blue, 255 }  ,fill opacity=1 ] (442.65,514.03) -- (453.54,503.14) -- (461.95,503.14) -- (461.95,519.55) -- (451.06,530.44) -- (442.65,530.44) -- cycle ; \draw  [color={rgb, 255:red, 128; green, 128; blue, 128 }  ,draw opacity=1 ] (461.95,503.14) -- (451.06,514.03) -- (442.65,514.03) ; \draw  [color={rgb, 255:red, 128; green, 128; blue, 128 }  ,draw opacity=1 ] (451.06,514.03) -- (451.06,530.44) ;
\draw  [fill={rgb, 255:red, 255; green, 255; blue, 255 }  ,fill opacity=1 ] (450.75,514.1) -- (461.7,503.14) -- (478.05,503.14) -- (478.05,519.48) -- (467.09,530.44) -- (450.75,530.44) -- cycle ; \draw   (478.05,503.14) -- (467.09,514.1) -- (450.75,514.1) ; \draw   (467.09,514.1) -- (467.09,530.44) ;
\draw  [fill={rgb, 255:red, 255; green, 255; blue, 255 }  ,fill opacity=1 ] (482.09,513.03) -- (492.98,502.14) -- (501.39,502.14) -- (501.39,518.55) -- (490.5,529.44) -- (482.09,529.44) -- cycle ; \draw   (501.39,502.14) -- (490.5,513.03) -- (482.09,513.03) ; \draw   (490.5,513.03) -- (490.5,529.44) ;
\draw  [color={rgb, 255:red, 128; green, 128; blue, 128 }  ,draw opacity=1 ][fill={rgb, 255:red, 255; green, 255; blue, 255 }  ,fill opacity=1 ] (383.08,570.62) -- (388.64,565.07) -- (404.78,565.07) -- (404.78,573.21) -- (399.22,578.77) -- (383.08,578.77) -- cycle ; \draw  [color={rgb, 255:red, 128; green, 128; blue, 128 }  ,draw opacity=1 ] (404.78,565.07) -- (399.22,570.62) -- (383.08,570.62) ; \draw  [color={rgb, 255:red, 128; green, 128; blue, 128 }  ,draw opacity=1 ] (399.22,570.62) -- (399.22,578.77) ;
\draw  [fill={rgb, 255:red, 255; green, 255; blue, 255 }  ,fill opacity=1 ] (399.15,570.67) -- (404.75,565.07) -- (436.85,565.07) -- (436.85,573.17) -- (431.25,578.77) -- (399.15,578.77) -- cycle ; \draw   (436.85,565.07) -- (431.25,570.67) -- (399.15,570.67) ; \draw   (431.25,570.67) -- (431.25,578.77) ;
\draw  [fill={rgb, 255:red, 255; green, 255; blue, 255 }  ,fill opacity=1 ] (447.5,570.62) -- (453.06,565.07) -- (469.2,565.07) -- (469.2,573.21) -- (463.65,578.77) -- (447.5,578.77) -- cycle ; \draw   (469.2,565.07) -- (463.65,570.62) -- (447.5,570.62) ; \draw   (463.65,570.62) -- (463.65,578.77) ;
\draw   (303.27,616.92) -- (305.98,614.2) -- (338.07,614.2) -- (338.07,618.28) -- (335.35,621) -- (303.27,621) -- cycle ; \draw   (338.07,614.2) -- (335.35,616.92) -- (303.27,616.92) ; \draw   (335.35,616.92) -- (335.35,621) ;
\draw  [color={rgb, 255:red, 255; green, 255; blue, 255 }  ,draw opacity=1 ][fill={rgb, 255:red, 74; green, 144; blue, 226 }  ,fill opacity=1 ] (336.84,616) -- (344.04,616) -- (344.04,613) -- (348.84,619) -- (344.04,625) -- (344.04,622) -- (336.84,622) -- cycle ;
\draw   (349.84,616.92) -- (352.56,614.2) -- (384.64,614.2) -- (384.64,618.28) -- (381.93,621) -- (349.84,621) -- cycle ; \draw   (384.64,614.2) -- (381.93,616.92) -- (349.84,616.92) ; \draw   (381.93,616.92) -- (381.93,621) ;
\draw  [color={rgb, 255:red, 255; green, 255; blue, 255 }  ,draw opacity=1 ][fill={rgb, 255:red, 74; green, 144; blue, 226 }  ,fill opacity=1 ] (383.42,616) -- (390.62,616) -- (390.62,613) -- (395.42,619) -- (390.62,625) -- (390.62,622) -- (383.42,622) -- cycle ;
\draw   (396.42,616.92) -- (399.14,614.2) -- (431.22,614.2) -- (431.22,618.28) -- (428.5,621) -- (396.42,621) -- cycle ; \draw   (431.22,614.2) -- (428.5,616.92) -- (396.42,616.92) ; \draw   (428.5,616.92) -- (428.5,621) ;
\draw  [fill={rgb, 255:red, 255; green, 255; blue, 255 }  ,fill opacity=1 ] (159.26,570.62) -- (164.82,565.07) -- (180.96,565.07) -- (180.96,573.21) -- (175.41,578.77) -- (159.26,578.77) -- cycle ; \draw   (180.96,565.07) -- (175.41,570.62) -- (159.26,570.62) ; \draw   (175.41,570.62) -- (175.41,578.77) ;
\draw  [color={rgb, 255:red, 255; green, 255; blue, 255 }  ,draw opacity=1 ][fill={rgb, 255:red, 74; green, 144; blue, 226 }  ,fill opacity=1 ] (176.41,571.07) -- (183.61,571.07) -- (183.61,568.07) -- (188.41,574.07) -- (183.61,580.07) -- (183.61,577.07) -- (176.41,577.07) -- cycle ;
\draw  [fill={rgb, 255:red, 255; green, 255; blue, 255 }  ,fill opacity=1 ] (189.41,570.62) -- (194.96,565.07) -- (211.11,565.07) -- (211.11,573.21) -- (205.55,578.77) -- (189.41,578.77) -- cycle ; \draw   (211.11,565.07) -- (205.55,570.62) -- (189.41,570.62) ; \draw   (205.55,570.62) -- (205.55,578.77) ;
\draw  [color={rgb, 255:red, 255; green, 255; blue, 255 }  ,draw opacity=1 ][fill={rgb, 255:red, 74; green, 144; blue, 226 }  ,fill opacity=1 ] (206.55,571.07) -- (213.75,571.07) -- (213.75,568.07) -- (218.55,574.07) -- (213.75,580.07) -- (213.75,577.07) -- (206.55,577.07) -- cycle ;
\draw  [fill={rgb, 255:red, 255; green, 255; blue, 255 }  ,fill opacity=1 ] (219.55,570.62) -- (225.11,565.07) -- (241.25,565.07) -- (241.25,573.21) -- (235.7,578.77) -- (219.55,578.77) -- cycle ; \draw   (241.25,565.07) -- (235.7,570.62) -- (219.55,570.62) ; \draw   (235.7,570.62) -- (235.7,578.77) ;
\draw  [color={rgb, 255:red, 255; green, 255; blue, 255 }  ,draw opacity=1 ][fill={rgb, 255:red, 74; green, 144; blue, 226 }  ,fill opacity=1 ] (236.7,571.07) -- (243.9,571.07) -- (243.9,568.07) -- (248.7,574.07) -- (243.9,580.07) -- (243.9,577.07) -- (236.7,577.07) -- cycle ;
\draw  [fill={rgb, 255:red, 255; green, 255; blue, 255 }  ,fill opacity=1 ] (249.7,570.62) -- (255.25,565.07) -- (271.4,565.07) -- (271.4,573.21) -- (265.84,578.77) -- (249.7,578.77) -- cycle ; \draw   (271.4,565.07) -- (265.84,570.62) -- (249.7,570.62) ; \draw   (265.84,570.62) -- (265.84,578.77) ;
\draw  [color={rgb, 255:red, 255; green, 255; blue, 255 }  ,draw opacity=1 ][fill={rgb, 255:red, 74; green, 144; blue, 226 }  ,fill opacity=1 ] (266.84,571.07) -- (274.04,571.07) -- (274.04,568.07) -- (278.84,574.07) -- (274.04,580.07) -- (274.04,577.07) -- (266.84,577.07) -- cycle ;
\draw  [fill={rgb, 255:red, 255; green, 255; blue, 255 }  ,fill opacity=1 ] (279.84,570.62) -- (285.4,565.07) -- (301.54,565.07) -- (301.54,573.21) -- (295.98,578.77) -- (279.84,578.77) -- cycle ; \draw   (301.54,565.07) -- (295.98,570.62) -- (279.84,570.62) ; \draw   (295.98,570.62) -- (295.98,578.77) ;
\draw  [color={rgb, 255:red, 255; green, 255; blue, 255 }  ,draw opacity=1 ][fill={rgb, 255:red, 74; green, 144; blue, 226 }  ,fill opacity=1 ] (296.98,571.07) -- (304.18,571.07) -- (304.18,568.07) -- (308.98,574.07) -- (304.18,580.07) -- (304.18,577.07) -- (296.98,577.07) -- cycle ;
\draw  [fill={rgb, 255:red, 255; green, 255; blue, 255 }  ,fill opacity=1 ] (309.98,570.62) -- (315.54,565.07) -- (331.68,565.07) -- (331.68,573.21) -- (326.13,578.77) -- (309.98,578.77) -- cycle ; \draw   (331.68,565.07) -- (326.13,570.62) -- (309.98,570.62) ; \draw   (326.13,570.62) -- (326.13,578.77) ;
\draw  [fill={rgb, 255:red, 255; green, 255; blue, 255 }  ,fill opacity=1 ] (95.94,514.03) -- (106.82,503.14) -- (115.24,503.14) -- (115.24,519.55) -- (104.35,530.44) -- (95.94,530.44) -- cycle ; \draw   (115.24,503.14) -- (104.35,514.03) -- (95.94,514.03) ; \draw   (104.35,514.03) -- (104.35,530.44) ;
\draw  [color={rgb, 255:red, 255; green, 255; blue, 255 }  ,draw opacity=1 ][fill={rgb, 255:red, 74; green, 144; blue, 226 }  ,fill opacity=1 ] (105.84,517.44) -- (113.04,517.44) -- (113.04,514.44) -- (117.84,520.44) -- (113.04,526.44) -- (113.04,523.44) -- (105.84,523.44) -- cycle ;
\draw  [fill={rgb, 255:red, 255; green, 255; blue, 255 }  ,fill opacity=1 ] (118.84,514.03) -- (129.72,503.14) -- (138.14,503.14) -- (138.14,519.55) -- (127.25,530.44) -- (118.84,530.44) -- cycle ; \draw   (138.14,503.14) -- (127.25,514.03) -- (118.84,514.03) ; \draw   (127.25,514.03) -- (127.25,530.44) ;
\draw  [color={rgb, 255:red, 255; green, 255; blue, 255 }  ,draw opacity=1 ][fill={rgb, 255:red, 74; green, 144; blue, 226 }  ,fill opacity=1 ] (128.74,517.44) -- (135.94,517.44) -- (135.94,514.44) -- (140.74,520.44) -- (135.94,526.44) -- (135.94,523.44) -- (128.74,523.44) -- cycle ;
\draw  [fill={rgb, 255:red, 255; green, 255; blue, 255 }  ,fill opacity=1 ] (141.74,514.03) -- (152.62,503.14) -- (161.04,503.14) -- (161.04,519.55) -- (150.15,530.44) -- (141.74,530.44) -- cycle ; \draw   (161.04,503.14) -- (150.15,514.03) -- (141.74,514.03) ; \draw   (150.15,514.03) -- (150.15,530.44) ;
\draw  [color={rgb, 255:red, 255; green, 255; blue, 255 }  ,draw opacity=1 ][fill={rgb, 255:red, 74; green, 144; blue, 226 }  ,fill opacity=1 ] (151.64,517.44) -- (158.84,517.44) -- (158.84,514.44) -- (163.64,520.44) -- (158.84,526.44) -- (158.84,523.44) -- (151.64,523.44) -- cycle ;
\draw  [fill={rgb, 255:red, 255; green, 255; blue, 255 }  ,fill opacity=1 ] (164.64,514.03) -- (175.52,503.14) -- (183.94,503.14) -- (183.94,519.55) -- (173.05,530.44) -- (164.64,530.44) -- cycle ; \draw   (183.94,503.14) -- (173.05,514.03) -- (164.64,514.03) ; \draw   (173.05,514.03) -- (173.05,530.44) ;
\draw  [fill={rgb, 255:red, 255; green, 255; blue, 255 }  ,fill opacity=1 ] (55.81,320.49) -- (101.06,275.24) -- (105.11,275.24) -- (105.11,339.29) -- (59.86,384.54) -- (55.81,384.54) -- cycle ; \draw   (105.11,275.24) -- (59.86,320.49) -- (55.81,320.49) ; \draw   (59.86,320.49) -- (59.86,384.54) ;
\draw  [fill={rgb, 255:red, 255; green, 255; blue, 255 }  ,fill opacity=1 ] (57.81,437.01) -- (80.27,414.54) -- (84.41,414.54) -- (84.41,446.68) -- (61.94,469.14) -- (57.81,469.14) -- cycle ; \draw   (84.41,414.54) -- (61.94,437.01) -- (57.81,437.01) ; \draw   (61.94,437.01) -- (61.94,469.14) ;
\draw  [color={rgb, 255:red, 255; green, 255; blue, 255 }  ,draw opacity=1 ][fill={rgb, 255:red, 74; green, 144; blue, 226 }  ,fill opacity=1 ] (63.41,448.94) -- (70.61,448.94) -- (70.61,445.94) -- (75.41,451.94) -- (70.61,457.94) -- (70.61,454.94) -- (63.41,454.94) -- cycle ;
\draw  [fill={rgb, 255:red, 255; green, 255; blue, 255 }  ,fill opacity=1 ] (76.41,437.01) -- (98.87,414.54) -- (103.01,414.54) -- (103.01,446.68) -- (80.54,469.14) -- (76.41,469.14) -- cycle ; \draw   (103.01,414.54) -- (80.54,437.01) -- (76.41,437.01) ; \draw   (80.54,437.01) -- (80.54,469.14) ;
\draw  [color={rgb, 255:red, 255; green, 255; blue, 255 }  ,draw opacity=1 ][fill={rgb, 255:red, 74; green, 144; blue, 226 }  ,fill opacity=1 ] (82.01,448.94) -- (89.21,448.94) -- (89.21,445.94) -- (94.01,451.94) -- (89.21,457.94) -- (89.21,454.94) -- (82.01,454.94) -- cycle ;
\draw  [fill={rgb, 255:red, 255; green, 255; blue, 255 }  ,fill opacity=1 ] (95.01,437.01) -- (117.47,414.54) -- (121.61,414.54) -- (121.61,446.68) -- (99.14,469.14) -- (95.01,469.14) -- cycle ; \draw   (121.61,414.54) -- (99.14,437.01) -- (95.01,437.01) ; \draw   (99.14,437.01) -- (99.14,469.14) ;
\draw [color={rgb, 255:red, 155; green, 155; blue, 155 }  ,draw opacity=0.9 ][line width=2.25]  [dash pattern={on 2.53pt off 3.02pt}]  (338.48,571.21) -- (348.48,571.21) ;
\draw [shift={(353.48,571.21)}, rotate = 180] [fill={rgb, 255:red, 155; green, 155; blue, 155 }  ,fill opacity=0.9 ][line width=0.08]  [draw opacity=0] (10,-4.8) -- (0,0) -- (10,4.8) -- cycle    ;
\draw [color={rgb, 255:red, 155; green, 155; blue, 155 }  ,draw opacity=0.9 ][line width=2.25]  [dash pattern={on 2.53pt off 3.02pt}]  (192.31,516.4) -- (431.31,516.4) ;
\draw [shift={(436.31,516.4)}, rotate = 180] [fill={rgb, 255:red, 155; green, 155; blue, 155 }  ,fill opacity=0.9 ][line width=0.08]  [draw opacity=0] (10,-4.8) -- (0,0) -- (10,4.8) -- cycle    ;
\draw [color={rgb, 255:red, 155; green, 155; blue, 155 }  ,draw opacity=0.9 ][line width=2.25]  [dash pattern={on 2.53pt off 3.02pt}]  (130.31,443.87) -- (475.31,443.87) ;
\draw [shift={(480.31,443.87)}, rotate = 180] [fill={rgb, 255:red, 155; green, 155; blue, 155 }  ,fill opacity=0.9 ][line width=0.08]  [draw opacity=0] (10,-4.8) -- (0,0) -- (10,4.8) -- cycle    ;
\draw [color={rgb, 255:red, 155; green, 155; blue, 155 }  ,draw opacity=0.9 ][line width=2.25]  [dash pattern={on 2.53pt off 3.02pt}]  (116.31,334.87) -- (518.31,334.87) ;
\draw [shift={(523.31,334.87)}, rotate = 180] [fill={rgb, 255:red, 155; green, 155; blue, 155 }  ,fill opacity=0.9 ][line width=0.08]  [draw opacity=0] (10,-4.8) -- (0,0) -- (10,4.8) -- cycle    ;
\draw  [draw opacity=0][fill={rgb, 255:red, 74; green, 74; blue, 74 }  ,fill opacity=1 ] (664.38,567.01) -- (672.62,567.01) -- (672.62,563.01) -- (678.11,571.01) -- (672.62,579.01) -- (672.62,575.01) -- (664.38,575.01) -- cycle ;
\draw [color={rgb, 255:red, 128; green, 128; blue, 128 }  ,draw opacity=0.6 ][line width=2.25]  [dash pattern={on 2.53pt off 3.02pt}]  (656.11,604.01) -- (673.11,604.01) ;
\draw [shift={(678.11,604.01)}, rotate = 180] [fill={rgb, 255:red, 128; green, 128; blue, 128 }  ,fill opacity=0.6 ][line width=0.08]  [draw opacity=0] (14.29,-6.86) -- (0,0) -- (14.29,6.86) -- cycle    ;
\draw  [draw opacity=0][fill={rgb, 255:red, 139; green, 87; blue, 42 }  ,fill opacity=1 ] (664.38,535.01) -- (672.62,535.01) -- (672.62,531.01) -- (678.11,539.01) -- (672.62,547.01) -- (672.62,543.01) -- (664.38,543.01) -- cycle ;
\draw  [draw opacity=0][fill={rgb, 255:red, 65; green, 117; blue, 5 }  ,fill opacity=1 ] (667.81,515.01) -- (667.81,505.41) -- (664.38,505.41) -- (671.24,499.01) -- (678.11,505.41) -- (674.67,505.41) -- (674.67,515.01) -- cycle ;
\draw  [draw opacity=0][fill={rgb, 255:red, 74; green, 144; blue, 226 }  ,fill opacity=1 ] (665.24,455.01) -- (673.47,455.01) -- (673.47,451.01) -- (678.96,459.01) -- (673.47,467.01) -- (673.47,463.01) -- (665.24,463.01) -- cycle ;
\draw  [draw opacity=0][fill={rgb, 255:red, 208; green, 2; blue, 27 }  ,fill opacity=1 ] (675.53,388.01) -- (675.53,397.61) -- (678.96,397.61) -- (672.1,404.01) -- (665.24,397.61) -- (668.67,397.61) -- (668.67,388.01) -- cycle ;
\draw  [draw opacity=0][fill={rgb, 255:red, 189; green, 16; blue, 224 }  ,fill opacity=1 ] (675.53,340.01) -- (675.53,349.61) -- (678.96,349.61) -- (672.1,356.01) -- (665.24,349.61) -- (668.67,349.61) -- (668.67,340.01) -- cycle ;
\draw  [color={rgb, 255:red, 255; green, 255; blue, 255 }  ,draw opacity=1 ][fill={rgb, 255:red, 139; green, 87; blue, 42 }  ,fill opacity=1 ] (604.42,167) -- (611.62,167) -- (611.62,164) -- (616.42,170) -- (611.62,176) -- (611.62,173) -- (604.42,173) -- cycle ;
\draw  [color={rgb, 255:red, 255; green, 255; blue, 255 }  ,draw opacity=1 ][fill={rgb, 255:red, 139; green, 87; blue, 42 }  ,fill opacity=1 ] (544.42,347) -- (551.62,347) -- (551.62,344) -- (556.42,350) -- (551.62,356) -- (551.62,353) -- (544.42,353) -- cycle ;
\draw  [color={rgb, 255:red, 255; green, 255; blue, 255 }  ,draw opacity=1 ][fill={rgb, 255:red, 139; green, 87; blue, 42 }  ,fill opacity=1 ] (503.42,447) -- (510.62,447) -- (510.62,444) -- (515.42,450) -- (510.62,456) -- (510.62,453) -- (503.42,453) -- cycle ;
\draw  [color={rgb, 255:red, 255; green, 255; blue, 255 }  ,draw opacity=1 ][fill={rgb, 255:red, 139; green, 87; blue, 42 }  ,fill opacity=1 ] (469.09,517.44) -- (476.29,517.44) -- (476.29,514.44) -- (481.09,520.44) -- (476.29,526.44) -- (476.29,523.44) -- (469.09,523.44) -- cycle ;
\draw  [color={rgb, 255:red, 255; green, 255; blue, 255 }  ,draw opacity=1 ][fill={rgb, 255:red, 139; green, 87; blue, 42 }  ,fill opacity=1 ] (433.25,570.77) -- (440.45,570.77) -- (440.45,567.77) -- (445.25,573.77) -- (440.45,579.77) -- (440.45,576.77) -- (433.25,576.77) -- cycle ;
\draw  [color={rgb, 255:red, 255; green, 255; blue, 255 }  ,draw opacity=1 ][fill={rgb, 255:red, 208; green, 2; blue, 27 }  ,fill opacity=1 ] (323.42,584) -- (323.42,591.2) -- (326.42,591.2) -- (320.42,596) -- (314.42,591.2) -- (317.42,591.2) -- (317.42,584) -- cycle ;
\draw  [color={rgb, 255:red, 255; green, 255; blue, 255 }  ,draw opacity=1 ][fill={rgb, 255:red, 208; green, 2; blue, 27 }  ,fill opacity=1 ] (176.42,535) -- (176.42,542.2) -- (179.42,542.2) -- (173.42,547) -- (167.42,542.2) -- (170.42,542.2) -- (170.42,535) -- cycle ;
\draw  [color={rgb, 255:red, 255; green, 255; blue, 255 }  ,draw opacity=1 ][fill={rgb, 255:red, 208; green, 2; blue, 27 }  ,fill opacity=1 ] (114.42,473) -- (114.42,480.2) -- (117.42,480.2) -- (111.42,485) -- (105.42,480.2) -- (108.42,480.2) -- (108.42,473) -- cycle ;
\draw  [color={rgb, 255:red, 255; green, 255; blue, 255 }  ,draw opacity=1 ][fill={rgb, 255:red, 208; green, 2; blue, 27 }  ,fill opacity=1 ] (85.42,381) -- (85.42,388.2) -- (88.42,388.2) -- (82.42,393) -- (76.42,388.2) -- (79.42,388.2) -- (79.42,381) -- cycle ;
\draw  [color={rgb, 255:red, 255; green, 255; blue, 255 }  ,draw opacity=1 ][fill={rgb, 255:red, 189; green, 16; blue, 224 }  ,fill opacity=1 ] (104.42,239) -- (104.42,246.2) -- (107.42,246.2) -- (101.42,251) -- (95.42,246.2) -- (98.42,246.2) -- (98.42,239) -- cycle ;
\draw  [color={rgb, 255:red, 255; green, 255; blue, 255 }  ,draw opacity=1 ][fill={rgb, 255:red, 65; green, 117; blue, 5 }  ,fill opacity=1 ] (598.46,252.24) -- (598.46,245.04) -- (595.46,245.04) -- (601.46,240.24) -- (607.46,245.04) -- (604.46,245.04) -- (604.46,252.24) -- cycle ;
\draw  [color={rgb, 255:red, 255; green, 255; blue, 255 }  ,draw opacity=1 ][fill={rgb, 255:red, 65; green, 117; blue, 5 }  ,fill opacity=1 ] (536.2,396.54) -- (536.2,389.34) -- (533.2,389.34) -- (539.2,384.54) -- (545.2,389.34) -- (542.2,389.34) -- (542.2,396.54) -- cycle ;
\draw  [color={rgb, 255:red, 255; green, 255; blue, 255 }  ,draw opacity=1 ][fill={rgb, 255:red, 65; green, 117; blue, 5 }  ,fill opacity=1 ] (493.53,484.24) -- (493.53,477.04) -- (490.53,477.04) -- (496.53,472.24) -- (502.53,477.04) -- (499.53,477.04) -- (499.53,484.24) -- cycle ;
\draw  [color={rgb, 255:red, 255; green, 255; blue, 255 }  ,draw opacity=1 ][fill={rgb, 255:red, 65; green, 117; blue, 5 }  ,fill opacity=1 ] (456.42,546) -- (456.42,538.8) -- (453.42,538.8) -- (459.42,534) -- (465.42,538.8) -- (462.42,538.8) -- (462.42,546) -- cycle ;
\draw  [color={rgb, 255:red, 255; green, 255; blue, 255 }  ,draw opacity=1 ][fill={rgb, 255:red, 65; green, 117; blue, 5 }  ,fill opacity=1 ] (412.42,595) -- (412.42,587.8) -- (409.42,587.8) -- (415.42,583) -- (421.42,587.8) -- (418.42,587.8) -- (418.42,595) -- cycle ;
\draw   (47.83,140.87) -- (137.96,50.74) -- (139.33,50.74) -- (139.33,179.12) -- (49.21,269.24) -- (47.83,269.24) -- cycle ; \draw   (139.33,50.74) -- (49.21,140.87) -- (47.83,140.87) ; \draw   (49.21,140.87) -- (49.21,269.24) ;
\draw   (638.83,108.87) -- (728.96,18.74) -- (730.33,18.74) -- (730.33,147.12) -- (640.21,237.24) -- (638.83,237.24) -- cycle ; \draw   (730.33,18.74) -- (640.21,108.87) -- (638.83,108.87) ; \draw   (640.21,108.87) -- (640.21,237.24) ;
\draw  [color={rgb, 255:red, 255; green, 255; blue, 255 }  ,draw opacity=1 ][fill={rgb, 255:red, 74; green, 74; blue, 74 }  ,fill opacity=1 ] (624.42,167) -- (631.62,167) -- (631.62,164) -- (636.42,170) -- (631.62,176) -- (631.62,173) -- (624.42,173) -- cycle ;
\draw   (577.83,314.97) -- (622.81,270) -- (623.81,270) -- (623.81,335.13) -- (578.83,380.1) -- (577.83,380.1) -- cycle ; \draw   (623.81,270) -- (578.83,314.97) -- (577.83,314.97) ; \draw   (578.83,314.97) -- (578.83,380.1) ;
\draw   (536.83,435.97) -- (557.81,415) -- (558.81,415) -- (558.81,446.6) -- (537.83,467.58) -- (536.83,467.58) -- cycle ; \draw   (558.81,415) -- (537.83,435.97) -- (536.83,435.97) ; \draw   (537.83,435.97) -- (537.83,467.58) ;
\draw  [color={rgb, 255:red, 255; green, 255; blue, 255 }  ,draw opacity=1 ][fill={rgb, 255:red, 74; green, 74; blue, 74 }  ,fill opacity=1 ] (564.42,347) -- (571.62,347) -- (571.62,344) -- (576.42,350) -- (571.62,356) -- (571.62,353) -- (564.42,353) -- cycle ;
\draw  [color={rgb, 255:red, 255; green, 255; blue, 255 }  ,draw opacity=1 ][fill={rgb, 255:red, 74; green, 74; blue, 74 }  ,fill opacity=1 ] (523.42,447) -- (530.62,447) -- (530.62,444) -- (535.42,450) -- (530.62,456) -- (530.62,453) -- (523.42,453) -- cycle ;
\draw   (507.93,511.2) -- (519.12,500) -- (520.12,500) -- (520.12,516.8) -- (508.93,528) -- (507.93,528) -- cycle ; \draw   (520.12,500) -- (508.93,511.2) -- (507.93,511.2) ; \draw   (508.93,511.2) -- (508.93,528) ;
\draw  [color={rgb, 255:red, 255; green, 255; blue, 255 }  ,draw opacity=1 ][fill={rgb, 255:red, 74; green, 74; blue, 74 }  ,fill opacity=1 ] (493.85,517.44) -- (501.05,517.44) -- (501.05,514.44) -- (505.85,520.44) -- (501.05,526.44) -- (501.05,523.44) -- (493.85,523.44) -- cycle ;
\draw   (480.93,570.41) -- (488.34,563) -- (489.34,563) -- (489.34,571.59) -- (481.93,579) -- (480.93,579) -- cycle ; \draw   (489.34,563) -- (481.93,570.41) -- (480.93,570.41) ; \draw   (481.93,570.41) -- (481.93,579) ;
\draw  [color={rgb, 255:red, 255; green, 255; blue, 255 }  ,draw opacity=1 ][fill={rgb, 255:red, 74; green, 74; blue, 74 }  ,fill opacity=1 ] (466.85,569.48) -- (474.05,569.48) -- (474.05,566.48) -- (478.85,572.48) -- (474.05,578.48) -- (474.05,575.48) -- (466.85,575.48) -- cycle ;
\draw (684.77,128.49) node [rotate=-315.6,xslant=0.99] {\includegraphics[width=95.18pt,height=69.8pt]{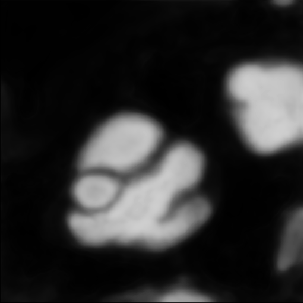}};
\draw (600.82,325.29) node [rotate=-315.6,xslant=0.99] {\includegraphics[width=48.5pt,height=35.18pt]{figures/example8550_output.png}};
\draw (547.82,440.79) node [rotate=-315.6,xslant=0.99] {\includegraphics[width=23.19pt,height=17.76pt]{figures/example8550_output.png}};
\draw (514.53,514) node [rotate=-315.6,xslant=0.99] {\includegraphics[width=11.81pt,height=8.88pt]{figures/example8550_output.png}};
\draw (485.63,571) node [rotate=-315.6,xslant=0.99] {\includegraphics[width=7.81pt,height=4.52pt]{figures/example8550_output.png}};

\draw (31.31,379.54) node  [font=\scriptsize] [align=left] {256x256};
\draw (82.1,405.54) node  [font=\footnotesize] [align=left] {64};
\draw (33.31,464.14) node  [font=\scriptsize] [align=left] {128x128};
\draw (100.7,405.54) node  [font=\footnotesize] [align=left] {64};
\draw (119.3,405.54) node  [font=\footnotesize] [align=left] {64};
\draw (425,526.57) node  [font=\scriptsize] [align=left] {64x64};
\draw (132.91,494.14) node  [font=\footnotesize] [align=left] {128};
\draw (110.01,494.14) node  [font=\footnotesize] [align=left] {128};
\draw (77.94,525.44) node  [font=\scriptsize] [align=left] {64x64};
\draw (155.81,494.14) node  [font=\footnotesize] [align=left] {128};
\draw (178.71,494.14) node  [font=\footnotesize] [align=left] {128};
\draw (172.64,556.07) node  [font=\footnotesize] [align=left] {256};
\draw (141.26,574.77) node  [font=\scriptsize] [align=left] {32x32};
\draw (202.78,556.07) node  [font=\footnotesize] [align=left] {256};
\draw (232.92,556.07) node  [font=\footnotesize] [align=left] {256};
\draw (263.07,556.07) node  [font=\footnotesize] [align=left] {256};
\draw (293.21,556.07) node  [font=\footnotesize] [align=left] {256};
\draw (323.36,556.07) node  [font=\footnotesize] [align=left] {256};
\draw (321.25,605.2) node  [font=\footnotesize] [align=left] {512};
\draw (285.27,618) node  [font=\scriptsize] [align=left] {16x16};
\draw (367.83,605.2) node  [font=\footnotesize] [align=left] {512};
\draw (414.41,605.2) node  [font=\footnotesize] [align=left] {512};
\draw (101.45,262.24) node  [font=\footnotesize] [align=left] {64};
\draw (365.11,574.77) node  [font=\scriptsize] [align=left] {32x32};
\draw (413.91,556.07) node  [font=\footnotesize] [align=left] {768};
\draw (458.88,556.07) node  [font=\footnotesize] [align=left] {256};
\draw (465.38,494.14) node  [font=\footnotesize] [align=left] {384};
\draw (496.36,493.95) node  [font=\footnotesize] [align=left] {128};
\draw (516.67,405.54) node  [font=\footnotesize] [align=left] {192};
\draw (539.94,405.54) node  [font=\footnotesize] [align=left] {64};
\draw (464.74,464.14) node  [font=\scriptsize] [align=left] {128x128};
\draw (509.65,375.54) node  [font=\scriptsize] [align=left] {256x256};
\draw (575.63,231.24) node  [font=\scriptsize] [align=left] {512x512};
\draw (582.12,261.24) node  [font=\footnotesize] [align=left] {128};
\draw (601.46,261.24) node  [font=\footnotesize] [align=left] {32};
\draw (687.96,396.01) node [anchor=west] [inner sep=0.75pt]  [font=\small] [align=left] {\textit{Residual Basic Block with downsampling}\\\textbf{(3x3 conv → Batch Norm → ReLU) →}\\\textbf{(3x3 conv, /2 → Batch Norm →}$\displaystyle \oplus $\textbf{Identity → ReLU)}};
\draw (687.96,459.01) node [anchor=west] [inner sep=0.75pt]  [font=\small] [align=left] {\textit{Residual Basic Block}\\\textbf{(3x3 conv → Batch Norm → ReLU) →}\\\textbf{(3x3 conv → Batch Norm →}$\displaystyle \oplus $\textbf{Identity → ReLU)}};
\draw (686.96,507.01) node [anchor=west] [inner sep=0.75pt]  [font=\small] [align=left] {\textit{Bilinear upsampling}};
\draw (686.96,539.01) node [anchor=west] [inner sep=0.75pt]  [font=\small] [align=left] {\textbf{(3x3 conv → Batch Norm → ReLU) x2}};
\draw (687.96,348.01) node [anchor=west] [inner sep=0.75pt]  [font=\small] [align=left] {\textbf{7x7 conv, /2 → Batch Norm → ReLU}};
\draw (686.96,571.01) node [anchor=west] [inner sep=0.75pt]  [font=\small] [align=left] {\textbf{1x1 conv → softmax}};
\draw (686.96,604.01) node [anchor=west] [inner sep=0.75pt]  [font=\small] [align=left] {\textit{Skip connection}\\\textbf{Copy}};
\draw (23.31,261.54) node  [font=\scriptsize] [align=left] {512x512};
\draw (493.5,570.74) node [anchor=west] [inner sep=0.75pt]  [font=\scriptsize]  {$\psi _{0}( x)$};
\draw (525.5,508.74) node [anchor=west] [inner sep=0.75pt]  [font=\scriptsize]  {$\psi _{1}( x)$};
\draw (565,431.74) node [anchor=west] [inner sep=0.75pt]  [font=\scriptsize]  {$\psi _{2}( x)$};
\draw (631,306.24) node [anchor=west] [inner sep=0.75pt]  [font=\scriptsize]  {$\psi _{3}( x)$};
\draw (738,89.74) node [anchor=west] [inner sep=0.75pt]  [font=\scriptsize]  {$\psi _{4}( x)$};
\draw (690.12,9.24) node  [font=\footnotesize] [align=left] {32};
\draw (709.46,9.24) node  [font=\footnotesize] [align=left] {16};

\end{tikzpicture}

%% file: figures/predictions.tex
\tikzset{every picture/.style={line width=0.75pt}} 

\begin{tikzpicture}[x=0.75pt,y=0.75pt,yscale=-1,xscale=1]

\draw (542.69,1693.52) node  {\includegraphics[width=111.53pt,height=236.44pt]{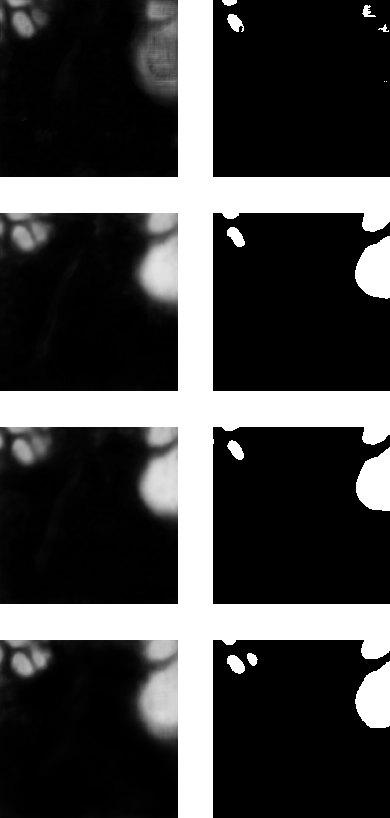}};
\draw (502.19,1478.35) node  {\includegraphics[width=50.78pt,height=51.19pt]{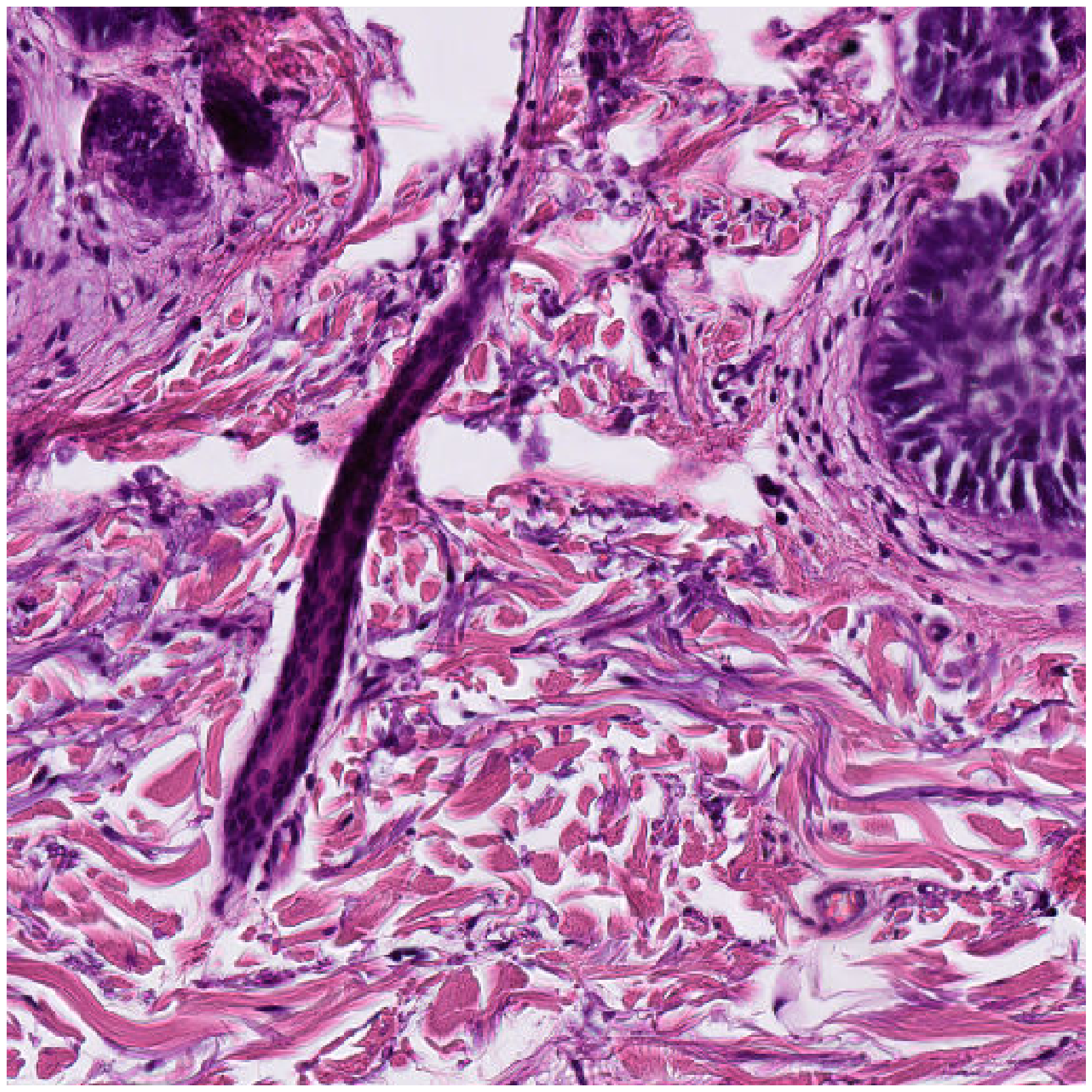}};
\draw (583.19,1478.02) node  {\includegraphics[width=50.78pt,height=51.19pt]{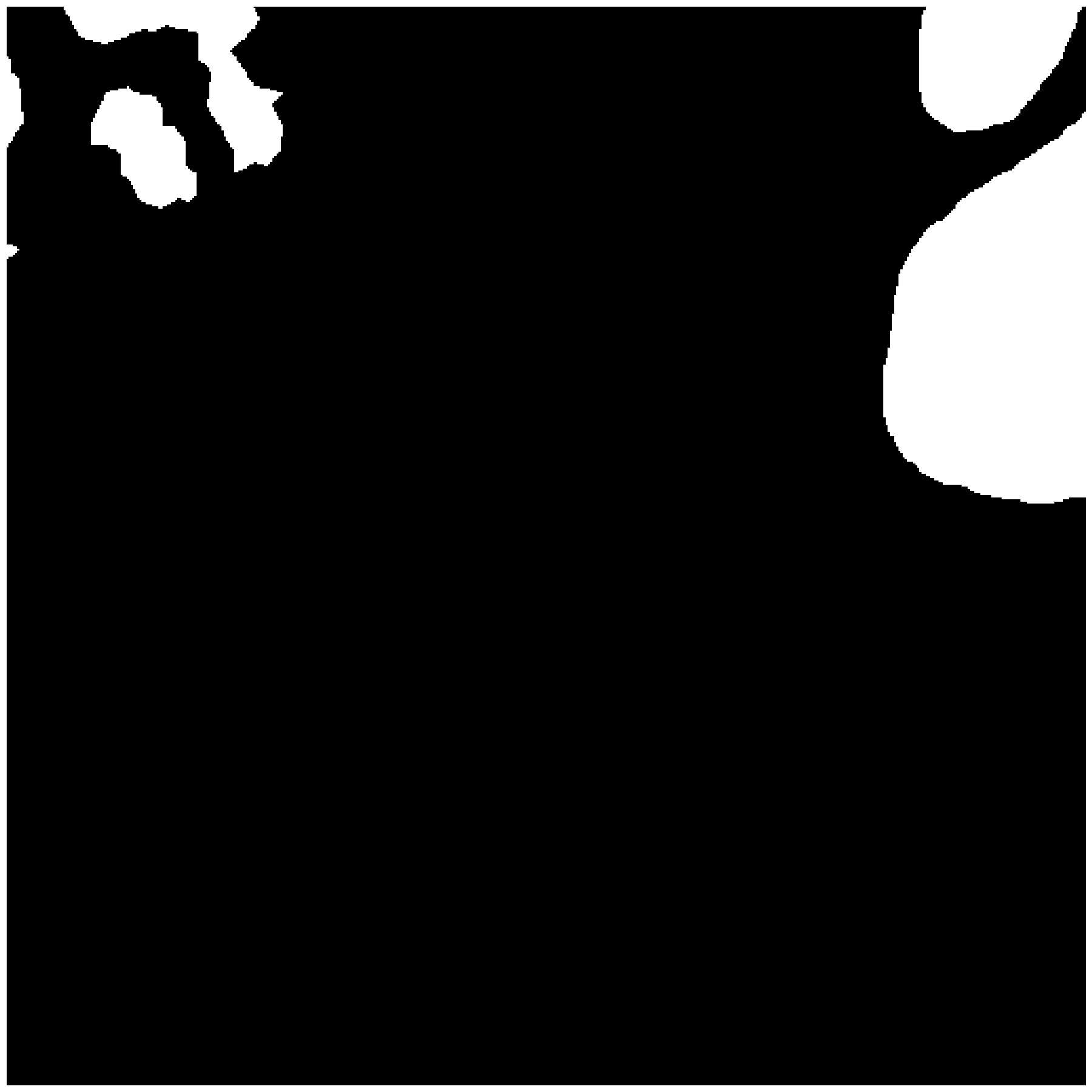}};
\draw (193.69,1693.52) node  {\includegraphics[width=111.53pt,height=236.44pt]{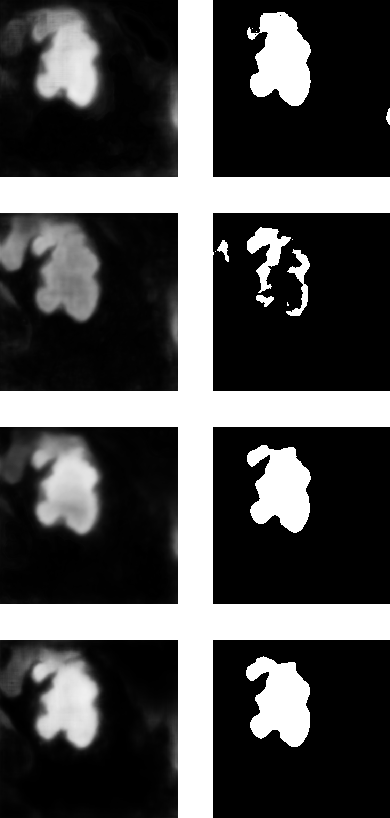}};
\draw (367.69,1694.15) node  {\includegraphics[width=111.53pt,height=238.5pt]{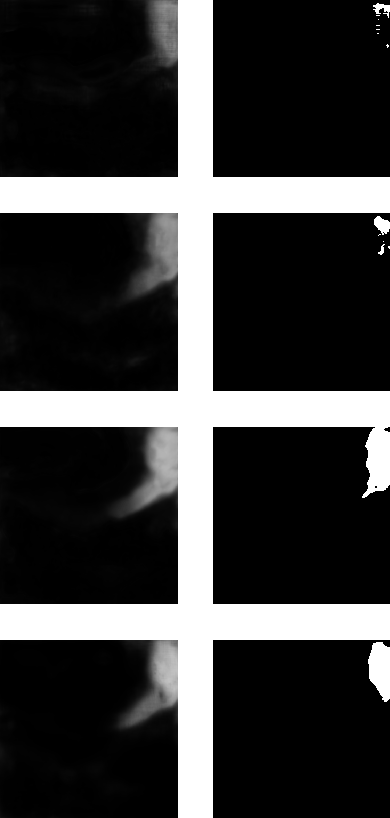}};
\draw (327.19,1478.35) node  {\includegraphics[width=50.78pt,height=51.19pt]{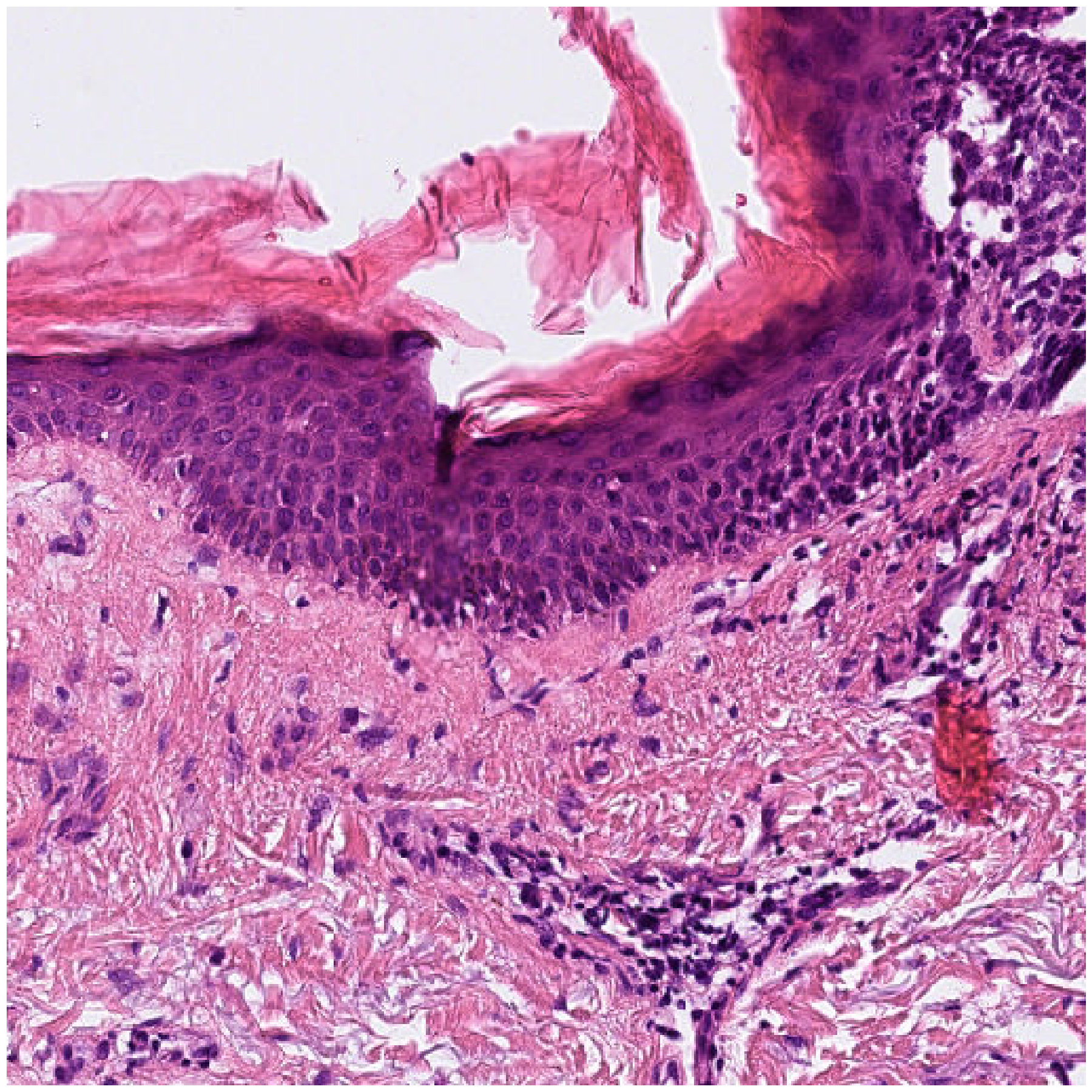}};
\draw (408.19,1478.02) node  {\includegraphics[width=50.78pt,height=51.19pt]{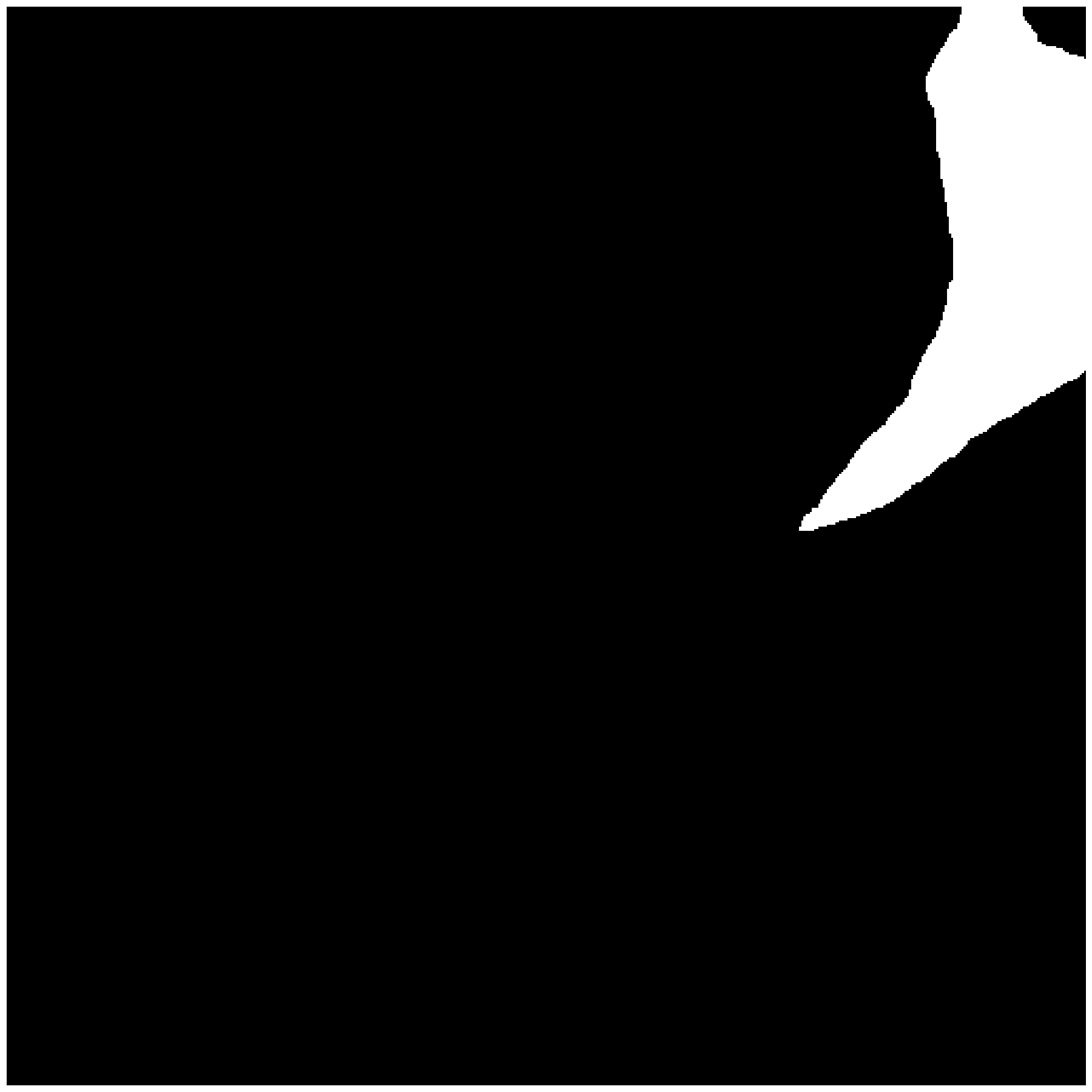}};
\draw (153.19,1477.48) node  {\includegraphics[width=50.78pt,height=52.5pt]{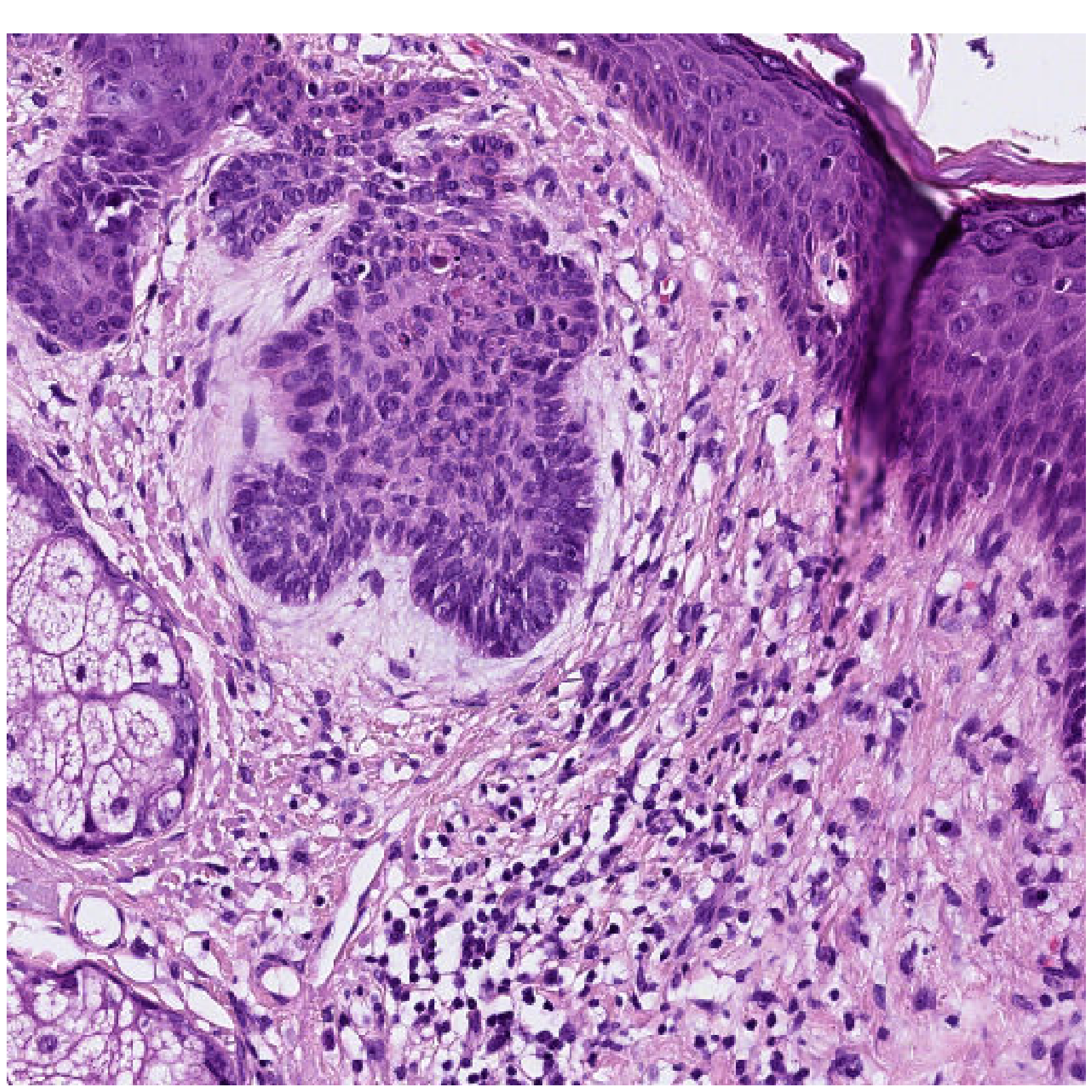}};
\draw (234.19,1478.02) node  {\includegraphics[width=50.78pt,height=51.19pt]{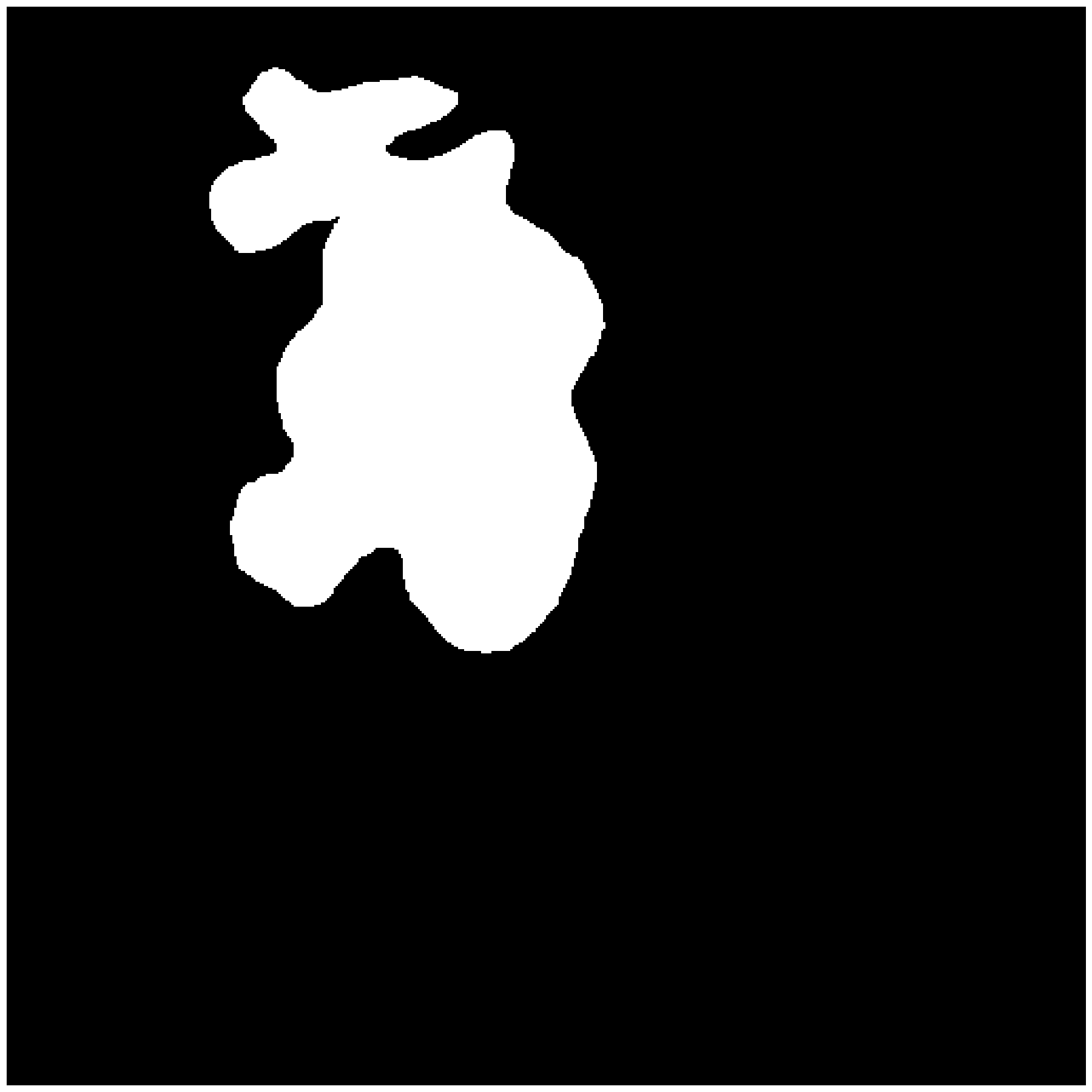}};
\draw (193.69,1260.56) node  {\includegraphics[width=111.54pt,height=236.5pt]{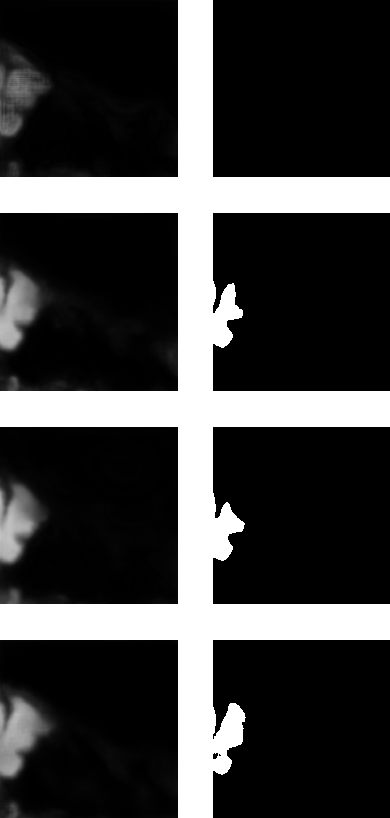}};
\draw (153.19,1045.35) node  {\includegraphics[width=50.78pt,height=51.19pt]{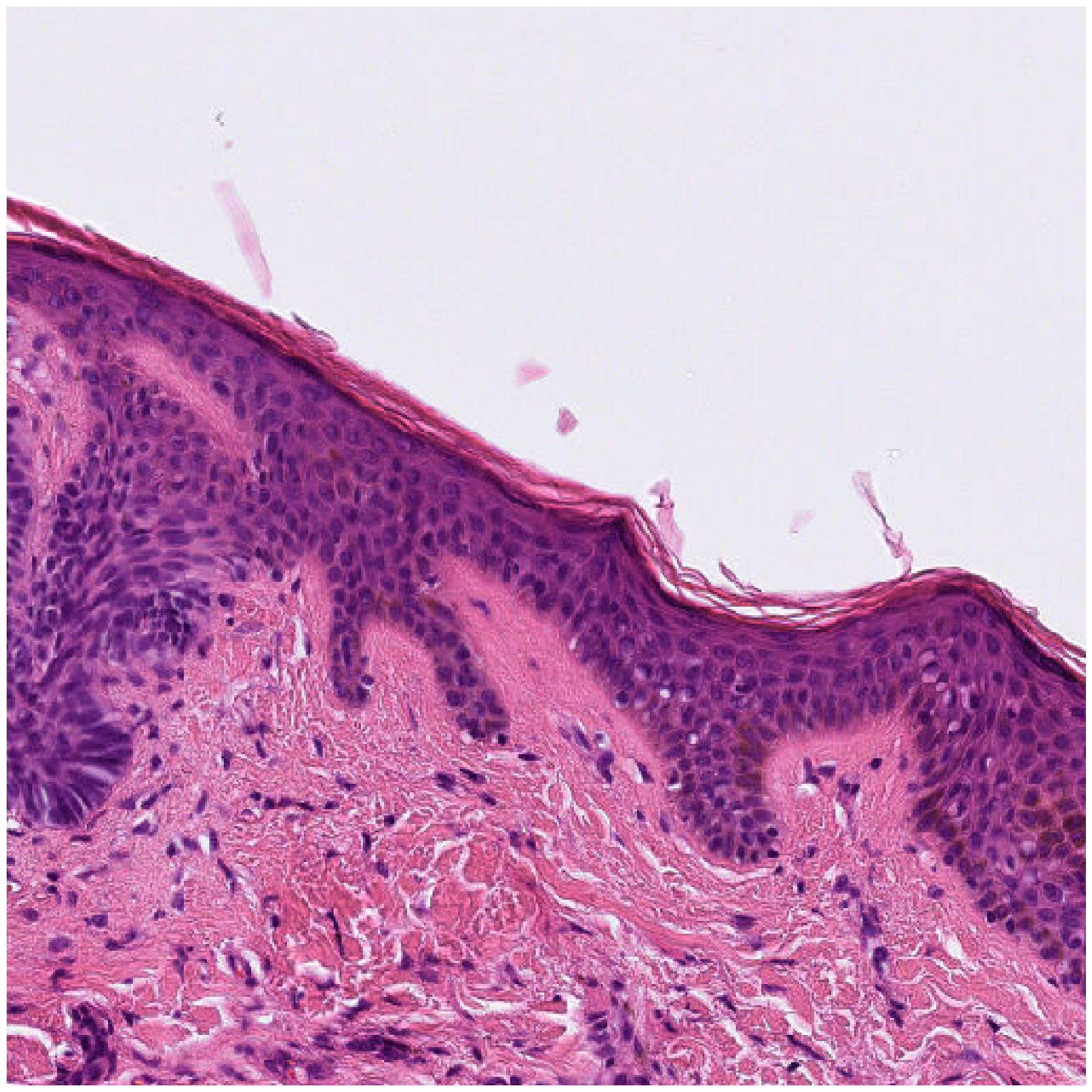}};
\draw (234.19,1045.02) node  {\includegraphics[width=50.78pt,height=51.19pt]{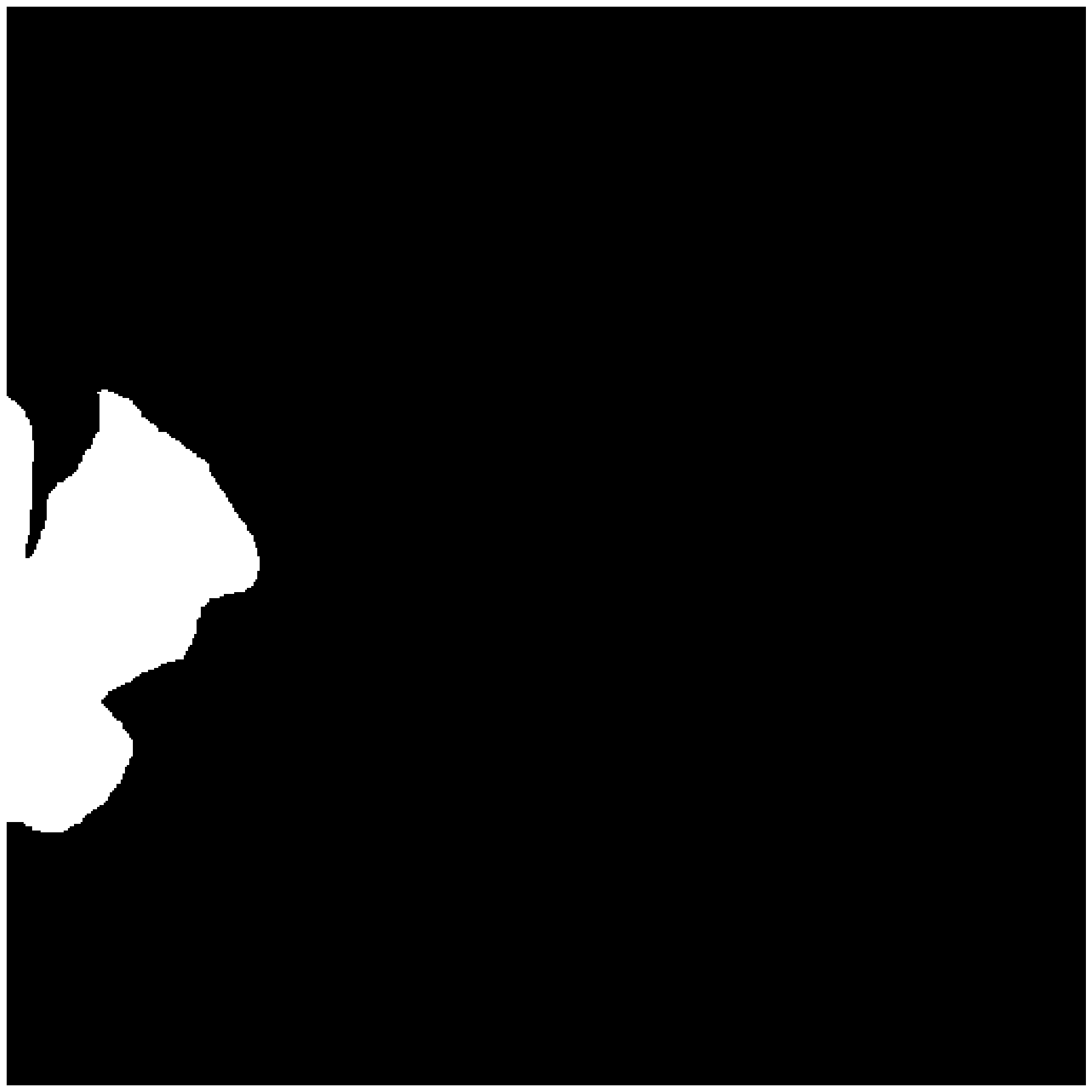}};
\draw (327.19,1045.35) node  {\includegraphics[width=50.78pt,height=51.19pt]{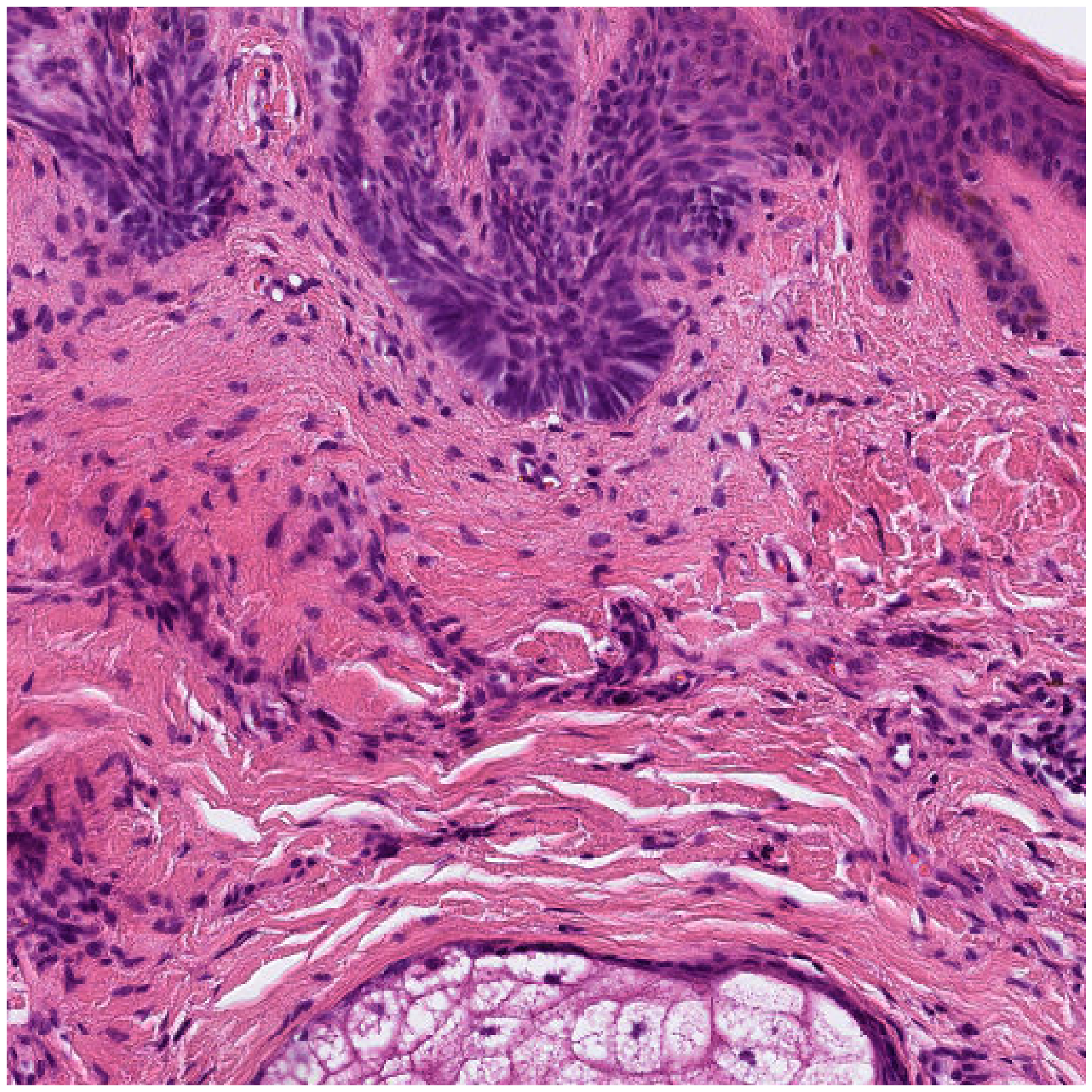}};
\draw (408.19,1045.02) node  {\includegraphics[width=50.78pt,height=51.19pt]{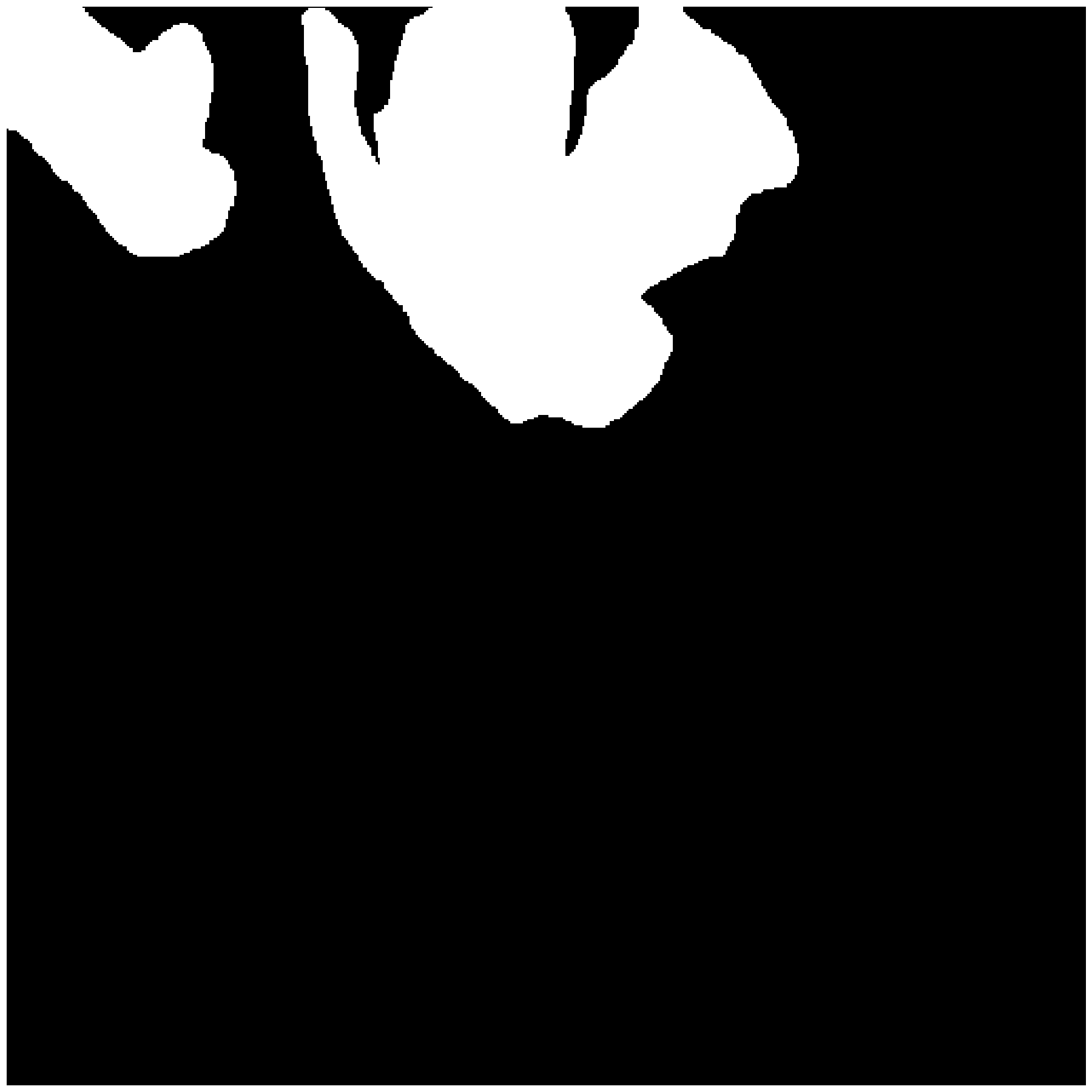}};
\draw (367.69,1261.61) node  {\includegraphics[width=111.54pt,height=238.08pt]{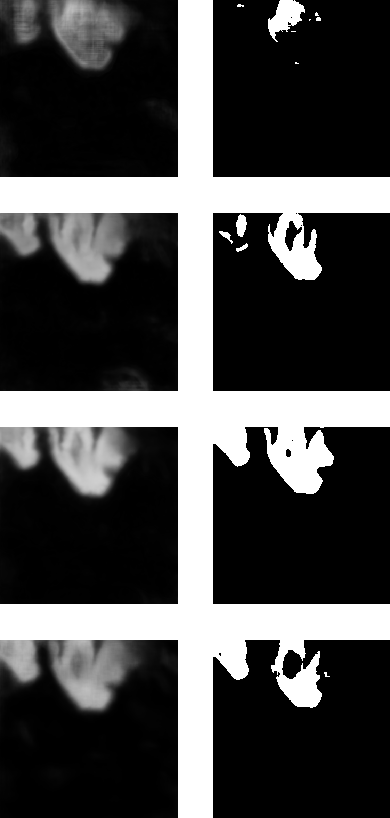}};
\draw (542.69,1260.52) node  {\includegraphics[width=111.54pt,height=236.44pt]{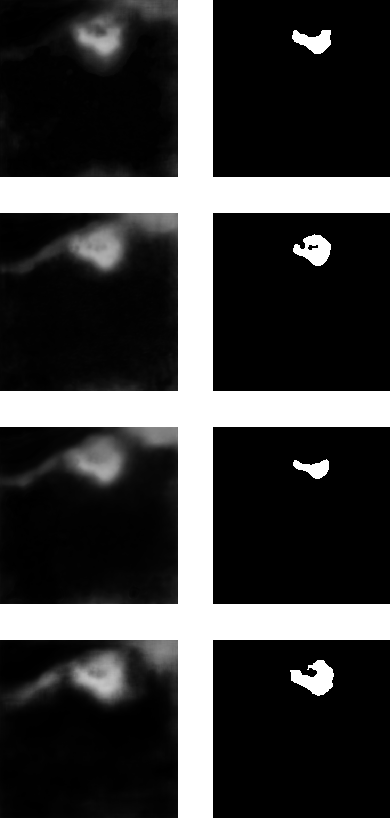}};
\draw (502.19,1045.21) node  {\includegraphics[width=50.78pt,height=51.4pt]{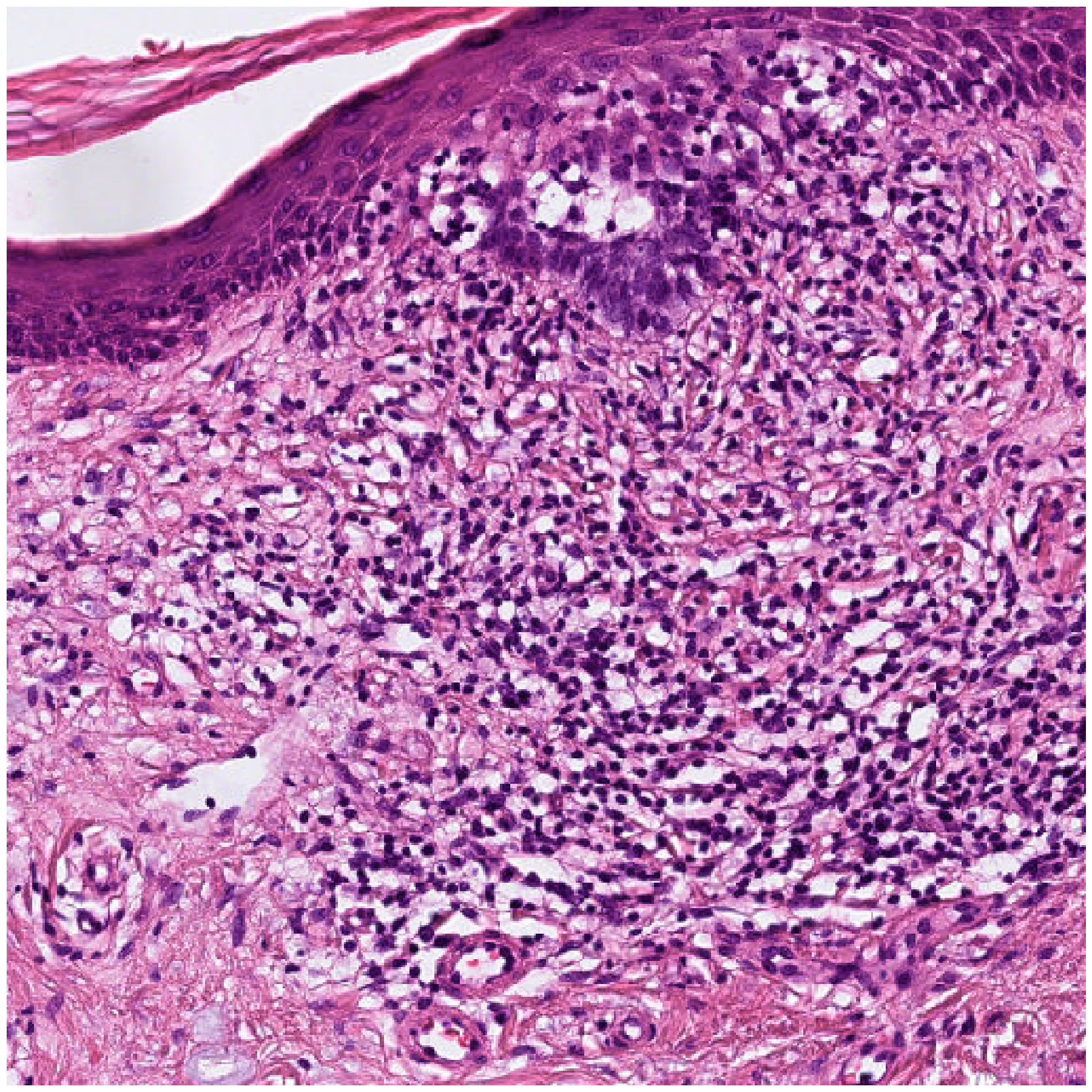}};
\draw   (293.33,1011.22) -- (361.04,1011.22) -- (361.04,1079.48) -- (293.33,1079.48) -- cycle ;
\draw   (374.33,1010.89) -- (442.04,1010.89) -- (442.04,1079.15) -- (374.33,1079.15) -- cycle ;
\draw   (293.33,1102.89) -- (361.04,1102.89) -- (361.04,1171.15) -- (293.33,1171.15) -- cycle ;
\draw   (374.33,1102.89) -- (442.04,1102.89) -- (442.04,1171.15) -- (374.33,1171.15) -- cycle ;
\draw   (119.33,1011.22) -- (187.04,1011.22) -- (187.04,1079.48) -- (119.33,1079.48) -- cycle ;
\draw   (200.33,1010.89) -- (268.04,1010.89) -- (268.04,1079.15) -- (200.33,1079.15) -- cycle ;
\draw   (119.33,1102.89) -- (187.04,1102.89) -- (187.04,1171.15) -- (119.33,1171.15) -- cycle ;
\draw   (119.33,1184.89) -- (187.04,1184.89) -- (187.04,1253.15) -- (119.33,1253.15) -- cycle ;
\draw   (200.33,1184.89) -- (268.04,1184.89) -- (268.04,1253.15) -- (200.33,1253.15) -- cycle ;
\draw   (200.33,1102.89) -- (268.04,1102.89) -- (268.04,1171.15) -- (200.33,1171.15) -- cycle ;
\draw   (119.33,1267.89) -- (187.04,1267.89) -- (187.04,1336.15) -- (119.33,1336.15) -- cycle ;
\draw   (200.33,1267.89) -- (268.04,1267.89) -- (268.04,1336.15) -- (200.33,1336.15) -- cycle ;
\draw   (468.33,1011.22) -- (536.04,1011.22) -- (536.04,1079.48) -- (468.33,1079.48) -- cycle ;
\draw   (549.33,1010.89) -- (617.04,1010.89) -- (617.04,1079.15) -- (549.33,1079.15) -- cycle ;
\draw   (468.33,1102.89) -- (536.04,1102.89) -- (536.04,1171.15) -- (468.33,1171.15) -- cycle ;
\draw   (549.33,1102.89) -- (617.04,1102.89) -- (617.04,1171.15) -- (549.33,1171.15) -- cycle ;
\draw   (119.33,1349.89) -- (187.04,1349.89) -- (187.04,1418.15) -- (119.33,1418.15) -- cycle ;
\draw   (200.33,1349.89) -- (268.04,1349.89) -- (268.04,1418.15) -- (200.33,1418.15) -- cycle ;
\draw   (293.33,1185.89) -- (361.04,1185.89) -- (361.04,1254.15) -- (293.33,1254.15) -- cycle ;
\draw   (374.33,1185.89) -- (442.04,1185.89) -- (442.04,1254.15) -- (374.33,1254.15) -- cycle ;
\draw   (293.33,1268.89) -- (361.04,1268.89) -- (361.04,1337.15) -- (293.33,1337.15) -- cycle ;
\draw   (374.33,1268.89) -- (442.04,1268.89) -- (442.04,1337.15) -- (374.33,1337.15) -- cycle ;
\draw   (293.33,1350.89) -- (361.04,1350.89) -- (361.04,1419.15) -- (293.33,1419.15) -- cycle ;
\draw   (374.33,1350.89) -- (442.04,1350.89) -- (442.04,1419.15) -- (374.33,1419.15) -- cycle ;
\draw   (468.33,1184.89) -- (536.04,1184.89) -- (536.04,1253.15) -- (468.33,1253.15) -- cycle ;
\draw   (549.33,1184.89) -- (617.04,1184.89) -- (617.04,1253.15) -- (549.33,1253.15) -- cycle ;
\draw   (468.33,1267.89) -- (536.04,1267.89) -- (536.04,1336.15) -- (468.33,1336.15) -- cycle ;
\draw   (549.33,1267.89) -- (617.04,1267.89) -- (617.04,1336.15) -- (549.33,1336.15) -- cycle ;
\draw   (468.33,1349.89) -- (536.04,1349.89) -- (536.04,1418.15) -- (468.33,1418.15) -- cycle ;
\draw   (549.33,1349.89) -- (617.04,1349.89) -- (617.04,1418.15) -- (549.33,1418.15) -- cycle ;
\draw (583.19,1045.02) node  {\includegraphics[width=50.78pt,height=51.19pt]{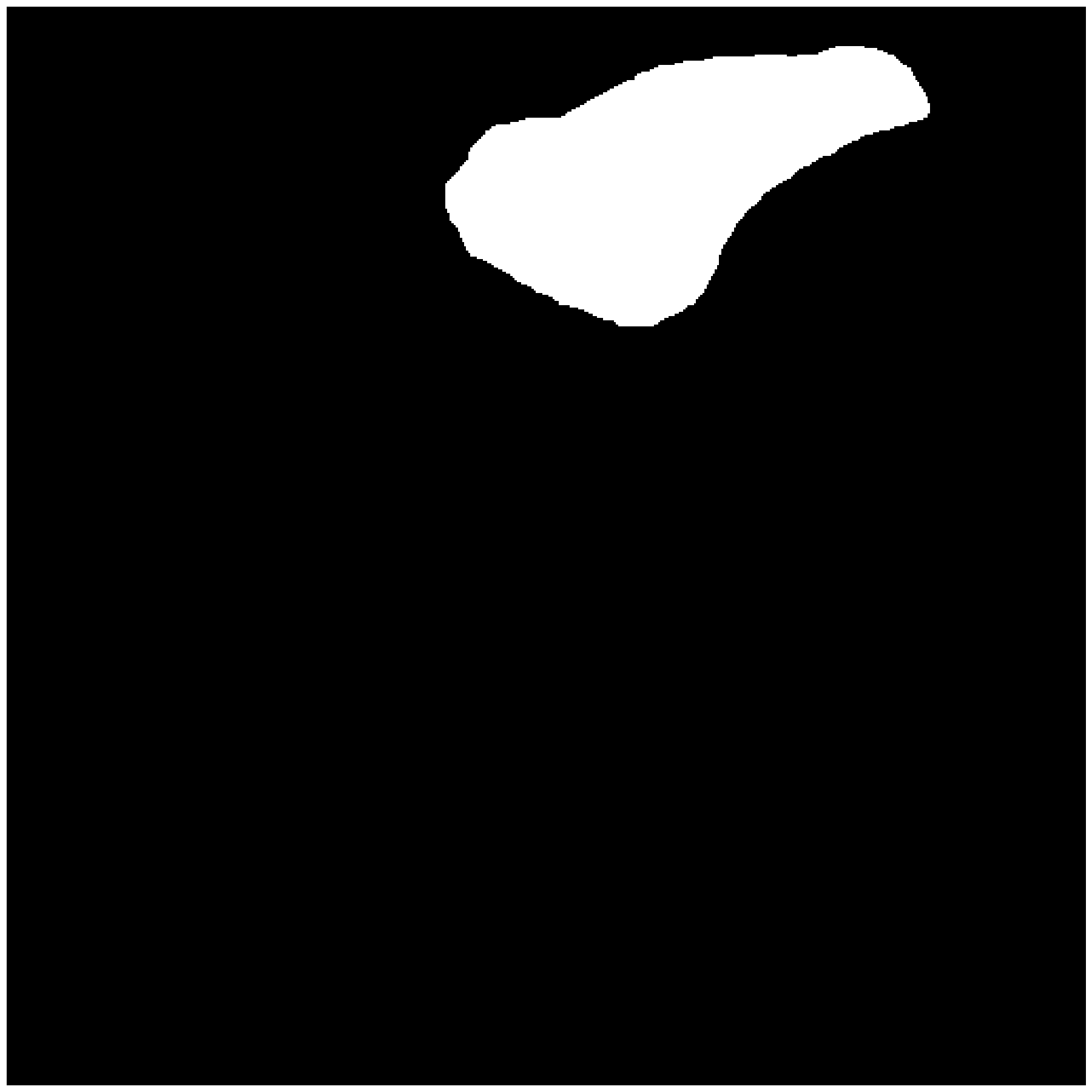}};
\draw   (293.33,1444.22) -- (361.04,1444.22) -- (361.04,1512.48) -- (293.33,1512.48) -- cycle ;
\draw   (374.33,1443.89) -- (442.04,1443.89) -- (442.04,1512.15) -- (374.33,1512.15) -- cycle ;
\draw   (293.33,1535.89) -- (361.04,1535.89) -- (361.04,1604.15) -- (293.33,1604.15) -- cycle ;
\draw   (374.33,1535.89) -- (442.04,1535.89) -- (442.04,1604.15) -- (374.33,1604.15) -- cycle ;
\draw   (119.33,1444.22) -- (187.04,1444.22) -- (187.04,1512.48) -- (119.33,1512.48) -- cycle ;
\draw   (200.33,1443.89) -- (268.04,1443.89) -- (268.04,1512.15) -- (200.33,1512.15) -- cycle ;
\draw   (119.33,1535.89) -- (187.04,1535.89) -- (187.04,1604.15) -- (119.33,1604.15) -- cycle ;
\draw   (119.33,1617.89) -- (187.04,1617.89) -- (187.04,1686.15) -- (119.33,1686.15) -- cycle ;
\draw   (200.33,1617.89) -- (268.04,1617.89) -- (268.04,1686.15) -- (200.33,1686.15) -- cycle ;
\draw   (200.33,1535.89) -- (268.04,1535.89) -- (268.04,1604.15) -- (200.33,1604.15) -- cycle ;
\draw   (119.33,1700.89) -- (187.04,1700.89) -- (187.04,1769.15) -- (119.33,1769.15) -- cycle ;
\draw   (200.33,1700.89) -- (268.04,1700.89) -- (268.04,1769.15) -- (200.33,1769.15) -- cycle ;
\draw   (468.33,1444.22) -- (536.04,1444.22) -- (536.04,1512.48) -- (468.33,1512.48) -- cycle ;
\draw   (549.33,1443.89) -- (617.04,1443.89) -- (617.04,1512.15) -- (549.33,1512.15) -- cycle ;
\draw   (468.33,1535.89) -- (536.04,1535.89) -- (536.04,1604.15) -- (468.33,1604.15) -- cycle ;
\draw   (549.33,1535.89) -- (617.04,1535.89) -- (617.04,1604.15) -- (549.33,1604.15) -- cycle ;
\draw   (119.33,1782.89) -- (187.04,1782.89) -- (187.04,1851.15) -- (119.33,1851.15) -- cycle ;
\draw   (200.33,1782.89) -- (268.04,1782.89) -- (268.04,1851.15) -- (200.33,1851.15) -- cycle ;
\draw   (293.33,1618.89) -- (361.04,1618.89) -- (361.04,1687.15) -- (293.33,1687.15) -- cycle ;
\draw   (374.33,1618.89) -- (442.04,1618.89) -- (442.04,1687.15) -- (374.33,1687.15) -- cycle ;
\draw   (293.33,1701.89) -- (361.04,1701.89) -- (361.04,1770.15) -- (293.33,1770.15) -- cycle ;
\draw   (374.33,1701.89) -- (442.04,1701.89) -- (442.04,1770.15) -- (374.33,1770.15) -- cycle ;
\draw   (293.33,1783.89) -- (361.04,1783.89) -- (361.04,1852.15) -- (293.33,1852.15) -- cycle ;
\draw   (374.33,1783.89) -- (442.04,1783.89) -- (442.04,1852.15) -- (374.33,1852.15) -- cycle ;
\draw   (468.33,1617.89) -- (536.04,1617.89) -- (536.04,1686.15) -- (468.33,1686.15) -- cycle ;
\draw   (549.33,1617.89) -- (617.04,1617.89) -- (617.04,1686.15) -- (549.33,1686.15) -- cycle ;
\draw   (468.33,1700.89) -- (536.04,1700.89) -- (536.04,1769.15) -- (468.33,1769.15) -- cycle ;
\draw   (549.33,1700.89) -- (617.04,1700.89) -- (617.04,1769.15) -- (549.33,1769.15) -- cycle ;
\draw   (468.33,1782.89) -- (536.04,1782.89) -- (536.04,1851.15) -- (468.33,1851.15) -- cycle ;
\draw   (549.33,1782.89) -- (617.04,1782.89) -- (617.04,1851.15) -- (549.33,1851.15) -- cycle ;

\draw (153.09,1000.86) node  [font=\scriptsize]  {$x$};
\draw (239.97,1000.86) node  [font=\scriptsize]  {$y$};
\draw (152.71,1092.95) node  [font=\scriptsize]  {$\Phi ( x)$};
\draw (239.38,1091.61) node  [font=\scriptsize]  {$\Phi ( x) \geq \delta $};
\draw (115.88,1302.58) node [anchor=east] [inner sep=0.75pt]  [font=\scriptsize] [align=left] {\begin{minipage}[lt]{65.22335600000001pt}\setlength\topsep{0pt}
\begin{flushright}
ResNet34-UNet\\+ DS
\end{flushright}

\end{minipage}};
\draw (115.52,1384.23) node [anchor=east] [inner sep=0.75pt]  [font=\scriptsize] [align=left] {\begin{minipage}[lt]{53.106096pt}\setlength\topsep{0pt}
\begin{flushright}
ResNet34-UNet\\+ Linear
\end{flushright}

\end{minipage}};
\draw (115.38,1219.08) node [anchor=east] [inner sep=0.75pt]  [font=\scriptsize] [align=left] {\begin{minipage}[lt]{53.106096pt}\setlength\topsep{0pt}
\begin{flushright}
ResNet34-UNet
\end{flushright}

\end{minipage}};
\draw (114.88,1135.08) node [anchor=east] [inner sep=0.75pt]  [font=\scriptsize] [align=left] {\begin{minipage}[lt]{18.992808pt}\setlength\topsep{0pt}
\begin{flushright}
UNet
\end{flushright}

\end{minipage}};
\draw (327.09,1000.86) node  [font=\scriptsize]  {$x$};
\draw (413.97,1000.86) node  [font=\scriptsize]  {$y$};
\draw (326.71,1092.95) node  [font=\scriptsize]  {$\Phi ( x)$};
\draw (413.38,1091.61) node  [font=\scriptsize]  {$\Phi ( x) \geq \delta $};
\draw (502.09,1000.86) node  [font=\scriptsize]  {$x$};
\draw (588.97,1000.86) node  [font=\scriptsize]  {$y$};
\draw (501.71,1092.95) node  [font=\scriptsize]  {$\Phi ( x)$};
\draw (588.38,1091.61) node  [font=\scriptsize]  {$\Phi ( x) \geq \delta $};
\draw (153.09,1433.86) node  [font=\scriptsize]  {$x$};
\draw (239.97,1433.86) node  [font=\scriptsize]  {$y$};
\draw (152.71,1525.95) node  [font=\scriptsize]  {$\Phi ( x)$};
\draw (239.38,1524.61) node  [font=\scriptsize]  {$\Phi ( x) \geq \delta $};
\draw (115.88,1735.58) node [anchor=east] [inner sep=0.75pt]  [font=\scriptsize] [align=left] {\begin{minipage}[lt]{65.22335600000001pt}\setlength\topsep{0pt}
\begin{flushright}
ResNet34-UNet\\+ DS
\end{flushright}

\end{minipage}};
\draw (115.52,1817.23) node [anchor=east] [inner sep=0.75pt]  [font=\scriptsize] [align=left] {\begin{minipage}[lt]{53.106096pt}\setlength\topsep{0pt}
\begin{flushright}
ResNet34-UNet\\+ Linear
\end{flushright}

\end{minipage}};
\draw (115.38,1652.08) node [anchor=east] [inner sep=0.75pt]  [font=\scriptsize] [align=left] {\begin{minipage}[lt]{53.106096pt}\setlength\topsep{0pt}
\begin{flushright}
ResNet34-UNet
\end{flushright}

\end{minipage}};
\draw (114.88,1568.08) node [anchor=east] [inner sep=0.75pt]  [font=\scriptsize] [align=left] {\begin{minipage}[lt]{18.992808pt}\setlength\topsep{0pt}
\begin{flushright}
UNet
\end{flushright}

\end{minipage}};
\draw (327.09,1433.86) node  [font=\scriptsize]  {$x$};
\draw (413.97,1433.86) node  [font=\scriptsize]  {$y$};
\draw (326.71,1525.95) node  [font=\scriptsize]  {$\Phi ( x)$};
\draw (413.38,1524.61) node  [font=\scriptsize]  {$\Phi ( x) \geq \delta $};
\draw (502.09,1433.86) node  [font=\scriptsize]  {$x$};
\draw (588.97,1433.86) node  [font=\scriptsize]  {$y$};
\draw (501.71,1525.95) node  [font=\scriptsize]  {$\Phi ( x)$};
\draw (588.38,1524.61) node  [font=\scriptsize]  {$\Phi ( x) \geq \delta $};

\end{tikzpicture}

%% file: figures/misclassifications_test.tex
\pgfplotstableread[row sep=\\,col sep=&]{
    model & fp & fn \\
    UNet & 133 & 33  \\
    ResNet34-UNet & 33 & 51 \\
    ResNet34-UNet + DS & 30 & 44 \\
    ResNet34-UNet + Linear & 25 & 59\\
}\mydata

\begin{tikzpicture}
    \begin{axis}[
            ybar,
            bar width=.4cm,
            width=0.9\textwidth,
            height=.35\textwidth,
            legend style={at={(0.5,1)},
                anchor=north,legend columns=-1},
            symbolic x coords={
                UNet,
                ResNet34-UNet,
                ResNet34-UNet + DS,
                ResNet34-UNet + Linear
            },
            xtick={
                UNet,
                ResNet34-UNet,
                ResNet34-UNet + DS,
                ResNet34-UNet + Linear
            },
            xticklabel style={text width=3cm, align=center},
            nodes near coords,
            nodes near coords align={vertical},
            ymin=0,ymax=180,
            ylabel={Misclassified sections}
        ]
        \addplot table[x=model,y=fp]{\mydata};
        \addplot table[x=model,y=fn]{\mydata};
        \legend{False Positives, False Negatives}
    \end{axis}
\end{tikzpicture}

%% file: figures/heatmaps.tex
\tikzset{every picture/.style={line width=0.75pt}} 

\begin{tikzpicture}[x=0.75pt,y=0.75pt,yscale=-1,xscale=1]

\draw (292,94.84) node  {\includegraphics[width=420pt,height=125.85pt]{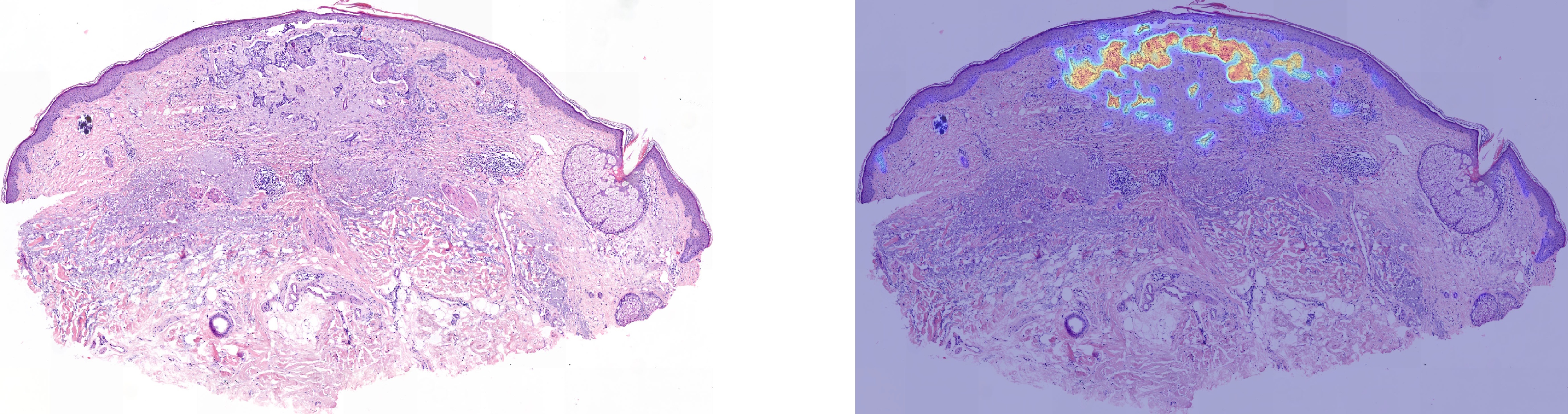}};
\draw (291.47,296.94) node  {\includegraphics[width=420.79pt,height=126pt]{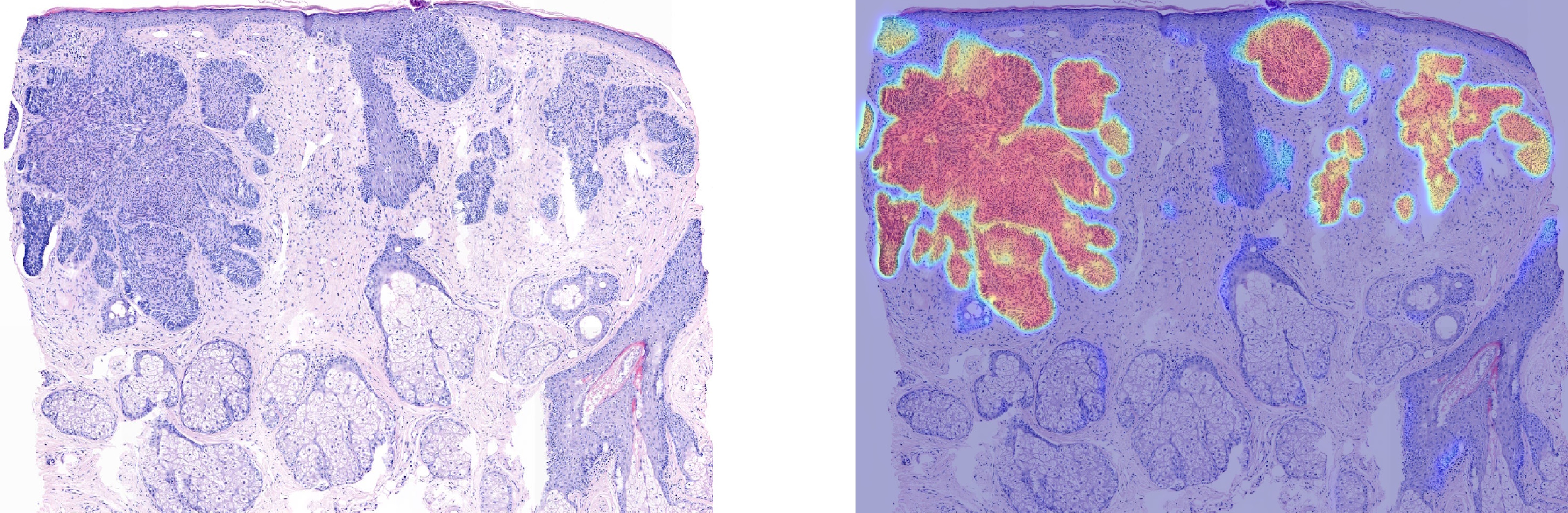}};
\draw (291.03,512.96) node  {\includegraphics[width=421.46pt,height=146.97pt]{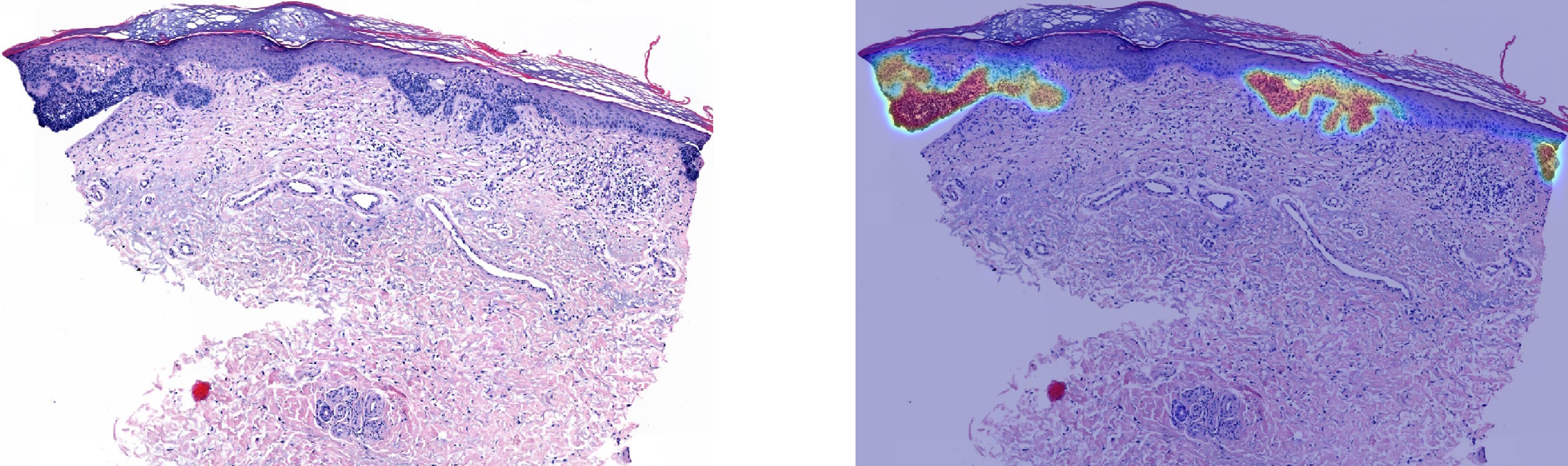}};
\draw (291,740.96) node  {\includegraphics[width=421.5pt,height=146.97pt]{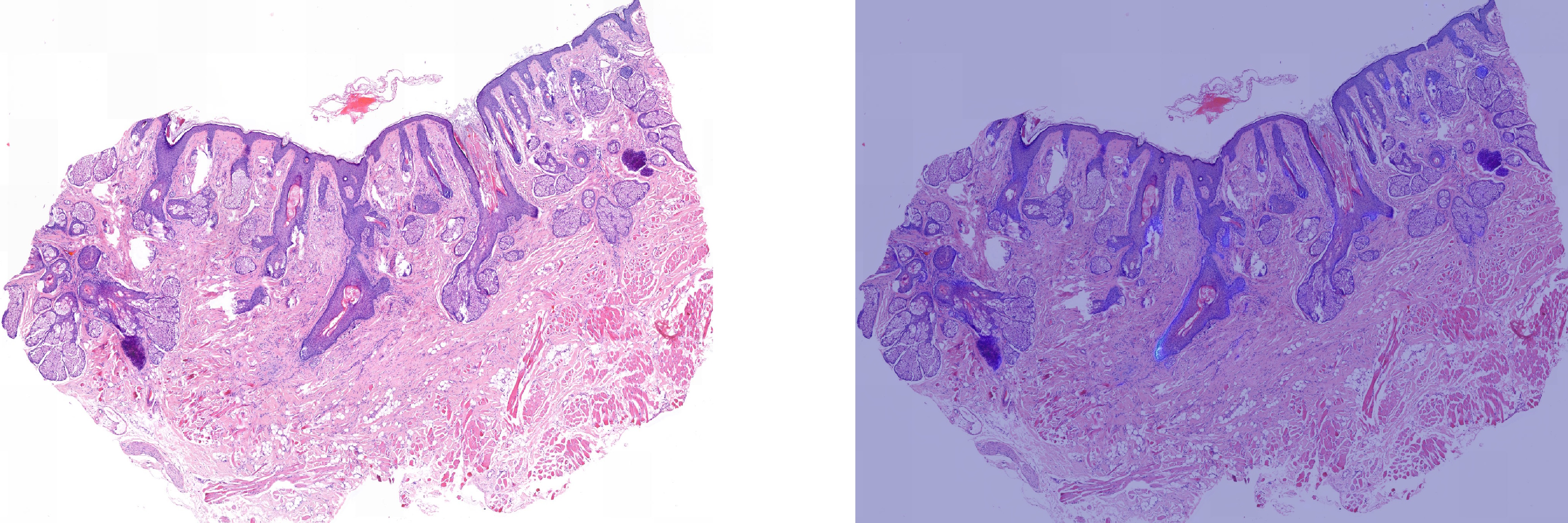}};
\draw   (317,642.97) -- (572,642.97) -- (572,838.94) -- (317,838.94) -- cycle ;
\draw   (10.06,642.97) -- (265.06,642.97) -- (265.06,838.94) -- (10.06,838.94) -- cycle ;
\draw   (317,414.97) -- (572,414.97) -- (572,610.94) -- (317,610.94) -- cycle ;
\draw   (10.06,414.97) -- (265.06,414.97) -- (265.06,610.94) -- (10.06,610.94) -- cycle ;
\draw   (317,212.94) -- (572,212.94) -- (572,380.94) -- (317,380.94) -- cycle ;
\draw   (10.94,212.94) -- (265.94,212.94) -- (265.94,380.94) -- (10.94,380.94) -- cycle ;
\draw   (12,10.94) -- (267,10.94) -- (267,178.94) -- (12,178.94) -- cycle ;
\draw   (317,10.94) -- (572,10.94) -- (572,178.94) -- (317,178.94) -- cycle ;

\draw (131,182) node [anchor=north west][inner sep=0.75pt]  [font=\footnotesize] [align=left] {(a)};
\draw (432,182) node [anchor=north west][inner sep=0.75pt]  [font=\footnotesize] [align=left] {(b)};
\draw (131,384) node [anchor=north west][inner sep=0.75pt]  [font=\footnotesize] [align=left] {(c)};
\draw (431,384) node [anchor=north west][inner sep=0.75pt]  [font=\footnotesize] [align=left] {(d)};
\draw (130,842) node [anchor=north west][inner sep=0.75pt]  [font=\footnotesize] [align=left] {(g)};
\draw (431,842) node [anchor=north west][inner sep=0.75pt]  [font=\footnotesize] [align=left] {(h)};
\draw (130,613) node [anchor=north west][inner sep=0.75pt]  [font=\footnotesize] [align=left] {(e)};
\draw (433,612) node [anchor=north west][inner sep=0.75pt]  [font=\footnotesize] [align=left] {(f)};

\end{tikzpicture}

%% file: figures/false_positive.tex
\tikzset{every picture/.style={line width=0.75pt}} 

\begin{tikzpicture}[x=0.75pt,y=0.75pt,yscale=-1,xscale=1]

\draw (297,1260.4) node  {\includegraphics[width=420pt,height=191.15pt]{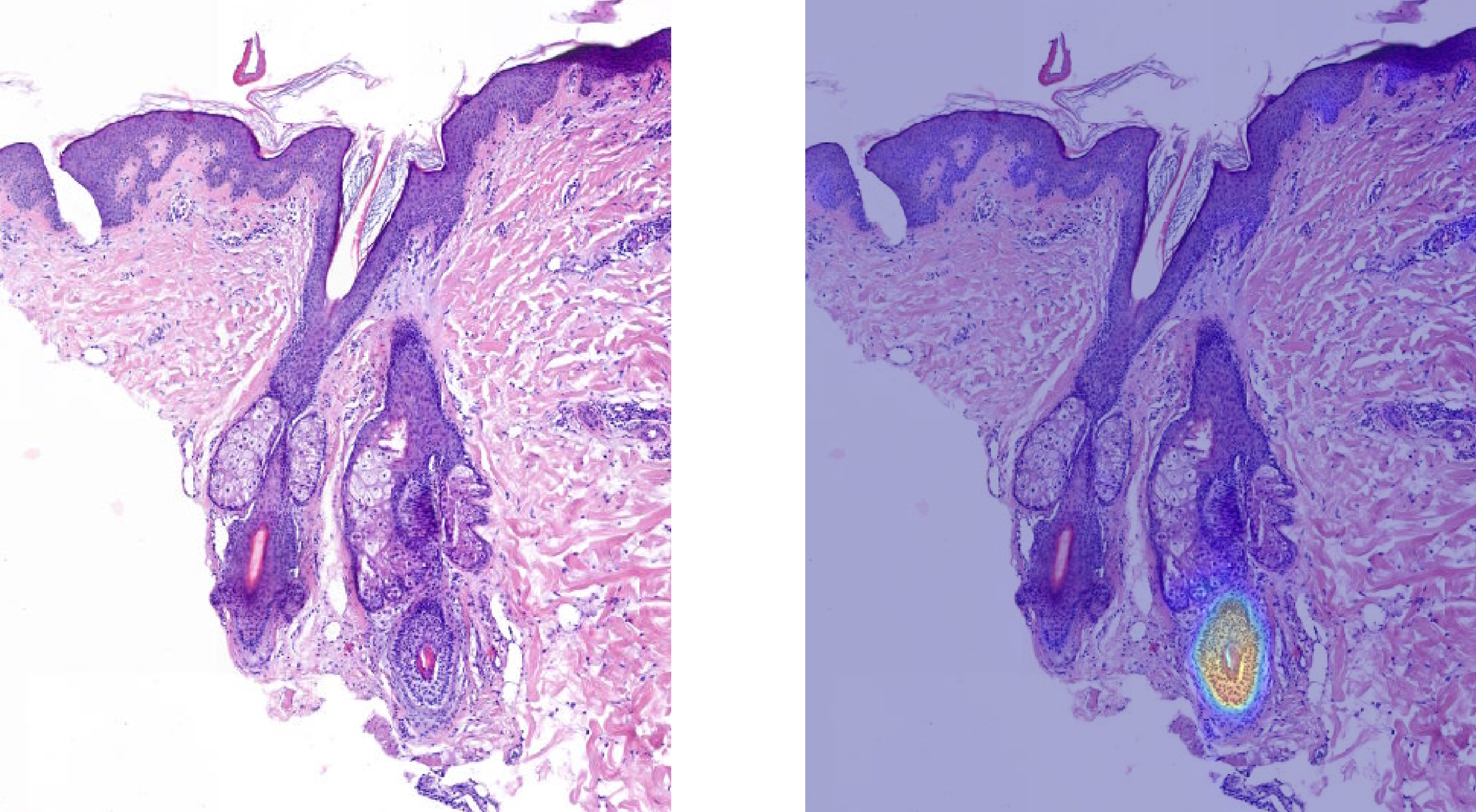}};
\draw   (322,1132.97) -- (577,1132.97) -- (577,1387.83) -- (322,1387.83) -- cycle ;
\draw   (17,1134.08) -- (272,1134.08) -- (272,1388.94) -- (17,1388.94) -- cycle ;

\draw (136,1393) node [anchor=north west][inner sep=0.75pt]  [font=\footnotesize] [align=left] {(a)};
\draw (445,1393) node [anchor=north west][inner sep=0.75pt]  [font=\footnotesize] [align=left] {(b)};

\end{tikzpicture}

%% file: figures/decoder_outputs.tex
\tikzset{every picture/.style={line width=0.75pt}} 

\begin{tikzpicture}[x=0.75pt,y=0.75pt,yscale=-1,xscale=1]

\draw (151.85,442.24) node  {\includegraphics[width=50.78pt,height=53.03pt]{figures/example27230_patch.png}};
\draw (236.96,443.02) node  {\includegraphics[width=50.94pt,height=51.19pt]{figures/example27230_mask.png}};
\draw (151.85,721.02) node  {\includegraphics[width=50.78pt,height=51.19pt]{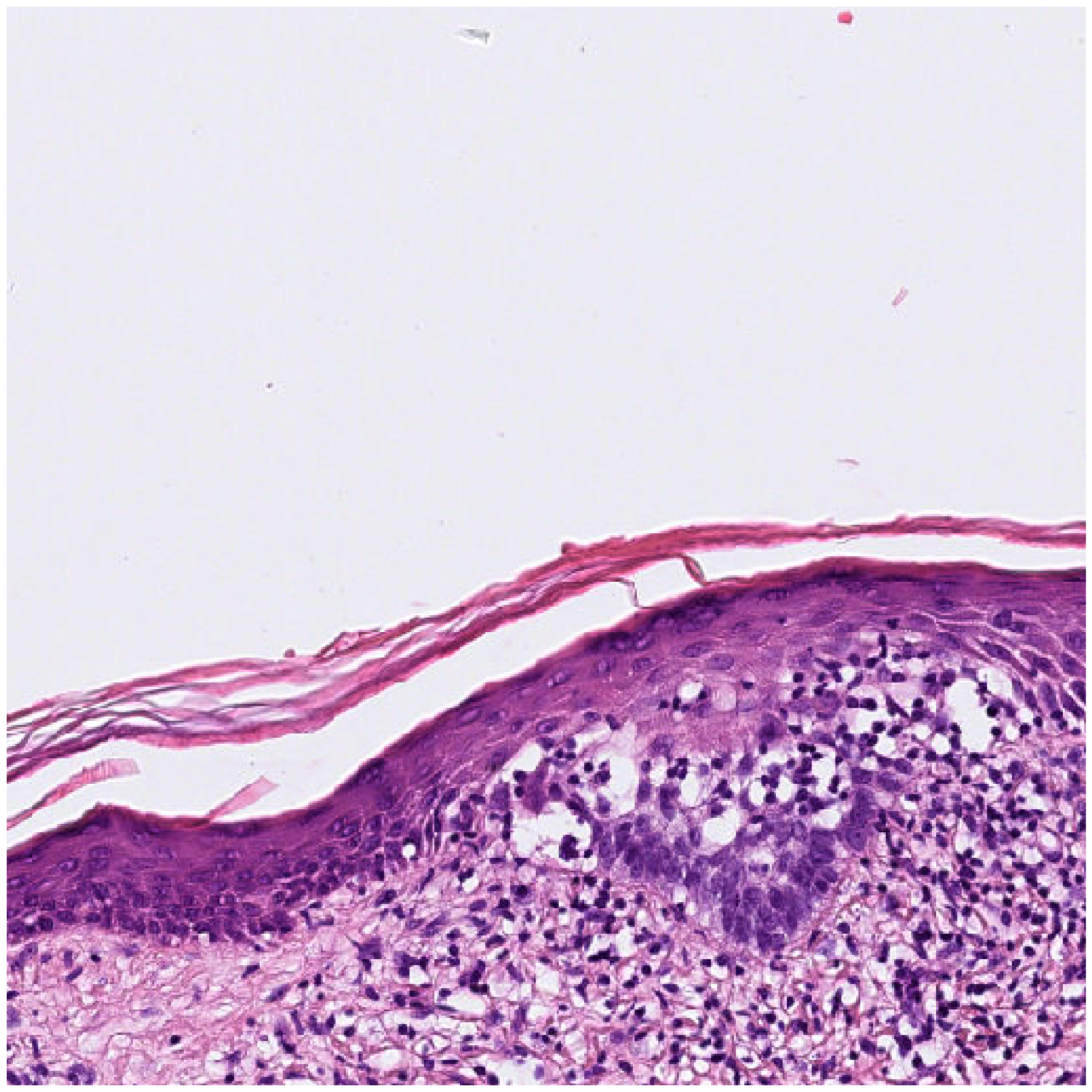}};
\draw (236.85,721.02) node  {\includegraphics[width=50.78pt,height=51.19pt]{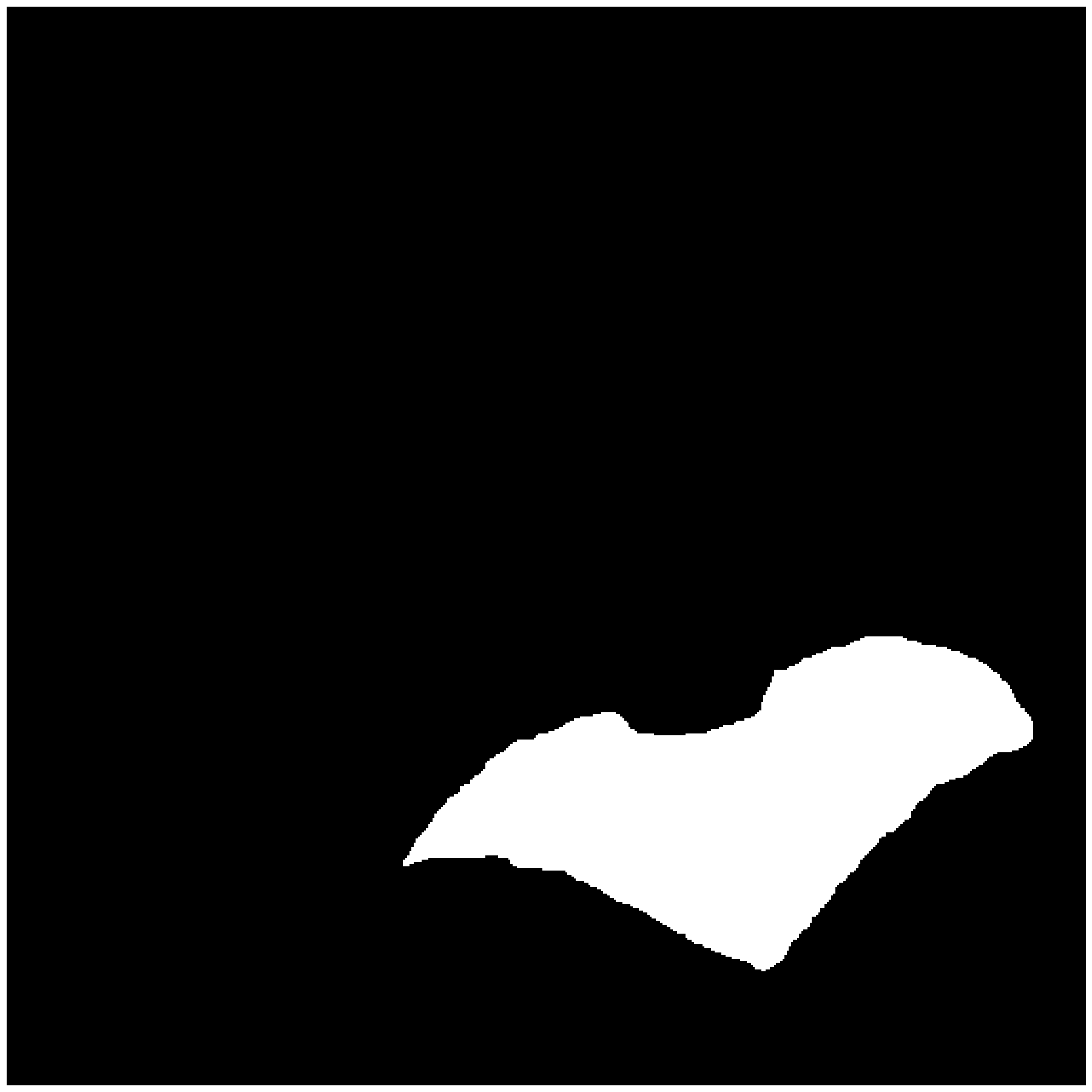}};
\draw (356.53,857.06) node  {\includegraphics[width=480.79pt,height=151.75pt]{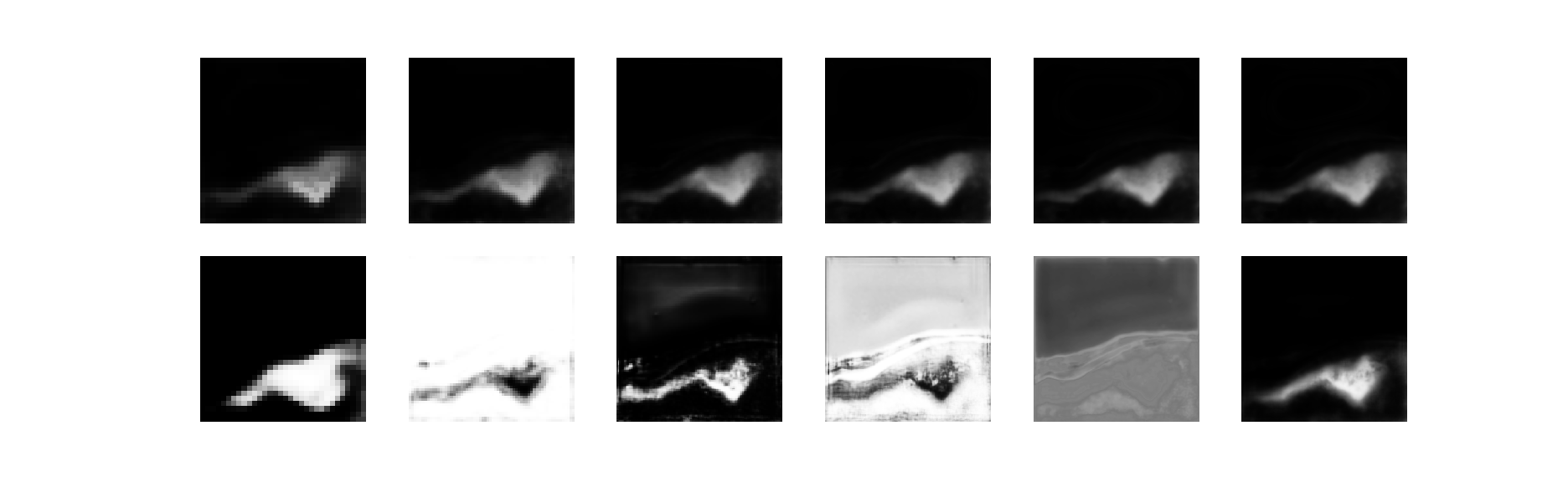}};
\draw (356.53,579.06) node  {\includegraphics[width=480.79pt,height=151.75pt]{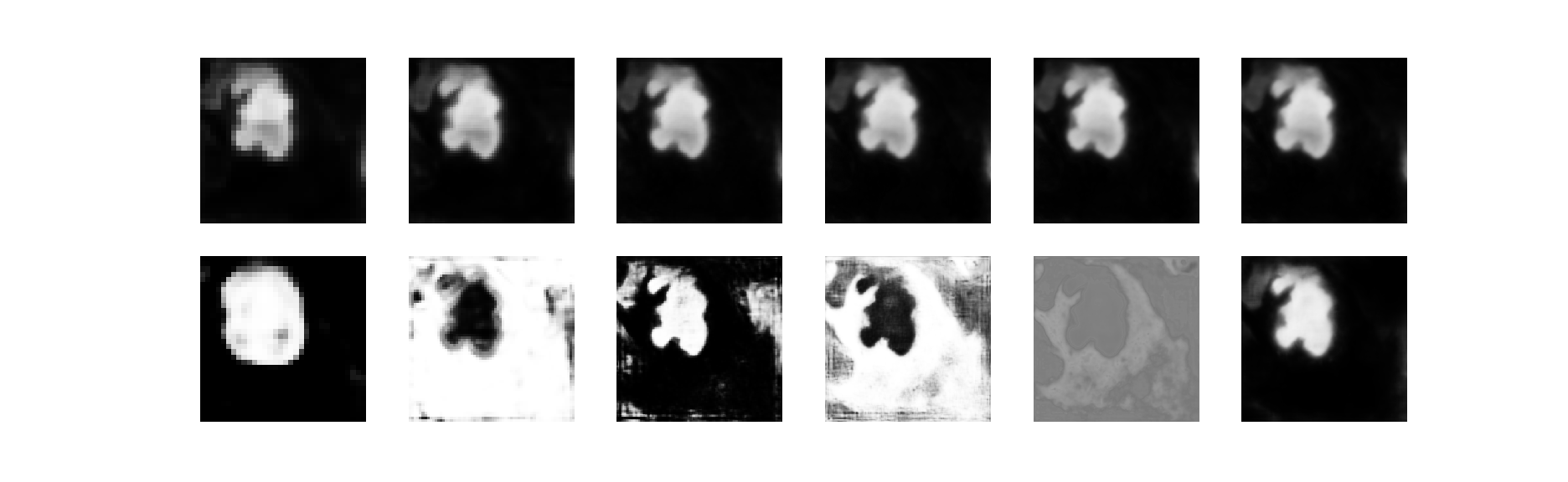}};
\draw (238,167.89) node  {\includegraphics[width=52.5pt,height=52.5pt]{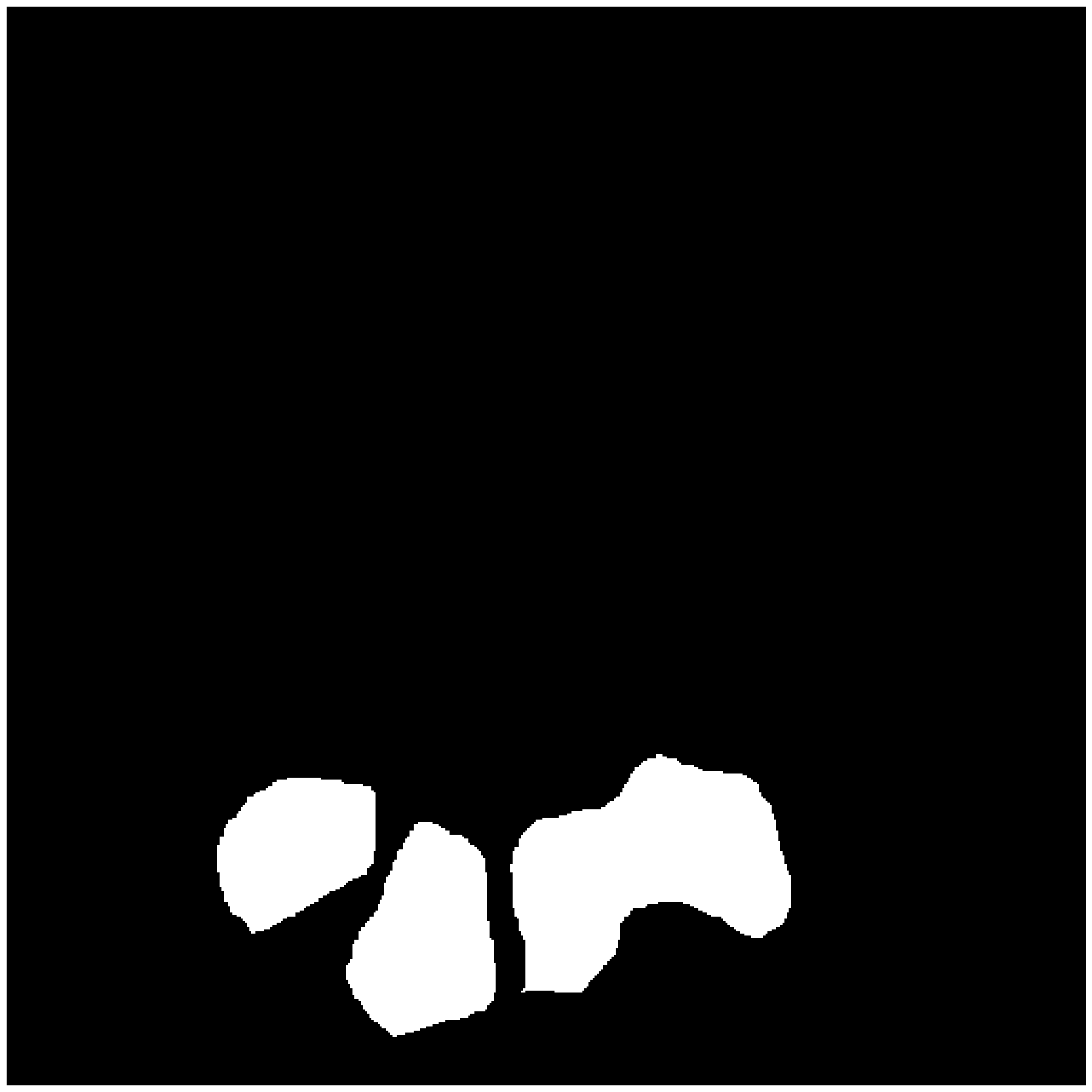}};
\draw (151.85,167.02) node  {\includegraphics[width=50.78pt,height=51.19pt]{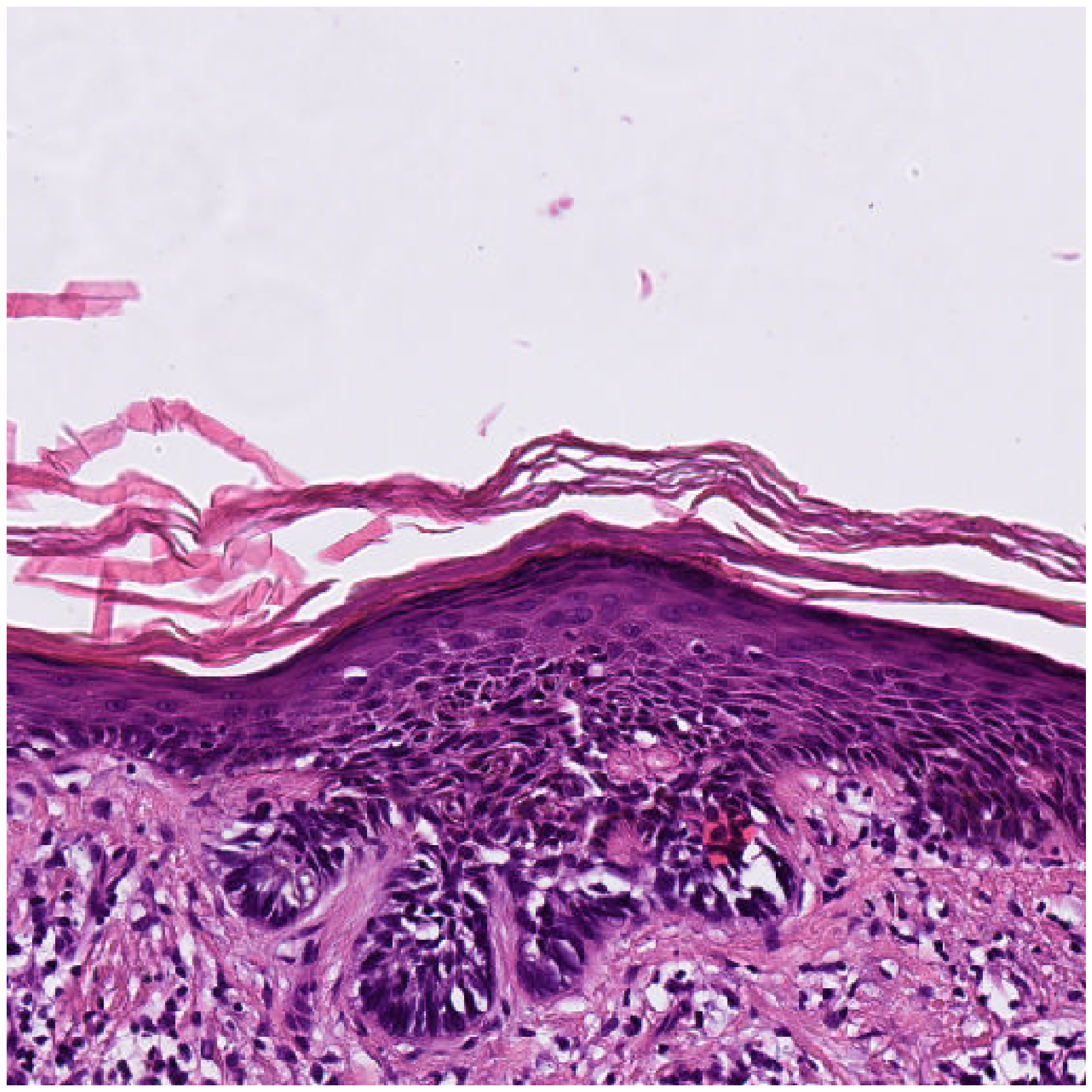}};
\draw (356.53,303.06) node  {\includegraphics[width=480.79pt,height=151.75pt]{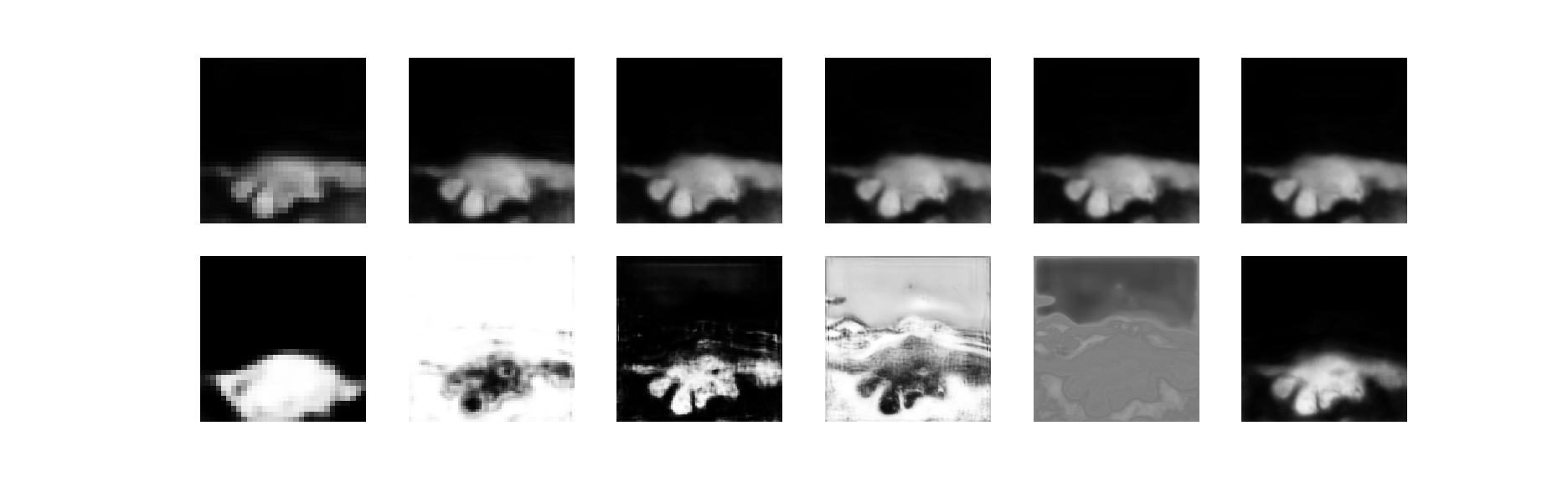}};
\draw   (118,132.89) -- (185.7,132.89) -- (185.7,201.15) -- (118,201.15) -- cycle ;
\draw   (203,132.89) -- (270.7,132.89) -- (270.7,201.15) -- (203,201.15) -- cycle ;
\draw   (118,225.89) -- (185.7,225.89) -- (185.7,294.15) -- (118,294.15) -- cycle ;
\draw   (118,309.89) -- (185.7,309.89) -- (185.7,378.15) -- (118,378.15) -- cycle ;
\draw   (203,309.89) -- (270.7,309.89) -- (270.7,378.15) -- (203,378.15) -- cycle ;
\draw   (203,225.89) -- (270.7,225.89) -- (270.7,294.15) -- (203,294.15) -- cycle ;
\draw   (288,225.89) -- (355.7,225.89) -- (355.7,294.15) -- (288,294.15) -- cycle ;
\draw   (288,309.89) -- (355.7,309.89) -- (355.7,378.15) -- (288,378.15) -- cycle ;
\draw   (373,309.89) -- (440.7,309.89) -- (440.7,378.15) -- (373,378.15) -- cycle ;
\draw   (373,225.89) -- (440.7,225.89) -- (440.7,294.15) -- (373,294.15) -- cycle ;
\draw   (459,225.89) -- (526.7,225.89) -- (526.7,294.15) -- (459,294.15) -- cycle ;
\draw   (459,309.89) -- (526.7,309.89) -- (526.7,378.15) -- (459,378.15) -- cycle ;
\draw   (544,309.89) -- (611.7,309.89) -- (611.7,378.15) -- (544,378.15) -- cycle ;
\draw   (544,225.89) -- (611.7,225.89) -- (611.7,294.15) -- (544,294.15) -- cycle ;
\draw   (118,408.89) -- (185.7,408.89) -- (185.7,477.15) -- (118,477.15) -- cycle ;
\draw   (203,408.89) -- (270.7,408.89) -- (270.7,477.15) -- (203,477.15) -- cycle ;
\draw   (118,501.89) -- (185.7,501.89) -- (185.7,570.15) -- (118,570.15) -- cycle ;
\draw   (118,585.89) -- (185.7,585.89) -- (185.7,654.15) -- (118,654.15) -- cycle ;
\draw   (203,585.89) -- (270.7,585.89) -- (270.7,654.15) -- (203,654.15) -- cycle ;
\draw   (203,501.89) -- (270.7,501.89) -- (270.7,570.15) -- (203,570.15) -- cycle ;
\draw   (288,501.89) -- (355.7,501.89) -- (355.7,570.15) -- (288,570.15) -- cycle ;
\draw   (288,585.89) -- (355.7,585.89) -- (355.7,654.15) -- (288,654.15) -- cycle ;
\draw   (373,585.89) -- (440.7,585.89) -- (440.7,654.15) -- (373,654.15) -- cycle ;
\draw   (373,501.89) -- (440.7,501.89) -- (440.7,570.15) -- (373,570.15) -- cycle ;
\draw   (459,501.89) -- (526.7,501.89) -- (526.7,570.15) -- (459,570.15) -- cycle ;
\draw   (459,585.89) -- (526.7,585.89) -- (526.7,654.15) -- (459,654.15) -- cycle ;
\draw   (544,585.89) -- (611.7,585.89) -- (611.7,654.15) -- (544,654.15) -- cycle ;
\draw   (544,501.89) -- (611.7,501.89) -- (611.7,570.15) -- (544,570.15) -- cycle ;
\draw   (118,686.89) -- (185.7,686.89) -- (185.7,755.15) -- (118,755.15) -- cycle ;
\draw   (203,686.89) -- (270.7,686.89) -- (270.7,755.15) -- (203,755.15) -- cycle ;
\draw   (118,779.89) -- (185.7,779.89) -- (185.7,848.15) -- (118,848.15) -- cycle ;
\draw   (118,863.89) -- (185.7,863.89) -- (185.7,932.15) -- (118,932.15) -- cycle ;
\draw   (203,863.89) -- (270.7,863.89) -- (270.7,932.15) -- (203,932.15) -- cycle ;
\draw   (203,779.89) -- (270.7,779.89) -- (270.7,848.15) -- (203,848.15) -- cycle ;
\draw   (288,779.89) -- (355.7,779.89) -- (355.7,848.15) -- (288,848.15) -- cycle ;
\draw   (288,863.89) -- (355.7,863.89) -- (355.7,932.15) -- (288,932.15) -- cycle ;
\draw   (373,863.89) -- (440.7,863.89) -- (440.7,932.15) -- (373,932.15) -- cycle ;
\draw   (373,779.89) -- (440.7,779.89) -- (440.7,848.15) -- (373,848.15) -- cycle ;
\draw   (459,779.89) -- (526.7,779.89) -- (526.7,848.15) -- (459,848.15) -- cycle ;
\draw   (459,863.89) -- (526.7,863.89) -- (526.7,932.15) -- (459,932.15) -- cycle ;
\draw   (544,863.89) -- (611.7,863.89) -- (611.7,932.15) -- (544,932.15) -- cycle ;
\draw   (544,779.89) -- (611.7,779.89) -- (611.7,848.15) -- (544,848.15) -- cycle ;

\draw (112.88,259.08) node [anchor=east] [inner sep=0.75pt]  [font=\scriptsize] [align=left] {\begin{minipage}[lt]{65.22335600000001pt}\setlength\topsep{0pt}
\begin{flushright}
ResNet34-UNet\\+ DS
\end{flushright}

\end{minipage}};
\draw (112.02,341.73) node [anchor=east] [inner sep=0.75pt]  [font=\scriptsize] [align=left] {\begin{minipage}[lt]{53.106096pt}\setlength\topsep{0pt}
\begin{flushright}
ResNet34-UNet\\+ Linear
\end{flushright}

\end{minipage}};
\draw (152.87,216.24) node  [font=\scriptsize]  {$\psi _{0}( x)$};
\draw (237.18,215.57) node  [font=\scriptsize]  {$\psi _{1}( x)$};
\draw (323.14,215.57) node  [font=\scriptsize]  {$\psi _{2}( x)$};
\draw (407.78,215.57) node  [font=\scriptsize]  {$\psi _{3}( x)$};
\draw (492.09,215.57) node  [font=\scriptsize]  {$\psi _{4}( x)$};
\draw (579.38,216.28) node  [font=\scriptsize]  {$\Phi ( x)$};
\draw (151.76,122.53) node  [font=\scriptsize]  {$x$};
\draw (238.63,122.53) node  [font=\scriptsize]  {$y$};
\draw (112.88,535.08) node [anchor=east] [inner sep=0.75pt]  [font=\scriptsize] [align=left] {\begin{minipage}[lt]{65.22335600000001pt}\setlength\topsep{0pt}
\begin{flushright}
ResNet34-UNet\\+ DS
\end{flushright}

\end{minipage}};
\draw (112.02,617.73) node [anchor=east] [inner sep=0.75pt]  [font=\scriptsize] [align=left] {\begin{minipage}[lt]{53.106096pt}\setlength\topsep{0pt}
\begin{flushright}
ResNet34-UNet\\+ Linear
\end{flushright}

\end{minipage}};
\draw (152.87,492.24) node  [font=\scriptsize]  {$\psi _{0}( x)$};
\draw (237.18,491.57) node  [font=\scriptsize]  {$\psi _{1}( x)$};
\draw (323.14,491.57) node  [font=\scriptsize]  {$\psi _{2}( x)$};
\draw (407.78,491.57) node  [font=\scriptsize]  {$\psi _{3}( x)$};
\draw (492.09,491.57) node  [font=\scriptsize]  {$\psi _{4}( x)$};
\draw (579.38,492.28) node  [font=\scriptsize]  {$\Phi ( x)$};
\draw (151.76,398.53) node  [font=\scriptsize]  {$x$};
\draw (238.63,398.53) node  [font=\scriptsize]  {$y$};
\draw (112.88,813.08) node [anchor=east] [inner sep=0.75pt]  [font=\scriptsize] [align=left] {\begin{minipage}[lt]{65.22335600000001pt}\setlength\topsep{0pt}
\begin{flushright}
ResNet34-UNet\\+ DS
\end{flushright}

\end{minipage}};
\draw (112.02,895.73) node [anchor=east] [inner sep=0.75pt]  [font=\scriptsize] [align=left] {\begin{minipage}[lt]{53.106096pt}\setlength\topsep{0pt}
\begin{flushright}
ResNet34-UNet\\+ Linear
\end{flushright}

\end{minipage}};
\draw (152.87,770.24) node  [font=\scriptsize]  {$\psi _{0}( x)$};
\draw (237.18,769.57) node  [font=\scriptsize]  {$\psi _{1}( x)$};
\draw (323.14,769.57) node  [font=\scriptsize]  {$\psi _{2}( x)$};
\draw (407.78,769.57) node  [font=\scriptsize]  {$\psi _{3}( x)$};
\draw (492.09,769.57) node  [font=\scriptsize]  {$\psi _{4}( x)$};
\draw (579.38,770.28) node  [font=\scriptsize]  {$\Phi ( x)$};
\draw (151.76,676.53) node  [font=\scriptsize]  {$x$};
\draw (238.63,676.53) node  [font=\scriptsize]  {$y$};

\end{tikzpicture}